\documentclass[11pt]{article}
\usepackage{mathrsfs}
\usepackage[centertags]{amsmath}
\usepackage{amsfonts}
\usepackage{amssymb}
\usepackage{appendix} 
\usepackage{amsthm}
\usepackage{cases}
\usepackage{comment}
\usepackage{booktabs}
\usepackage{epsfig}
\usepackage {graphicx}
\usepackage{color}
\usepackage{CJK}
\usepackage{hyperref}
\usepackage{setspace}
\usepackage{algorithm}
\usepackage{algorithmicx}
\usepackage{algpseudocode}
\usepackage{subcaption}
\usepackage{natbib}
\usepackage{rotating} 
\usepackage{lscape}   
\usepackage{siunitx}              
\usepackage{geometry}             
\usepackage{multirow}
\usepackage{array}
\usepackage{tabularx}
\usepackage{makecell}
\usepackage{tablefootnote}

\usepackage{anysize}

\marginsize{1.1in}{1.1in}{0.55in}{0.8in} %

\newtheorem{theorem}{Theorem}

\newtheorem{remark}{Remark}

\def\bB{\mathbf{B}}
\def\bgamma{\boldsymbol{\gamma}}
\def\bbeta{\boldsymbol{\beta}}
\def\balpha{\boldsymbol{\alpha}}
\def\btheta{\boldsymbol{\theta}}
\def\beps{\boldsymbol{\epsilon}}
\def\bdelta{\boldsymbol{\delta}}

\def\bz{\mathbf{z}}
\def\by{\mathbf{y}}

\def\bx{\mathbf{x}}
\def\bSigma{\boldsymbol{\Sigma}}

\def\Var{\operatorname{Var}}
\def\R{\mathbb{R}}
\def\bA{\mathbf{A}}
\def\bD{\mathbf{D}}
\def\bTheta{\boldsymbol{\Theta}}
\def\bB{\mathbf{B}}

\def\ARX{\operatorname{ARX}}
\def\E{\mathbb{E}}

\def\bE{\mathbf{E}}
\def\bGamma{\boldsymbol{\Gamma}}

\def\be{\mathbf{e}}
\def\hbA{\widehat{\mathbf{A}}}
\def\hbD{\widehat{\mathbf{D}}}
\def\hbB{\widehat{\mathbf{B}}}
\def\hbTheta{\widehat{\boldsymbol{\Theta}}}
\def\Pr{\mathbb{P}}
\def\calI{\mathbb{I}}
\def\Score{\operatorname{Score}}
\def\hbGamma{\widehat{\boldsymbol{\Gamma}}}
\def\bGamma{{\boldsymbol{\Gamma}}}

\usepackage{letltxmacro}
\makeatletter
\let\oldr@@t\r@@t
\def\r@@t#1#2{%
\setbox0=\hbox{$\oldr@@t#1{#2\,}$}\dimen0=\ht0
\advance\dimen0-0.2\ht0
\setbox2=\hbox{\vrule height\ht0 depth -\dimen0}%
{\box0\lower0.4pt\box2}}
\LetLtxMacro{\oldsqrt}{\sqrt}
\renewcommand*{\sqrt}[2][\ ]{\oldsqrt[#1]{#2}}
\renewcommand{\ldots}{\cdots}
\renewcommand{\hat}{\widehat}
\renewcommand{\tilde}{\widetilde}

\begin{document}
\begin{CJK}{UTF8}{gbsn}

\title{LLM-Powered CPI Prediction Inference with Online Text Time Series%
\thanks{
Yingying Fan is Centennial Chair in Business Administration and Professor, Data Sciences and Operations Department, Marshall School of Business, University of Southern California, Los Angeles, CA 90089 (E-mail: \textit{fanyingy@marshall.usc.edu}). %
Jinchi Lv is Kenneth King Stonier Chair in Business Administration and Professor, Data Sciences and Operations Department, Marshall School of Business, University of Southern California, Los Angeles, CA 90089 (E-mail: \textit{jinchilv@marshall.usc.edu}). %
Ao Sun is Postdoctoral Scholar, Data Sciences and Operations Department, Marshall School of Business, University of Southern California, Los Angeles, CA 90089 (E-mail: \textit{ao.sun@marshall.usc.edu}). 
Yurou Wang is Ph.D. candidate, Paula and Gregory Chow Institute for Studies in Economics, Xiamen University, China (E-mail: \textit{yurouwang.xmu@gmail.com}). %
}
\date{June 8, 2025}
\author{Yingying Fan$^1$, Jinchi Lv$^1$, Ao Sun$^1$ and Yurou Wang$^2$
\medskip\\
University of Southern California$^1$ and Xiamen University$^2$
\\
} %
}

\maketitle

\begin{abstract} 
Forecasting the Consumer Price Index (CPI) is an important yet challenging task in economics, where most existing approaches rely on low-frequency, survey-based data. With the recent advances of large language models (LLMs), there is growing potential to leverage high-frequency online text data for improved CPI prediction, an area still largely unexplored. This paper proposes LLM-CPI, an LLM-based approach for CPI prediction inference incorporating online text time series. We collect a large set of high-frequency online texts from a popularly used Chinese social network site and employ LLMs such as ChatGPT and the trained BERT models to construct continuous inflation labels for posts that are related to inflation. Online text embeddings are extracted via LDA and BERT. 
We develop a joint time series framework that combines monthly CPI data with LLM-generated daily CPI surrogates. The monthly model employs an ARX structure combining observed CPI data with text embeddings and macroeconomic variables, while the daily model uses a VARX structure built on LLM-generated CPI surrogates and text embeddings. We establish the asymptotic properties of the method and provide two forms of constructed prediction intervals. The finite-sample performance and practical advantages of LLM-CPI 
are demonstrated through both simulation and real data examples. 
\end{abstract}

\textit{Running title}: LLM-CPI


\section{Introduction} \label{Sec.intro}


The consumer price index (CPI) is a key macroeconomic indicator that plays a crucial role in shaping monetary policy and reflecting societal welfare. It is closely linked to interest rates, labor market conditions (e.g., wage growth and employment levels), production costs, and financial market trends \citep{taylor1993discretion, gurkaynak2005sensitivity, borio2007globalisation, blanchard2016phillips}. Traditional methods for forecasting CPI rely on structural economic models and field-collected macroeconomic data. While these methods are fully interpretable, they face two major challenges: declining prediction accuracy \citep{atkeson2001phillips, stock2007has} and high costs of large-scale data collection.

Emerging alternative data streams, particularly text data from news media and social platforms, demonstrate the viability of unstructured content as novel inputs for inflation prediction \citep{larsen2019value, thorsrud2020words, larsen2021news, angelico2022can, hong2025forecasting}. The fact that large language models (LLMs) have capabilities in processing complex linguistic patterns suggests largely untapped potential for economic modeling and research \citep{agrawal2022prediction,brynjolfsson2025generative}. For instance, the bidirectional encoder representations from Transformers (BERT)-based architectures \citep{devlin2018bert} have proven effective in financial volatility modeling \citep{araci2019finbert}, highlighting their adaptability to economic contexts. However, the prediction advantages of LLMs are counterbalanced by their inherent black-box architectures and lack of explaining the sources of prediction power.

The above dilemmas give rise to a core research question: How can we effectively combine the prediction capabilities of LLMs with the interpretability of established economic models? Our suggested framework addresses such challenge through an integrated forecasting system that harmoniously combines traditional econometric models with the LLM-powered prediction models. The method tackles two key obstacles: 1) the simultaneous use of limited high-accuracy low-frequency official data (e.g., monthly government surveys) and abundant high-frequency but less robust LLM-powered surrogates; and 2) the inherent conflict between complex machine learning architectures and the need for interpretable results. Through carefully modeling connections between traditional inflation structure models and LLM-powered surrogates, our approach enhances the prediction inference accuracy while maintaining clear explanations for economic relationships. We name our new framework as the LLM-powered CPI prediction inference (LLM-CPI). 

\begin{figure}[t]
    \centering
    \includegraphics[width=0.9\linewidth]{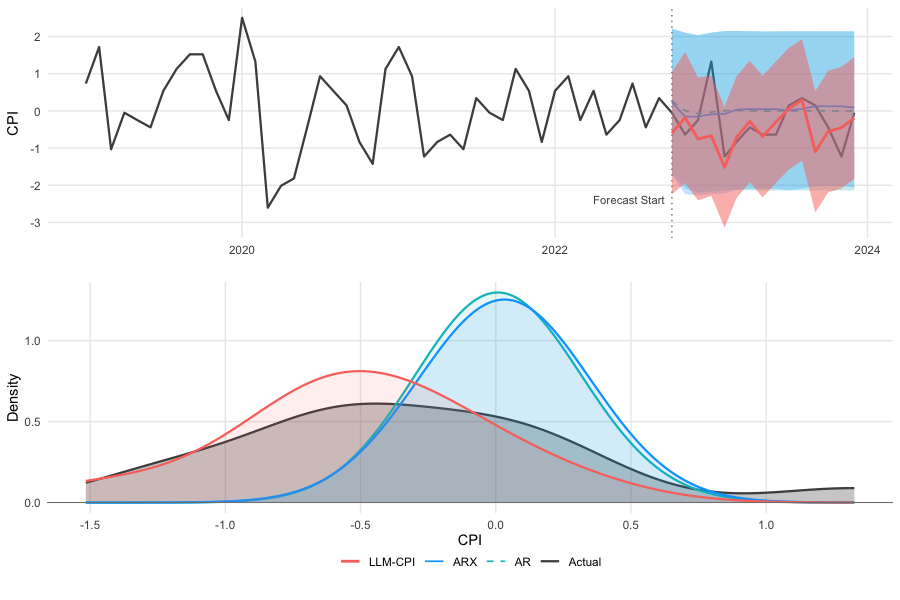}
    \caption{Top panel: The CPI forecasts and corresponding prediction intervals across different dates given by the AR, ARX (unemployment rate as exogenous), and LLM-CPI models. Bottom panel: The kernel density plots of predicted CPI values given by different models compared to the actual CPI values. The LLM-CPI model exploits the LLM-generated synthetic surrogates constructed using  ChatGPT and the trained BERT model, and the LDA embeddings for online text embeddings, as discussed in Section \ref{sec:real_data.descripreproc}. 
    } 
    \label{fig:illus_preidction}
\end{figure}

Figure \ref{fig:illus_preidction} glimpses the prediction power of the suggested LLM-CPI method in forecasting high-frequency inflation dynamics. The black solid curve represents the standardized actual CPI values. We adopt the period from January 2019 to September 2022 as the training sample, and the period from October 2022 to December 2023 to evaluate the out-of-sample forecasts using three models: 
\begin{itemize}
\item A classical autoregressive (AR) model relying on the historical CPI values (green dashed curve);
\item An autoregressive model with the unemployment rate as an exogenous predictor (ARX), also referred to as Gordon's ``triangle model'' in \cite{gordon1988us} (blue solid curve);
\item Our suggested LLM-CPI model that combines the prediction power of the structural ARX model and the LLM-powered surrogate model (red solid curve)\footnote{The full details on the model construction can be found in Section \ref{Sec.LLM-CPI}.}.
\end{itemize}
The results clearly show that the LLM-CPI most closely tracks the actual CPI trajectory, successfully capturing turning points and underlying trends. More importantly, the prediction interval given by the LLM-CPI model is substantially narrower than those given by the AR and ARX models, while still maintaining the desired nominal coverage rate (i.e., $95\%$). In contrast, the prediction intervals from traditional models are excessively wide and less informative. These findings highlight the effectiveness of the LLM-CPI in delivering both accurate point forecasts and 
tight uncertainty quantification, especially in the presence of complex economic signals.

\subsection{Related works} \label{Sec.relawork}

Our paper is related to the literature on economic narratives. Economic narratives, stories that provide interpretations of economic events or suggest theories about the economy, spread in a manner similar to infectious diseases \citep{shiller2020narrative}. Within this paradigm, competing economic narratives exhibit distinct lifecycles characterized by viral spread, transient prominence, or rapid obsolescence. Crucially, these narrative ecosystems generate behavioral externalities: the aggregate behavioral responses to these competing narratives--whether welfare-enhancing or distortionary--constitute a fundamental driver of economic fluctuations \citep{shiller2020popular}. Many theoretical, experimental, and empirical studies have examined the important role of economic narratives in economic fluctuations \citep{larsen2019value, chahrour2021sectoral, cookson2024social}. \cite{larsen2021news} integrated large-scale news data analysis and economic theoretical models to explore the role of economic narratives in shaping household inflation expectations. \cite{weber2022subjective} highlighted the declining effectiveness of traditional media for public communication, driven by reduced readership and lower perceived credibility compared to social media and personal networks, as revealed by direct survey evidence. Existing studies predominantly employ 
classical text analysis of traditional media sources to assess economic narratives' predictive capacity. In contrast, our approach leverages interdependencies between the observed monthly CPI data and the LLM-generated daily inflation proxies from social media to produce prediction intervals with statistically validated asymptotic coverage. 

Our study contributes to recent advances in LLMs empowering economic and business research. Recent advancements in LLMs have shown potential to support economic and business research through their ability to generate synthetic data that approximates real-world patterns \citep{brand2023using, horton2023large}. These models enable cost-effective experimentation in applications such as market simulations and policy evaluations, offering researchers a flexible tool for preliminary analysis. However, the reliability of LLM-driven insights remains uncertain due to challenges such as inherent biases, reliance on outdated training data, and limited adaptability to real-time events \citep{goli2023can, de2025chatgpt, ye2025lola}. These limitations highlight the importance of developing complementary methods to evaluate and refine the LLM outputs. In parallel, recent methodological works in \textit{prediction-powered inference} and \textit{synthetic surrogate joint modeling} have sought to address these issues by integrating LLM-generated predictions into statistical testing workflows. For example, \cite{angelopoulos2023prediction} and \cite{zrnic2024cross} proposed combining experimental and synthetic data to improve estimation accuracy through cross-validation, while \cite{mccaw2024synthetic} mitigated instability in raw LLM predictions by treating them as synthetic proxies to augment traditional models by the joint likelihood approach. Our work explores the integration of two parallel research streams and suggests a framework that aligns their complementary strengths in time-series prediction settings.

Our work also contributes to the literature on modern inflation forecasting, a long-studied challenge in economics \citep{gordon1988us, stock1999forecasting, atkeson2001phillips}. Recent advances in data availability have enabled new approaches to this problem. For instance, \cite{medeiros2021forecasting} demonstrated improved inflation predictions by applying machine learning methods to a broad set of macroeconomic indicators from \cite{mccracken2016fred}. More recently, \cite{hong2025forecasting} incorporated text data from Wall Street Journal articles to enhance forecasting accuracy. Building upon these developments, we explore an alternative approach that integrates both text and macroeconomic data sources while leveraging the power of LLMs for prediction and inference. Our 
results suggest that this combined method can offer significant improvements over existing approaches. 

\subsection{Our innovations} \label{Sec.ourinnov}


Our study first suggests a novel approach to constructing an LLM-generated daily inflation index for the China economy. Leveraging five years of data from Sina Weibo (\url{https://www.weibo.com}), the largest social media platform in China, spanning the period from January 1, 2019 to December 31, 2023, we address the challenge of identifying inflation-related content within a massive and noisy text data set by developing a three-stage LLM-based learning framework. Specifically, we first employ the chain-of-thought prompting strategy \citep{wei2022chain} to guide LLMs in annotating a randomly selected subset of the text data. We then utilize this annotated subset to fine-tune three distinct BERT models to sequentially filter out irrelevant or advertisement content, identify posts related to inflation, and assign a continuous inflation score to each identified post. Finally, we construct an online text-based daily inflation index by averaging the inflation scores of validated inflation-related posts on a daily basis. We name the constructed daily average score as the \textit{LLM-generated daily inflation index}. Such LLM-generated daily inflation index has the potential to serve as a high-frequency inflation monitoring tool, providing a valuable complement to the official monthly CPI.

We further propose the LLM-powered CPI prediction inference (LLM-CPI) framework by integrating the LLM-generated daily inflation index with conventional monthly CPI measurements. LLM-CPI only requires the error correlation assumption, rather than precise alignment between the social media-based daily inflation index and official monthly inflation. The LLM-CPI framework combines a text-embedding-augmented autoregressive model for the observed monthly CPI and a vector autoregressive model for the LLM-generated daily inflation index. These two models are interconnected by their cross-sectional correlation structure of errors \citep{mccaw2024synthetic}. Thus, our framework extends the method of \cite{mccaw2024synthetic} to the time-series data settings. Such architecture achieves superior forecast accuracy compared to conventional approaches using only historical CPI, macroeconomic indicators, or text embedding features. In addition, our framework provides tight prediction intervals that enjoy theoretical guarantees, as confirmed by the simulation and real data examples.


The rest of the paper is organized as follows. Section \ref{Sec.textinflaembed} introduces the collected online text data as well as the high-frequency online text-based inflation index and text embeddings. We suggest the new method of LLM-powered CPI prediction inference (LLM-CPI) and present its asymptotic theory in Section \ref{Sec.LLM-CPI}. Section \ref{sec:simu} provides several simulation examples verifying the finite-sample performance of our method. We showcase the practical advantages of the newly suggested method through a real data application on CPI prediction inference in Section \ref{sec:real_data}. Section \ref{sec:discu} discusses some implications and extensions of our work. All the proofs and additional technical and empirical details are provided in the Supplementary Material.

\section{High-frequency online text-based inflation index and text embeddings} \label{Sec.textinflaembed}

We introduce in this section our collected online text data set, the construction of high-frequency online text-based inflation index via LLMs, and text embeddings.

\subsection{Online text data collection} \label{Sec.textcollec}

Sina Weibo, a prominent social media platform in China, serves as a critical real-time information channel for journalists and consumers alike, with 588 million monthly active users as of March 2024 \citep{weibo2024}. This platform hosts discussions spanning diverse topics, including politics, technology, and economic trends, and has been widely studied for its role in shaping public discourse within financial and economic contexts \citep{feng2019top, qin2024social}. As a public forum for sharing personal opinions and experiences, Weibo provides unique insights into consumer perspectives on inflation. We suggest leveraging such platform to capture real-time inflation-related narratives, offering temporal granularity comparable to traditional survey methods while reflecting grassroots economic sentiments.



When users publish posts on Weibo, these posts become immediately visible to their followers and can be reshared through subsequent reposts, mirroring Twitter's information dissemination model. These posts, encompassing news articles, hyperlinks, opinion statements, advertisements, and personal updates, remain publicly accessible via the platform's search interface.


To effectively capture consumer perceptions of inflation, we exploit a keyword filtering method based on \cite{angelico2022can}, adapted to the unique features of the Chinese language, such as words with multiple meanings. Since housing costs play a major role in influencing consumption and savings behavior in China, our keyword list includes terms related to real estate prices. Using Weibo's advanced search tools, we collect posts that contain any of the specified keywords, with full details of the procedure provided in Section \ref{new.Sec.textdatacoll.1}. The final keyword list consists of $25$ Chinese terms, grouped into categories (with English translations in parentheses)\footnote{{ We intentionally include broad keywords, such as ``Price'' and ``House price,'' to ensure the completeness of important search results.}}:
\begin{itemize}
    \item \textbf{General price}: 价格 (Price), 租金  (Rental fee), 成本 (Cost), 费用 (Fee), 钱 (Money), 油价 (Oil price), 房屋价格 (House price), 房租 (Housing rent), 房贷 (Mortgage), 房地产 (Real estate), 楼市 (Property market), 新房 (New house), 二手房 (Used house), 租房 (Renting a house), 买房 (House buying), 卖房 (House selling), 房价 (House price (abbr.));
\item \textbf{Inflation}: 通货膨胀 (Inflation), 涨价 (Price rise), 贵 (Expensive), 涨 (Rise/Increase);
\item \textbf{Deflation}: 通货紧缩 (Deflation), 降价 (Price reduction), 便宜 (Cheap), 跌 (Decline/Decrease).
\end{itemize}

We collect all posts that contain at least one of the selected keywords on 
Weibo from January 1, 2019 to December 31, 2023. The initial data set comprises approximately $119.8$ million posts; see Table \ref{tab:social_media_stati} in Section \ref{new.Sec.textdatacoll.1} for details. We should emphasize that the collected raw data set not only pertains to inflation-related contents, but also includes contents related to advertisements, E-commerce websites, and sales promotions.

\subsection{High-frequency online text-based inflation index constructed via LLMs} \label{textinflaconstr}



Our text analysis begins with standard text preprocessing steps to remove invalid text and posts with incomplete timestamps from the raw Weibo data set, as discussed in Section \ref{new.Sec.textdatacoll.2}. A more challenging issue is the overwhelming presence of commercial contents unrelated to inflation. Identifying inflation-related patterns in social media posts is particularly difficult due to three key factors: 1) overlapping content types, such as promotional posts versus personal experiences, 2) the informal and varied language used in user-generated content, and 3) diverse ways users express opinions about price changes. An even greater obstacle is the lack of natural labels to identify commercial posts. Directly applying unsupervised learning to identify inflation-related posts has limited effectiveness because of severe class imbalance, with inflation-specific contents being far less common compared to general commercial posts.

To address these practical challenges, we develop an LLM-based learning framework. Our process starts with employing a chain-of-thought prompting strategy using ChatGPT (i.e., GPT-4-turbo-2024-04-09) to annotate a stratified random sample of $20,000$ Weibo posts. The ChatGPT model has demonstrated superior performance in text annotation tasks due to its advanced contextual understanding and few-shot learning capabilities \citep{gilardi2023chatgpt}. The annotation procedure is hierarchical in that it first annotates all sampled posts into advertisement and non-advertisement categories, and then non-advertisement posts are further classified into five distinct categories: [Inflation, Lifestyle, Entertainment, Emotion, News]. The posts categorized as inflation are further assigned specific inflation severity scores, as detailed in Section \ref{new.Sec.inflation}. Such prompting strategy enables hierarchical decision-making in text annotation by sequentially evaluating conditional logic (e.g., first assessing topic relevance before sentiment polarity), significantly improving annotation accuracy for multi-layered tasks \citep{wei2022chain}.

We next fine-tune two BERT models, \textit{Advertisement-BERT} and \textit{Category-BERT}, to accurately identify post categories related to price fluctuations.  Given the Chinese-language nature of our text data, we utilize the ``bert-base-chinese'' architecture\footnote{\url{https://huggingface.co/google-bert/bert-base-chinese}}, a pre-trained LLM optimized for Chinese text processing through whole-word masking and character-level tokenization. Our suggested framework operates sequentially. The fine-tuned Advertisement-BERT model performs binary classification to filter commercial content using labeled training data, and then the fine-tuned Category-BERT model categorizes non-advertisement posts into five thematic groups, with Inflation-defined posts capturing explicit price-related narratives. Such cascaded filtering reduces the original data set to $5.79$ million inflation-relevant posts (i.e.,  $4.8\%$ retention rate) from $1.49$ million unique users, representing authentic consumer perspectives on price dynamics; see Section \ref{new.Sec.inflation} for details. The refined corpus enables two downstream applications: high-frequency online text-based inflation index construction through temporal aggregation of post-level sentiment scores, and text embedding construction for the CPI prediction inference.

In the final processing stage, we fine-tune the \textit{CPI-BERT} model using continuous inflation scores ($\operatorname{Score}_i \in [0,1]$) generated by ChatGPT. Such regression-optimized architecture predicts the fine-grained sentiment intensity for each of $N=5,790,457$ validated posts, where each score quantifies perceived inflation at the individual post level; see Section \ref{new.Sec.inflation} for more details. Figure \ref{fig:indexup_down} in Section \ref{new.Sec.LLM-geninflfluc} depicts the daily volumes of inflation-up and inflation-down posts on this final stage results. Such high-frequency narrative data provides complementary insights to the official inflation metrics, capturing grassroots economic sentiments often omitted from traditional indicators.

\begin{figure}[t]
    \centering
    \includegraphics[width=14cm]{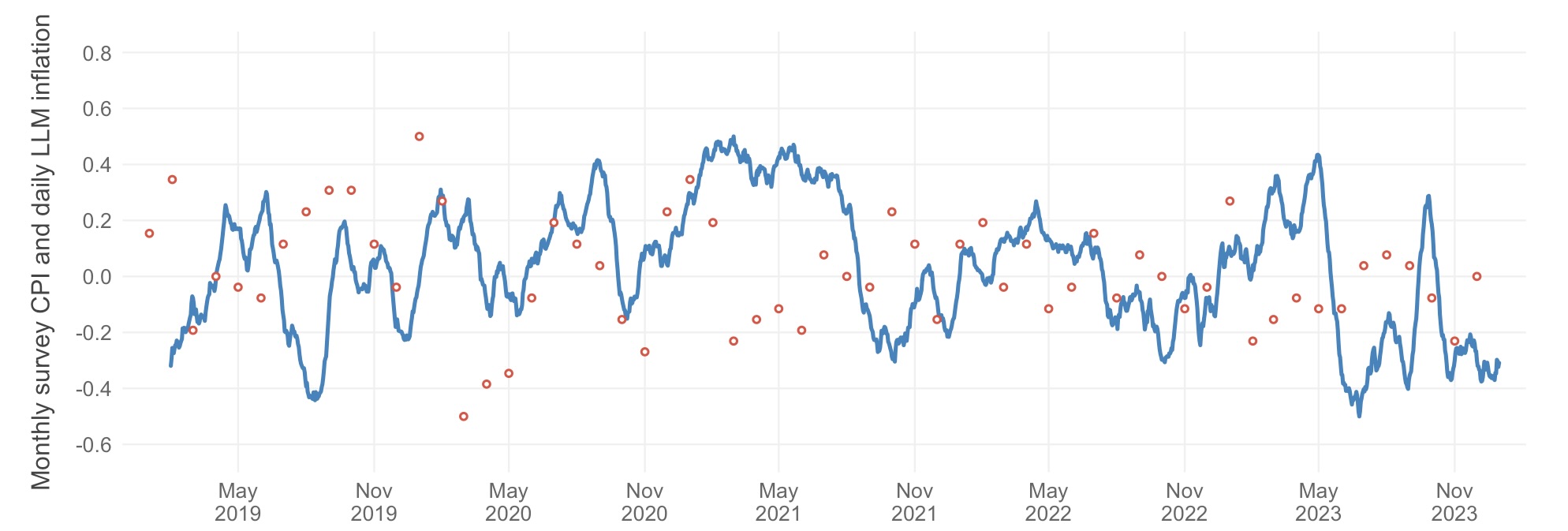}
    \caption{A time-series comparison between the LLM-generated daily inflation index and the observed CPI from January 2019 to December 2023. The blue curve represents the LLM-generated daily inflation index capturing high-frequency variations through analysis of unstructured Weibo posts, and the red circles depict the monthly observed CPI.}
    \label{fig:WB_cpi}
\end{figure}

To construct the LLM-generated daily inflation index, we pair each post's continuous inflation score with its publication date, forming a data set $\{(\Score_i, \operatorname{Date}_i), \, i=1,\ldots, N\}$, where $\operatorname{Date}_i$ denotes the posting date of the $i$th Weibo entry. The \textit{LLM-generated daily inflation index} for day $d$ is defined as 
\begin{equation}\label{equ:CPI_daily_inflation}
\operatorname{Inflation}_d = \frac{\sum^{N}_{i=1}\Score_i \mathbb{I}(\operatorname{Date}_i = d)}{\sum^{N}_{i=1}\mathbb{I}(\operatorname{Date}_i = d)},    
\end{equation}
where $\mathbb{I}(\cdot)$ is the indicator function. 
Figure \ref{fig:WB_cpi} plots the LLM-generated daily inflation index $\operatorname{Inflation}_d$ in (\ref{equ:CPI_daily_inflation}) and the monthly CPI that we collect from China National Bureau of Statistics (CNBS). To facilitate the comparison of trend changes, we apply a $30$-day moving average smoothing to $\operatorname{Inflation}_d$ in the figure. Additionally, both the monthly survey-based CPI and the LLM-generated daily inflation index are standardized and shifted by subtracting $0.5$, ensuring that their fluctuations are centered around zero. The LLM-generated daily inflation index (blue curve) exhibits high-frequency fluctuations while maintaining a consistent long-term trend with the monthly survey-based CPI (red circles), demonstrating alignment between real-time unstructured text analysis and traditional macroeconomic measurement methodologies. Both time series reflect similar inflationary patterns over the observed period, with the LLM-generated daily inflation index capturing finer-grained volatility that converges toward the monthly survey-based benchmark. 

\subsection{Online text embeddings} \label{Sec.textembed}


Our suggested LLM-CPI framework incorporates two text embedding methods: the topic probability embeddings from the latent Dirichlet allocation (LDA) \citep{blei2003latent}, and the BERT embeddings extracted from the fine-tuned CPI-BERT model architecture. Specifically, we implement the LDA model to derive topic probability distributions from the text data. Each document is represented as a $K$-dimensional vector, where elements correspond to posterior probabilities of membership in $K$ latent thematic clusters. Following established optimization criterion \citep{blei2003latent}, we configure $K=20$ to balance semantic coherence against model complexity. These document-topic distributions are temporally aggregated to 
monthly through 
averaging. For the BERT embeddings, we extract $768$-dimensional vectors through mean pooling of the final hidden layer right before the output layer of the fine-tuned CPI-BERT model (i.e., a deep neural network), capturing semantic patterns in individual posts. These post-level embeddings are averaged within each 
month to create monthly 
LLM-based economic text features. More details on how to construct different economic text embedding features can be found in Section \ref{app:sec:embeddings}.

We emphasize that these embedding features, particularly those derived from the trained BERT models, are high-dimensional, while the number of target observations is limited to $60$ monthly CPI index values spanning from January 2019 to December 2023. Consequently, it is crucial to apply a suitable model selection technique to identify the most informative and predictive components of the text embeddings; see Section \ref{app:sec:model_selection} for details on time-series model selection. We denote the selected LDA embedding features as $\bx_t^{\text{LDA}}$, and the selected BERT embedding features as $\bx_t^{\text{BERT}}$. When not distinguishing between the two, we refer to them generically as text embedding features $\bx_t$. 

\section{LLM-powered CPI prediction inference} \label{Sec.LLM-CPI}

In this section, we introduce the framework of the LLM-powered CPI prediction inference (LLM-CPI) exploiting the online text time series obtained in Section \ref{Sec.textinflaembed}, and establish its theoretical justifications. 

\subsection{A joint time-series model for CPI and text-based inflation index} \label{Sec.jointmodel}

A key ingredient of the suggested LLM-CPI method is a joint time-series model integrating a target CPI model on the observed monthly CPIs and an LLM-powered surrogate model on the LLM-generated daily inflation index constructed in Section \ref{textinflaconstr}. Let $\{y_{t} \in \R, t =1,\ldots, T\}$ be the observed standardized monthly CPI time series\footnote{The CPI measures the price inflation of the current month relative to the previous month, with the previous month set as the baseline of $100$, and the detailed definition of  $y_t$  can be found in Section \ref{sec:real_data.descripreproc}.}. The sign of $y_t$ (positive or negative) indicates whether inflation has increased or decreased compared to the prior month. The CPI index is widely regarded as the golden standard for measuring inflation \citep{stock1999forecasting}. In addition to the CPI, we collect other monthly macroeconomic indicators, such as the unemployment rate, that are potentially related to inflation. These macroeconomic indicators are denoted as $\bz_t \in \R^d$, where $d$ represents the dimensionality of the macroeconomic covariate vector. Further, as in \cite{hong2025forecasting} we incorporate the monthly economic 
text embedding features $\bx_t \in \mathbb{R}^p$, constructed in Section \ref{Sec.textembed}, into modeling the dynamics of CPI.

We begin with introducing the target CPI model on the observed monthly CPIs {$\{y_{t}, t =1,\ldots, T\}$}. To this end, we employ an autoregressive model with exogenous variables of order $q_1$, referred to as $\ARX(q_1)$ model. Then the target CPI model on the observed monthly CPIs $y_{t}$ is defined as the $\ARX(q_1)$ model 
\begin{equation}\label{equ:cpi}
y_{t} =  \sum_{l=1}^{q_1} \alpha_{l} y_{t-l} + \bz_t^\top \btheta + \bx_t^\top \bbeta + \epsilon_t
\end{equation}
with $t =1,\ldots, T$, where $\balpha = (\alpha_1,\ldots, \alpha_{q_1})^\top \in \R^{q_1}$ denotes the autoregressive coefficient vector, $\btheta \in \R^d$ represents the regression coefficient vector for the macroeconomic indicators, $\bbeta \in \R^p$ stands for the regression coefficient vector for the text embedding features, and $\epsilon_t$ is the scalar model error.


Our key innovation is to leverage the LLM-generated daily inflation index, constructed in Section \ref{textinflaconstr}, to empower the CPI prediction inference. Since the LLM-generated daily inflation index highly fluctuates, we smooth it into three ten-day periods within each month: the first ten days, the middle ten days, and the remaining days of the month. Such three-period granularity each month allows the LLM-generated inflation index to provide more timely insights compared to the observed monthly CPI, while reducing the randomness of the LLM-generated daily inflation index. We denote the resulting LLM-generated inflation index as $\{y_{t,k}^S \in \R, t =1,\ldots, T, k =1,\ldots, K\}$, where $K=3$ represents the three periods within each month, and each $y_{t,k}^S$ corresponds to a surrogate of the CPI generated by LLMs (i.e., ChatGPT and the trained BERT models) for the $k$th period of the $t$th month.

While the LLM-generated inflation index might be related to the CPI, it cannot be directly used as a substitute or prediction of the true inflation measurements. This is due to two intrinsic limitations of the LLM-based predictions. First, the LLM-based predictions are inherently stochastic, meaning that different runs or slight variations in prompts can yield different results. Such intrinsic randomness incurs the reproducibility challenges, making it difficult to rely on the LLM-based predictions alone for consistent, reliable forecasting. Second, the black-box nature of LLMs obscures the underlying mechanisms driving their predictions, and causes the potential risk of bias. Such lack of transparency raises concerns on the interpretability and reliability of the LLM-based forecasts alone.

Instead of directly using the LLM-generated inflation index values as the final outputs, we incorporate them as the synthetic surrogates to enhance the prediction and inference of the target CPI by exploiting the correlations between the target CPI and the LLM-generated inflation index. Our LLM-CPI method is inspired by the recent work of \cite{mccaw2024synthetic}, who introduced a general framework for integrating synthetic surrogates to empower the testing procedure for genome-wide association studies. Building upon this, we adapt and extend their approach to the time series prediction framework, enabling the incorporation of LLM-generated synthetic surrogates into joint time-series modeling. To formalize this idea, we introduce the LLM-powered surrogate model on the LLM-generated inflation index $\{y_{t,k}^S \in \R, t =1,\ldots, T, k =1,\ldots, K\}$ that is defined as the vector autoregressive model with exogenous variables of order $q_2$ (without loss of generality, we assume that $q_2 \le q_1$), referred to as VARX$(q_2)$ model,
\begin{equation}\label{equ:sur}
    \by^S_t = \sum_{l=1}^{q_2} \bA^S_{l}  \by^S_{t-l} + \bB^S \bx_t + \beps^S_t,
\end{equation}
where $\by^S_t = (y_{t,1}^S, \ldots, y^S_{t,K})^\top \in \R^K$ contains the LLM-generated inflation index values over three periods in the $t$th month, $\bA^S_{l} \in \R^{K \times K}$ with $l =1,\ldots, q_2$ denote the autoregressive coefficient matrices, $\bB^S  \in \R^{K \times p}$ represents the regression coefficient matrix for the exogenous text embedding features, and $\beps^S_t = (\epsilon^S_{t,1}, \cdots, \epsilon^S_{t,K})^\top$ is the model error vector. The LLM-powered surrogate model (\ref{equ:sur}) is not intended to represent the true data-generating process of the LLM predictions, but serves as a working model to capture their temporal dynamics and relationships with the text data. Such model is expected to be useful when the text embedding features $\bx_t$ carry economically meaningful signals related to inflation. 

\begin{figure}[t]
    \centering
    \includegraphics[width=1\linewidth, height=0.4\linewidth]{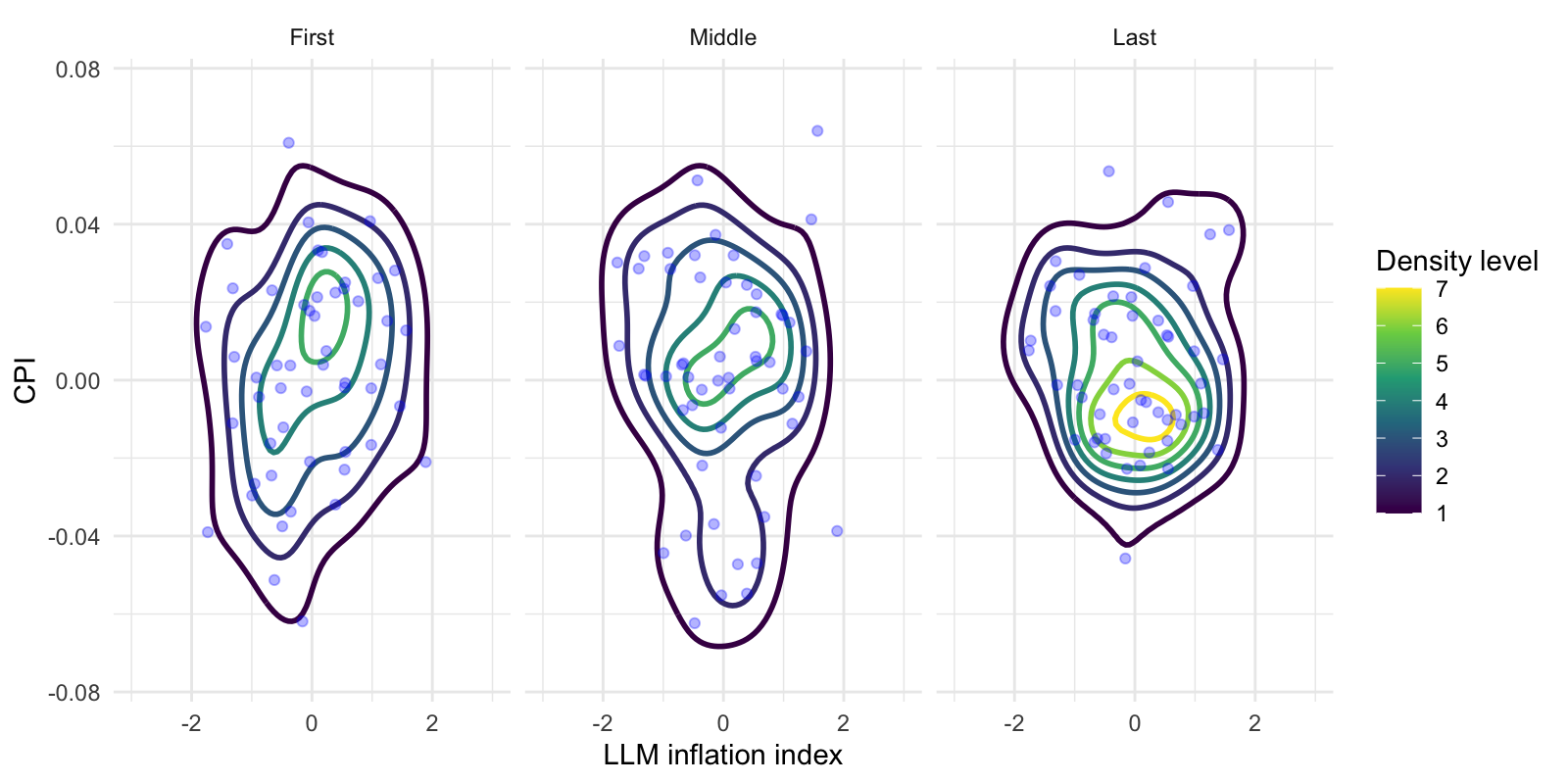}
    \caption{Density plots of residual vectors from fitting the LLM-CPI model, integrating the target CPI model \eqref{equ:cpi} and the surrogate model \eqref{equ:sur} on the LLM-generated daily inflation index, across three time periods: 1) first (initial $10$ days of the month), 2) middle (middle $10$ days of the month), and 3) last (remaining days of the month). The approximately elliptical contours indicate the normality with correlations; see Section \ref{sec:real_data} for model fitting details.
    }
    \label{fig:error_density}
\end{figure}

We are now ready to introduce our joint LLM-CPI model that links both the target CPI model (\ref{equ:cpi}) and the LLM-powered surrogate model (\ref{equ:sur}). Specifically, to bridge the target CPI model and the LLM-powered surrogate model, we assume that the errors of both models jointly follow a multivariate normal distribution as in \cite{mccaw2024synthetic}
\begin{equation}\label{equ:joint_model}
    (\epsilon_t, (\beps_t^S)^\top)^\top \sim N(\mathbf{0}^\top, \bSigma), 
\end{equation}
where the error covariance matrix is given by 
$$
\bSigma = \begin{bmatrix}
    \sigma^2_{TT} &\bSigma_{TS}\\
    \bSigma_{ST} &\bSigma_{SS}
\end{bmatrix}.
$$
We further assume that the errors are independent across different months, meaning that the autoregressive terms in the models account for all cross-temporal correlations across months. Here, $\sigma^2_{TT}$ represents the variance of the target CPI model errors, $\bSigma_{SS}$ denotes the covariance matrix of the LLM-powered surrogate model errors, and $\bSigma_{TS}$ captures the cross-covariance between the target CPI model and the LLM-powered surrogate model errors. Such formulation allows us to model the correlation structures between the target CPI and LLM-powered  surrogate predictions. 
The joint normality assumption in the LLM-CPI model (\ref{equ:joint_model}) could be validated in our real data set, as depicted in Figure \ref{fig:error_density}. In particular, we fit the target CPI model \eqref{equ:cpi} and the surrogate model \eqref{equ:sur} on our online text data with CPI , and calculate the resulting residuals. We then plot the 
density for the estimated distribution of residuals $(\widehat{\epsilon}_t, \widehat{\epsilon}^S_{t,k})$ for each $1 \leq k \leq 3$. The approximately elliptical contours of these three density plots indicate the normality with correlations, demonstrating the practical utility of the LLM-CPI model (\ref{equ:joint_model}). 


\subsection{Theoretical justifications of LLM-CPI} \label{Sec.asymtheo}

It is important to emphasize that the target CPI model and the LLM-powered surrogate model do \textit{not} share any model parameters; their only connection is through the correlations of their model errors. Our LLM-CPI method only leverages the correlation structure between both model errors to enhance the prediction and inference accuracy. In view of the joint LLM-CPI model (\ref{equ:joint_model}), we can rewrite the random error of the target CPI model \eqref{equ:cpi} using the conditional normal distribution. Specifically, the error term $\epsilon_t$ of the target CPI model \eqref{equ:cpi} admits the representation 
\begin{equation}\label{Feb22:equ01}
 \epsilon_t = \bSigma_{TS}\bSigma_{SS}^{-1}\beps^S_t + e_t := \bgamma^\top \beps^S_t + e_t,    
\end{equation}
where $\bgamma = \bSigma_{SS}^{-1} \bSigma_{ST}$ represents the regression coefficient vector linking the LLM-generated surrogate model errors to the target CPI model errors, and random error $e_t \sim N(0, \sigma^2_{e})$ with $\sigma^2_{e}= \sigma_{TT} - \bSigma_{TS}\bSigma_{SS}^{-1}\bSigma_{ST}$. The decomposition in (\ref{Feb22:equ01}) above allows us to isolate the portion of variation in the target CPI model errors that can be explained by the LLM-powered surrogate model errors, thereby reducing the overall variance of the target CPI model errors and ensuring more accurate prediction and inference. To further leverage the LLM-powered surrogate model \eqref{equ:sur}, we can write
\begin{equation}\label{Feb22:equ02}
 \beps^S_t = \by^S_t - \sum_{l=1}^{q_2} \bA^S_{l}  \by^S_{t-l} - \bB^S \bx_t.
\end{equation}

By substituting expression (\ref{Feb22:equ02}) into decomposition \eqref{Feb22:equ01}, it holds that 
\begin{equation}\label{Feb22:equ03}
    \epsilon_t =  \bgamma^\top \left( \by^S_t - \sum_{l=1}^{q_2} \bA^S_{l}  \by^S_{t-l} \right) - \bgamma^\top \bB^S \bx_t + e_t:= \bgamma^\top \bD (\by^S_t)- \bgamma^\top \bB^S \bx_t + e_t,
\end{equation}
where $\bD (\by^S_t) := \by^S_t - \sum_{l=1}^{q_2} \bA^S_{l}  \by^S_{t-l}$ captures the 
error component of the LLM predictions after accounting for their autoregressive structure. Plugging expression (\ref{Feb22:equ03}) into the target CPI model \eqref{equ:cpi}, we can rewrite the joint LLM-CPI model (\ref{equ:joint_model}) as a joint LLM-powered ARX model (i.e., an equivalent representation) given by 
\begin{equation}\label{equ:trans}
\begin{aligned}
    y_{t} =&  \sum_{l=1}^{q_1} \alpha_{l} y_{t-l} +\bz_t^\top \btheta + \bx_t^\top \left(\bbeta -\bgamma^\top \bB^S \right)+ \bgamma^\top \bD(\by^S_t)+ e_t\\
    =&\sum_{l=1}^{q_1} \alpha_{l} y_{t-l} +\bz_t^\top \btheta + \bx_t^\top \bdelta + \bgamma^\top \bD(\by^S_t)+ e_t,
\end{aligned}
\end{equation}
where $\bdelta = \bbeta -\bgamma^\top \bB^S$. Such joint model (\ref{equ:trans}) benefits from a reduced error variance due to the law of total variance, with the extent of improvement directly tied to the strength of the correlations between the target CPI and the LLM-generated surrogate predictions. The stronger their correlation, the greater gains in the prediction and inference accuracy. Throughout the rest of the paper, the joint LLM-CPI model is implicitly referred to as both versions (\ref{equ:joint_model}) and (\ref{equ:trans}).

To estimate the parameters of the joint LLM-powered ARX model \eqref{equ:trans}, we exploit a two-step approach. We first estimate parameters of the LLM-powered surrogate model \eqref{equ:sur} by solving the optimization problem 
\begin{equation}\label{equ:obj}
    (\widehat{\bA}^S_1, \ldots, \widehat{\bA}^S_{q_2}, \widehat{\bB}^S) = \arg\min_{\bA^S_1, \ldots, \bA^S_{q_2}, \bB^S} \frac{1}{T} \sum_{t=q_2+1}^{T}\Big\| \by^S_t - \sum_{l=1}^{q_2} \bA^S_{l}  \by^S_{t-l} - \bB^S \bx_t\Big\|^2.
\end{equation}
Given the estimated parameters of the LLM-generated surrogate model above, we compute the residual component of the LLM predictions after accounting for their autoregressive structure (i.e., the empirical version of $\bD(\by_t^{S})$) given by 
\begin{equation}\label{equ:hD}
    \widehat{\bD}(\by_t^{S}) = \by^S_t - \sum_{l=1}^{q_2} \widehat{\bA}^S_{l}  \by^S_{t-l}.
\end{equation}
We then estimate the parameters of the joint LLM-powered ARX model \eqref{equ:trans} by solving the optimization problem 
\begin{equation}\label{equ:obj_cpi}
\begin{aligned}
(\hat{\balpha}, \hat{\btheta}, \hat{\bdelta}, \hat{\bgamma}) =& \arg\min_{\balpha, \btheta, \bdelta, \bgamma} \frac{1}{T} \sum_{t=q_q+1}^{T} \Big(y_{t}  -  \sum_{l=1}^{q_1} \alpha_{l} y_{t-l} -\bz_t^\top \btheta - \bx_t^\top \bdelta - \bgamma^\top \widehat{\bD}(\by^S_t)\Big)^2.
\end{aligned}    
\end{equation}

With the estimates given in (\ref{equ:obj})--(\ref{equ:obj_cpi}) above, we can construct the one-step-ahead forecast $\hat{y}_{T+1}$ using the joint LLM-powered ARX model \eqref{equ:trans} as 
\begin{equation}\label{equ:prediction}
    \hat{y}_{T+1} = \sum_{l=1}^{q_1} \hat{\alpha}_{l} y_{T+1-l} +\bz_{T+1}^\top \hat{\btheta} + \bx_{T+1}^\top \hat{\bdelta} + \hat{\bgamma}^\top \widehat{\bD}(\by^S_{T+1}),
\end{equation}
where $\widehat{\bD}(\by^S_{T+1})$ is calculated using \eqref{equ:hD} with $t= T+1$. For the multi-step-ahead forecasts, we employ a rolling horizon approach. Specifically, the $h$-step-ahead forecast $\hat{y}_{T+h}$ can be constructed as 
\begin{equation}\label{equ:hstep}
    \hat{y}_{T+h} = \sum_{l=1}^{q_1} \hat{\alpha}_{l} \hat{y}_{T+h-l} +\bz_{T+h}^\top \hat{\btheta} + \bx_{T+h}^\top \hat{\bdelta} + \hat{\bgamma}^\top \widehat{\bD}(\by^S_{T+h}),
\end{equation}
where $\hat{y}_{T+h-1}, \hat{y}_{T+h-2}, \ldots$ are iteratively computed based on the forecasts $\hat{y}_{t}$ from previous time stamps with $\hat{y}_{t}= y_t$ for each $t \le T$, and covariates  $\bx_{T+h}$, $\bz_{T+h}$, and $\widehat{\bD}(\by^S_{T+h})$ are used as the inputs.  

We say that an estimator $\hat{\bB} \in \R^{p_1 \times p_2}$ of parameter $\bB \in \R^{p_1 \times p_2}$ is $\zeta_T$-consistent if $\|\hat{\bB} - \bB\|_F = O_p(\zeta^{-1}_T)$, where $\zeta_T$ denotes the convergence rate with $\zeta_T \to \infty$ as $T \to \infty$. Typically for parametric estimation, we have $\zeta_T = \sqrt{T}$, although this may vary depending on the model setting. The theorem below characterizes the asymptotic property of the LLM-CPI method.

\begin{theorem} \label{lem:prediction}
Assume that the target CPI model (\ref{equ:cpi}) and LLM-powered surrogate model (\ref{equ:sur}) integrated in the LLM-CPI model (\ref{equ:joint_model}) or (\ref{equ:trans}) are both for stationary processes (which ensures that $y_t = O_p(1)$ and $\| \by_t^S\| = O_p(1)$ for each $t =1,\ldots, T$), and estimation procedures \eqref{equ:obj} and \eqref{equ:obj_cpi} are $\zeta_T$-consistent. 
 Then for each 
 fixed $h \geq 1$, we have
    $$
    \hat{y}_{T+h} - y_{T+h} = \sum_{r=0}^{h-1}(\bA^r)_{11} e_{T+h-r} +  O_p(\zeta_T^{-1}),
    $$
    where matrix $\bA$ is as defined in \eqref{def:A} and $\bA^0$ is defined as the identity matrix $\mathbf{I}$.
\end{theorem}

Theorem \ref{lem:prediction} above primarily addresses the large-sample properties, demonstrating that the LLM-CPI method is asymptotically unbiased when the sample size $T$ is sufficiently large. The regularity conditions required by this theorem are commonly imposed in the literature. In particular, the stationarity assumption can be satisfied by any weakly stationary processes;  see, e.g., the discussions in Section 2 of \cite{lutkepohl2013introduction}. Further, the least squares estimation can be applied to solve optimization problems \eqref{equ:obj} and \eqref{equ:obj_cpi}, where the corresponding consistency rates under different types of conditions can be found in Section 3.3 of \cite{lutkepohl2013introduction} and \cite{lai1982least}, as well as the references therein.
\begin{remark}\label{remark1}
There is also a previous literature exploring the small-sample properties of autoregressive models \citep{phillips1979sampling, series1981properties}, highlighting that predictions from such models are often biased and require corrections when sample size $T$ is relatively small. We observe from our numerical analysis that the bias is also affected by the variance of the errors, and a major advantage of the LLM-powered prediction inference framework is to reduce such error variance. By doing so, our approach has the potential to mitigate the prediction bias. However, to maintain focus on the primary contributions of this paper, we leave a detailed exploration of this aspect for future work.
\end{remark}

In contrast, if we ignore the effect of the LLM-generated surrogate model, and rely solely on the traditional \textit{non-LLM-powered} ARX model, the $h$-step-ahead forecast will be given by
\begin{equation}\label{equ:arx_prediction}
\hat{y}^{a}_{T+h} =  \sum_{l=1}^{q_1} \hat{\alpha}_{l} \hat{y}^{a}_{T+h-l} + \bz_{T+h}^\top \hat{\btheta} + \bx_{T+h}^\top \hat{\bbeta}, 
\end{equation}
where $\hat{y}^a_{T+h-1}, \hat{y}^a_{T+h-2}, \ldots$ are iteratively computed via the ARX model prediction with $\hat{y}^a_t = y_{t}$ for each $t \le T$. The prediction error 
for this benchmark model is 
$$
\hat{y}^a_{T+h} - y^a_{T+h} = \sum_{r=0}^{h-1}(\bA^r)_{11} \epsilon_{T+h-r} +  O_p(\zeta_T^{-1}).
$$
Since $O_p(\zeta_T^{-1})$ above is negligible, the efficiency gain of the LLM-powered prediction inference method, compared to the traditional ARX model without the use of the LLM-generated surrogates, can be quantified by the ratio of prediction error variances
\begin{equation} \label{new.eq.effigain}
\operatorname{Efficiency} = \frac{\Var\left(\sum_{r=0}^{h-1}(\bA^r)_{11} \epsilon_{T+h-r}\right)}{\Var\left(\sum_{r=0}^{h-1}(\bA^r)_{11} e_{T+h-r}\right)} =  \frac{\sigma^2_{TT}}{\sigma^2_{TT} - \bSigma_{TS}\bSigma_{SS}^{-1}\bSigma_{ST}}.
\end{equation}

Under the simplified assumptions of $\bSigma_{SS} = \mathbf{I}$ and $\bSigma_{ST} = \rho \mathbf{1}$ with $\mathbf{I}$ and $\mathbf{1}$ the identity matrix and the vector of ones, respectively, the efficiency gain in (\ref{new.eq.effigain}) of the LLM-CPI relative to the traditional ARX benchmark above reduces to 
\begin{equation} \label{new.eq.effigain.simp}
\operatorname{Efficiency} = \frac{1}{1-|S| \rho^2}.
\end{equation}
Here, we require $|S| \rho^2 < 1$ to guarantee the positive definiteness of the joint covariance matrix $\bSigma$. In light of (\ref{new.eq.effigain.simp}), we see the practical benefits of incorporating the LLM-generated synthetic surrogates in the LLM-CPI, and that the stronger the correlations between the target CPI and these LLM-generated surrogates, the greater the efficiency gain. 
This highlights the significant improvement in prediction and inference accuracy achieved by the LLM-CPI, as unveiled in the simulation and real data results in Sections \ref{sec:simu} and \ref{sec:real_data}, respectively. 

\subsection{CPI prediction inference via LLM-CPI} \label{Sec.CPIpredinf}

We now introduce two ways of constructing the LLM-CPI prediction intervals for CPI prediction inference.

\subsubsection{Box--Jenkins prediction interval} \label{Sec.CPIpredinf.bj}

Based on the results in Theorem \ref{lem:prediction}, we can construct an asymptotic prediction interval for the $h$-step-ahead prediction $\hat{y}_{T+h}$ once we obtain a consistent estimator of the error variance. One common approach to estimating such variance is through the sum of squared residuals \citep{lai1982least}
$$
\hat{\sigma}^2_e = \frac{1}{T-q_1}\sum_{t=q_1+1}^T \left( y_t - \sum_{l=1}^{q_1} \hat{\alpha}_{l} y_{t+1-l} -\bz_t^\top \hat{\btheta} - \bx_t^\top \hat{\bdelta} - \hat{\bgamma}^\top \widehat{\bD}(\by^S_{t}) \right)^2 := \frac{1}{T-q_1}  \sum_{t=q_1+1}^T  \hat{e}_t^2.
$$
Using the above error variance estimator, we can construct the Box--Jenkins (BJ) prediction interval \citep{box2015time} with confidence level $1 - \alpha$ as 
\begin{equation}\label{equ:pre_interval}
   \operatorname{PI}^{BJ}(\hat{y}_{T+h})= \left[\hat{y}_{T+h} - |z_{\alpha/2}| \sqrt{\sum_{r=0}^{h-1}(\hbA^r)^2_{11}}\hat{\sigma}_e,  \, \hat{y}_{T+h} + |z_{\alpha/2}|\sqrt{ \sum_{r=0}^{h-1}(\hbA^r)^2_{11}} \hat{\sigma}_e\right],
\end{equation}
where $z_{\alpha/2}$ is the $\alpha/2$ quantile of the standard normal distribution and $\alpha \in (0, 1)$. The BJ prediction interval asymptotically covers the true value $y_{T+h}$ if $\hat{\sigma}^2_e$ is a consistent estimator of $\sigma^2_e$. The theorem below verifies such consistency under certain regularity conditions. 
Let us define $\lambda_{\max,z} = \lambda_{\max}\left(\sum_{t=q_1+1}^T \bz_t \bz_t^\top/T\right)$, $\lambda_{\max,x} = \lambda_{\max}\left(\sum_{t=q_1+1}^T \bx_t \bx_t^\top/T\right)$, and 
$\lambda_{\max,D} = \lambda_{\max}\left(\sum_{t=q_1+1}^T \bD(\by_t^S) \bD(\by_t^S)^\top/T\right)$ with $\lambda_{\max}(\cdot)$ representing the maximum eigenvalue of a given symmetric matrix.

\begin{theorem} \label{lem:variance}
Assume that all the conditions of Theorem \ref{lem:prediction} and the additional conditions 
   \begin{align}
        &\max_{t=1,\ldots,T} \E(y^4_{t})=  O(1), \ \max_{t=1,\ldots,T}  \E\left(\| \by_{t}^S\|^4 \right)= O\left(1\right), \label{con:lem2:1}\\
       &\lambda_{\max,z} = o_p(\zeta^{2}_T), \  \lambda_{\max,x} = o_p(\zeta^2_T), \text{ and } \lambda_{\max,D} = o_p(\zeta^2_T)\label{con:lem2:2}
   \end{align}
   are satisfied. Then we have $\hat{\sigma}^2_e \overset{p}{\to} \sigma^2_e$, and  for each fixed $h \geq 1$, the BJ prediction interval in (\ref{equ:pre_interval}) satisfies that 
   $$
   \liminf_{T \to \infty}\Pr\left\{ y_{T+h} \in \operatorname{PI}^{BJ}(\hat{y}_{T+h}) \right\} \ge 1-\alpha.
   $$
\end{theorem}


Theorem \ref{lem:variance} above justifies the asymptotic validity of the LLM-CPI method with the BJ prediction interval. Condition (\ref{con:lem2:1}) requires that the fourth moments exist, and Condition (\ref{con:lem2:2}) demands upper bounds on the largest eigenvalues of design matrices, which is standard one in the literature; see, e.g., \cite{lai1982least} and \cite{medeiros2016l1}.

\subsubsection{Bootstrap prediction interval} \label{Sec.CPIpredinf.boot}

We also suggest a residual-based bootstrap prediction interval for the LLM-CPI \citep{bickel1981some,freedman1981bootstrapping}.
Given the estimated parameters $(\hat{\balpha}, \hat{\btheta}, \hat{\bdelta}, \hat{\bgamma})$, we compute the residuals
\begin{equation}\label{equ:residual}
\hat{e}_t = y_{t}  - \sum_{l=1}^{q_1} \hat{\alpha}_{l} y_{t-l} -\bz_t^\top \hat{\btheta} - \bx_t^\top \hat{\bdelta} - \hat{\bgamma}^\top \widehat{\bD}(\by^S_t), \ t =q_1+1,\ldots, T.
\end{equation}
We can generate the bootstrap residuals by resampling $T+h$ bootstrap samples from $\{\hat{e}_t - \hat{\mu}, \, t = q_1+1,\ldots, T\}$ with replacement, where $\hat{\mu} = \sum_{t=q_1+1}^T \hat{e}_t/(T-q_1)$. Denote by $\{e^{*}_t, \, t = 1,\ldots, T+h\}$ the bootstrap residuals. By recursive calculations, it holds that 
\begin{equation}\label{equ:boot_y}
    {y}^{*}_{t}  = \sum_{l=1}^{q_1} \hat{\alpha}_{l} {y}^*_{t-l} +\bz_{t}^\top \hat{\btheta} + \bx_{t}^\top \hat{\bdelta} + \hat{\bgamma}^\top \widehat{\bD}(\by^S_{t})+{e}_{t}^*, \  t=q_1+1,\ldots,T+h
\end{equation}
with initial points $\{ {y}^{*}_{t} = e_t^*, \, t \le q_1\}$. We then refit the joint LLM-powered ARX model \eqref{equ:trans} on the bootstrap sample $\{\hat{y}^{*}_{t}, \, t=q_1+1,\ldots,T\}$ and denote the refitted parameters as $\{\hat{\balpha}^*, \hat{\btheta}^*, \hat{\bdelta}^*, \hat{\bgamma}^*\}$. The $h$-step-ahead forecast for the bootstrap sample is given by
\begin{equation}\label{equ:boot_hstep}
    \hat{y}^*_{T+h} = \sum_{l=1}^{q_1} \hat{\alpha}^*_{l} \hat{y}^*_{T+h-l} +\bz_{T+h}^\top \hat{\btheta}^* + \bx_{T+h}^\top \hat{\bdelta}^* + (\hat{\bgamma}^*)^\top \widehat{\bD}(\by^S_{T+h}),
\end{equation}
where $\hat{y}_{T+h-1}^*, \hat{y}^*_{T+h-2}, \ldots$ are iteratively computed with $\hat{y}_{t}^*$ understood as ${y}^*_{t}$ for each $t \le T$. Using $\hat{y}^*_{T+h}$ introduced above, the bootstrap residual is calculated as  $\hat{e}^*_{T+h} = {y}^*_{T+h} - \hat{y}^*_{T+h}$. 

We repeat the bootstrap procedure (i.e., \eqref{equ:boot_y} and \eqref{equ:boot_hstep})  $B \geq 1$ times to obtain a sequence of bootstrap residuals $\{\hat{e}^{*,(b)}_{T+h}, \, b = 1,\ldots, B\}$. For each $\alpha \in (0, 1)$, denote the $\alpha/2$ quantile and $1-\alpha/2$ quantile of the bootstrap residuals as $\hat{q}^h_{\alpha/2}$ and $\hat{q}^h_{1-\alpha/2}$, respectively. Then we can construct the bootstrap prediction interval with confidence level $1 - \alpha$ as 
\begin{equation}\label{equ:bo_interval}
   \operatorname{PI}^{BOOT}(\hat{y}_{T+h})= \left[\hat{y}_{T+h} + \hat{q}^h_{\alpha/2}, \,  \hat{y}_{T+h} + \hat{q}^h_{1-\alpha/2}\right].
\end{equation}
Let $\mathbf{g}_t := (y^*_{t-1}, \ldots, y^*_{t-q_1}, \bz_t^\top, \bx_t^\top, (\bD(\by_t^S))^\top )^\top \in \R^{q_1 + b + p + K}$ be the joint feature vector, and $\mathbf{G} := (\mathbf{g}_{q_1+1},\ldots, \mathbf{g}_{T})^\top  \in \R^{(T-q_1) \times (q_1 + b + p + K)}$ the bootstrap design matrix. Define  $\lambda_{\max, G} = \lambda_{\max}(\mathbf{G}^\top \mathbf{G})$ and $\lambda_{\min, G} = \lambda_{\min}(\mathbf{G}^\top \mathbf{G})$, where $\lambda_{\max}(\cdot)$ and $\lambda_{\min}(\cdot)$ represent the maximum and minimum eigenvalues of a given symmetric matrix, respectively. We specify the parameter estimators as the least squares estimators as in \cite{lai1982least}.

\begin{theorem} \label{lem:variance.boot}
 Assume that all the conditions of Theorem \ref{lem:variance} are satisfied, the bootstrap time series given in \eqref{equ:boot_y} is stationary, and 
 \begin{equation}\label{con:lem3:1}
     \lambda_{\min, G}/T \gg  \zeta_T^{-1} \text{ and }  \  \lambda_{\min, G}^{-1} \log(\lambda_{\max, G}) \to 0.
 \end{equation}
Then for each fixed $h \geq 1$, the bootstrap prediction interval in (\ref{equ:bo_interval}) satisfies that 
   $$
   \liminf_{T,B \to \infty}\Pr\left\{ y_{T+h} \in \operatorname{PI}^{BOOT}(\hat{y}_{T+h}) \right\} \ge 1-\alpha.
   $$
\end{theorem}

Theorem \ref{lem:variance.boot} above establishes the asymptotic coverage of the LLM-CPI method with the bootstrap prediction interval. Intuitively, under the above regularity conditions, the empirical distribution of $\{\hat{e}^{*,(b)}_{T+h}, \, b = 1, \ldots, B\}$ converges to the distribution of $y_{T+h} - \hat{y}_{T+h}$ asymptotically. Consequently, the quantiles of $\{\hat{e}^{*,(b)}_{T+h}, \, b = 1, \ldots, B\}$ can be used to approximate those of distribution of $y_{T+h} - \hat{y}_{T+h}$. Therefore, the bootstrap prediction interval provides asymptotically valid coverage for the true value.


\section{Simulation examples} \label{sec:simu}

We provide in this section several simulation examples to investigate the finite-sample performance of the LLM-CPI method, in which the generated synthetic data sets are real data based.

\subsection{Simulation settings} \label{Sec.simusett}

To closely mimic the underlying structure of the real data set, we select two LDA embedding features as the predictors, denoted as $\bx_t\in\R^2$, as discussed in Section \ref{sec:real_data.descripreproc}. See Section \ref{app:sec:embeddings} for details on how these LDA embeddings are constructed. We then generate synthetic observations following the data-generating process given in the LLM-CPI model integrating the target CPI model \eqref{equ:cpi} and the LLM-powered surrogate model \eqref{equ:sur}. The simulation studies examine the performance of the LLM-CPI model in comparison to several popular benchmark models, under varying correlation levels between the error terms of the target and surrogate models. We specify the model settings as follows:
\begin{itemize}
    \item \textit{Target CPI model  \eqref{equ:cpi}}.  We define the autoregressive coefficient vector $\balpha = (0.5, -0.3)^\top$ and exogenous coefficient vector $\bbeta = (0.7, -0.2)^\top$. Since $\bz_t$ and $\bx_t$ are both treated as exogenous variables with the same status, we set $\btheta = \mathbf{0}$ in the simulation studies. Then the autoregressive order of the target model is $q_1  = 2$.

    \item \textit{LLM-powered surrogate model \eqref{equ:sur}}. We define the autoregressive coefficient matrix and the exogenous coefficient matrix as
$$
\bA_1^S = \begin{bmatrix}
    0.2, &0.2, &0.2\\
    -0.2, &-0.2, &-0.2\\
    -0.1, &-0.1, &-0.1
\end{bmatrix}, \quad  
\bB^S =  \begin{bmatrix}
    0.1, &0.1\\
    -0.1, &-0.1\\
    -0.3, &-0.3
\end{bmatrix},
$$
respectively. The autoregressive order of the surrogate model is set to $q_2=1$.

\item \textit{Error structure}. The errors for the target CPI model and the LLM-powered surrogate model are generated from the multivariate normal distribution $(\epsilon_t, \beps^S_t)^\top \sim N(\mathbf{0}, \bSigma)$, where covariance matrix $\bSigma$ of errors consists of equal values $\rho$ except that the diagonal entries are all equal to $1$.
\end{itemize}

For each model setting, we generate synthetic data from the LLM-CPI model linking both the target CPI and LLM-powered surrogate models using the specified model parameters. We then estimate the parameters of 
the joint LLM-CPI model with the two-step procedure suggested in Section \ref{Sec.asymtheo}. To evaluate the performance, we compare the LLM-CPI model to the AR model \eqref{equ:pure_AR}, random walk (RW) model \eqref{equ:random_walk} \citep{atkeson2001phillips}, and the historical average (AVE) model \eqref{equ:mean_model} which is frequently used in financial market forecasting \citep{welch2008comprehensive}\footnote{The details of these forecasting models can be found in Section \ref{new.Sec.forecastmods}.}.

\begin{table}[t]
\centering
\small
\caption{The $\operatorname{rPMSE}^{AR}_m(H)$ results across different prediction steps $H$ and correlation levels $\rho$.
}
\label{tab:simu_true_mse}
\begin{tabular}{cccccccccc}
\toprule
$H$ & 8 & 9 & 10 & 11 & 12 & 13 & 14 & 15 & Ave. \\
\midrule
\multicolumn{10}{c}{$\rho = 0.1$} \\
\midrule
RW       & 0.768 & 1.174 & 0.802 & 0.857 & 2.073 & 3.556 & 14.661 & 19.358 & 5.906 \\
AVE      & 0.686 & 1.102 & 2.327 & 4.933 & 7.849 & 9.188 & 10.753 & 8.133 & 5.371 \\
LLM-CPI  & 0.204 & 0.209 & 0.250 & 0.317 & 0.334 & 0.301 & 0.325 & 0.408 & 0.294 \\
\midrule
\multicolumn{10}{c}{$\rho = 0.2$} \\
\midrule
RW       & 0.761 & 1.164 & 0.798 & 0.871 & 2.152 & 3.638 & 15.100 & 19.845 & 6.291 \\
AVE      & 0.677 & 1.130 & 2.397 & 5.084 & 8.082 & 9.416 & 10.999 & 8.199 & 5.748 \\
LLM-CPI  & 0.205 & 0.206 & 0.247 & 0.311 & 0.329 & 0.297 & 0.319 & 0.403 & 0.290 \\
\midrule
\multicolumn{10}{c}{$\rho = 0.3$} \\
\midrule
RW       & 0.761 & 1.171 & 0.795 & 0.833 & 2.096 & 3.721 & 15.065 & 19.784 & 6.279 \\
AVE      & 0.675 & 1.115 & 2.366 & 5.010 & 8.044 & 9.488 & 10.987 & 8.218 & 5.738 \\
LLM-CPI  & 0.196 & 0.200 & 0.242 & 0.309 & 0.325 & 0.294 & 0.314 & 0.393 & 0.284 \\
\midrule
\multicolumn{10}{c}{$\rho = 0.4$} \\
\midrule
RW       & 0.759 & 1.173 & 0.796 & 0.855 & 2.082 & 3.606 & 15.014 & 19.751 & 6.254 \\
AVE      & 0.676 & 1.120 & 2.380 & 5.058 & 8.023 & 9.376 & 10.989 & 8.217 & 5.730 \\
LLM-CPI  & 0.185 & 0.190 & 0.236 & 0.302 & 0.314 & 0.282 & 0.305 & 0.381 & 0.274 \\
\bottomrule
\end{tabular}\\
{\footnotesize Note: The relative PMSE values compared to the AR benchmark. Smaller values indicate better performance.}
\end{table}

\begin{table}[t]
\centering
\small
\caption{The $\operatorname{rSign}^{AR}_m(H)$ results across different prediction steps $H$ and correlation levels $\rho$.}
\label{tab:simu_true_sign}
\begin{tabular}{cccccccccc}
\toprule
$H$ & 8 & 9 & 10 & 11 & 12 & 13 & 14 & 15 & Ave. \\
\midrule
\multicolumn{10}{c}{$\rho = 0.1$} \\
\midrule
RW       & 0.637 & 0.932 & 0.613 & 0.591 & 0.675 & 0.545 & 0.688 & 0.695 & 0.672 \\
AVE      & 0.631 & 0.932 & 0.517 & 0.588 & 0.675 & 0.545 & 0.688 & 0.695 & 0.659 \\
LLM-CPI  & 0.359 & 0.599 & 0.279 & 0.410 & 0.471 & 0.361 & 0.458 & 0.439 & 0.422 \\
\midrule
\multicolumn{10}{c}{$\rho = 0.2$} \\
\midrule
RW       & 0.638 & 0.937 & 0.597 & 0.583 & 0.681 & 0.543 & 0.700 & 0.707 & 0.673 \\
AVE      & 0.630 & 0.937 & 0.513 & 0.583 & 0.681 & 0.543 & 0.700 & 0.707 & 0.662 \\
LLM-CPI  & 0.362 & 0.612 & 0.279 & 0.398 & 0.467 & 0.359 & 0.471 & 0.445 & 0.424 \\
\midrule
\multicolumn{10}{c}{$\rho = 0.3$} \\
\midrule
RW       & 0.630 & 0.930 & 0.615 & 0.581 & 0.671 & 0.541 & 0.698 & 0.711 & 0.672 \\
AVE      & 0.625 & 0.930 & 0.511 & 0.581 & 0.671 & 0.541 & 0.698 & 0.711 & 0.658 \\
LLM-CPI  & 0.331 & 0.573 & 0.264 & 0.399 & 0.465 & 0.361 & 0.467 & 0.445 & 0.413 \\
\midrule
\multicolumn{10}{c}{$\rho = 0.4$} \\
\midrule
RW       & 0.609 & 0.930 & 0.604 & 0.587 & 0.663 & 0.532 & 0.682 & 0.688 & 0.662 \\
AVE      & 0.605 & 0.930 & 0.514 & 0.587 & 0.663 & 0.532 & 0.682 & 0.688 & 0.638 \\
LLM-CPI  & 0.298 & 0.569 & 0.264 & 0.394 & 0.442 & 0.338 & 0.438 & 0.413 & 0.394 \\
\bottomrule
\end{tabular}\\
{\footnotesize Note: The relative sign prediction error values compared to the AR benchmark. Smaller values indicate better performance.}
\end{table}

\begin{table}[t]
\centering
\small
\caption{The $\operatorname{Coverage}_m(H)$ and $\operatorname{Length}_m(H)$ results across different prediction steps $H$ and correlation levels $\rho$.}
\label{tab:simu_true}
{\small
\begin{tabular}{cccccccccc}
\toprule
Method & 8 & 9 & 10 & 11 & 12 & 13 & 14 & 15 & Ave. \\
\midrule
\multicolumn{10}{c}{$\rho = 0.1$} \\
\midrule
AR          & 1.000 & 1.000 & 1.000 & 1.000 & 1.000 & 1.000 & 1.000 & 1.000 & 1.000 \\
            & (3.623) & (3.719) & (3.762) & (3.830) & (3.898) & (3.951) & (4.049) & (4.126) & (3.870) \\
BJ      & 0.945 & 0.934 & 0.934 & 0.914 & 0.910 & 0.925 & 0.933 & 0.902 & 0.924 \\
            & (0.802) & (0.804) & (0.796) & (0.790) & (0.792) & (0.797) & (0.800) & (0.799) & (0.798) \\
BOOT    & 0.938 & 0.930 & 0.936 & 0.928 & 0.927 & 0.942 & 0.916 & 0.883 & 0.925 \\
            & (1.044) & (1.048) & (1.052) & (1.065) & (1.084) & (1.089) & (1.087) & (1.083) & (1.069) \\
\midrule
\multicolumn{10}{c}{$\rho = 0.2$} \\
\midrule
AR          & 1.000 & 1.000 & 1.000 & 1.000 & 1.000 & 1.000 & 1.000 & 1.000 & 1.000 \\
            & (3.623) & (3.720) & (3.763) & (3.831) & (3.899) & (3.952) & (4.049) & (4.125) & (3.870) \\
BJ      & 0.949 & 0.934 & 0.934 & 0.911 & 0.912 & 0.926 & 0.936 & 0.907 & 0.926 \\
            & (0.789) & (0.792) & (0.783) & (0.777) & (0.780) & (0.785) & (0.789) & (0.788) & (0.785) \\
BOOT    & 0.939 & 0.934 & 0.940 & 0.927 & 0.930 & 0.944 & 0.917 & 0.888 & 0.927 \\
            & (1.035) & (1.042) & (1.045) & (1.059) & (1.077) & (1.085) & (1.081) & (1.081) & (1.063) \\
\midrule
\multicolumn{10}{c}{$\rho = 0.3$} \\
\midrule
AR          & 1.000 & 1.000 & 1.000 & 1.000 & 1.000 & 1.000 & 1.000 & 1.000 & 1.000 \\
            & (3.625) & (3.723) & (3.764) & (3.832) & (3.900) & (3.954) & (4.051) & (4.129) & (3.872) \\
BJ      & 0.950 & 0.936 & 0.937 & 0.909 & 0.910 & 0.926 & 0.933 & 0.906 & 0.927 \\
            & (0.782) & (0.784) & (0.775) & (0.768) & (0.770) & (0.774) & (0.777) & (0.776) & (0.776) \\
BOOT    & 0.943 & 0.938 & 0.937 & 0.928 & 0.924 & 0.943 & 0.915 & 0.886 & 0.927 \\
            & (1.028) & (1.035) & (1.038) & (1.052) & (1.069) & (1.074) & (1.073) & (1.074) & (1.055) \\
\midrule
\multicolumn{10}{c}{$\rho = 0.4$} \\
\midrule
AR          & 1.000 & 1.000 & 1.000 & 1.000 & 1.000 & 1.000 & 1.000 & 1.000 & 1.000 \\
            & (3.628) & (3.725) & (3.767) & (3.835) & (3.903) & (3.956) & (4.053) & (4.130) & (3.875) \\
BJ      & 0.953 & 0.937 & 0.934 & 0.909 & 0.911 & 0.928 & 0.937 & 0.909 & 0.927 \\
            & (0.770) & (0.774) & (0.765) & (0.759) & (0.761) & (0.767) & (0.770) & (0.770) & (0.767) \\
BOOT    & 0.938 & 0.932 & 0.937 & 0.925 & 0.922 & 0.939 & 0.913 & 0.880 & 0.923 \\
            & (1.021) & (1.027) & (1.031) & (1.042) & (1.061) & (1.067) & (1.066) & (1.063) & (1.047) \\
\bottomrule
\end{tabular}}\\
{\footnotesize Note: The values in the parentheses are interval length, and coverage near the nominal level of $0.95$ with smaller interval length is preferred.}
\end{table}

Given a prediction horizon of $H$ months, we train each forecasting model using a training sample that spans from January 2019 to December 2023 (matching the real data set in Section \ref{sec:real_data.descripreproc}) excluding the last $H$ months. The corresponding out-of-sample prediction performance is evaluated on a testing sample covering the final $H$ months, from December 2023 minus $H-1$ months through December 2023. We employ the AR model as the baseline for comparison. For each simulation repetition, we assess the prediction accuracy using two performance measures: the $H$-step relative root prediction mean squared error (rPMSE) and the $H$-step relative sign prediction error (rSign) for each model $m$. 
The $H$-step rPMSE performance measure is defined as 
\begin{equation}\label{equ:rPMSE}
    \operatorname{rPMSE}^{AR}_m(H) = \sqrt{\frac{\overline{\operatorname{PMSE}}_m(H)}{\overline{\operatorname{PMSE}}_{AR}(H)}},
\end{equation}
where $\overline{\operatorname{PMSE}}_m(H) = \sum_{i=1}^{Q} \left( \sum_{h=1}^{H} (\hat{y}^{(i)}_{T+h,m} - y^{(i)}_{T+h})^2/H\right)/Q$ with $\hat{y}^{(i)}_{T+h,m}$ the $h$-step-ahead inflation prediction based on specific model $m$ for the $i$th simulation repetition (out of $Q$ total). rPMSE measures the relative prediction power compared to the baseline AR model, and lower rPMSE values indicate better prediction accuracy. The $H$-step rSign performance measure is defined as
\begin{equation}\label{equ:rSign}
  \operatorname{rSign}^{AR}_m(H) = \frac{\overline{\operatorname{Sign}}_m(H)}{\overline{\operatorname{Sign}}_{AR}(H)},  
\end{equation}
where $\overline{\operatorname{Sign}}_m(H) = \sum_{i=1}^{Q} \left(\sum_{h=1}^{H} (\calI(\operatorname{sign}(\hat{y}^{(i)}_{T+h,m}) - \operatorname{sign}(y^{(i)}_{T+h}))/H\right)/Q$ is the sign prediction error, $\calI(\cdot)$ represents the indicator function, and $\operatorname{sign}(\cdot)$ denotes the sign of a given number. For each forecasting model, we consider the $H$-step-ahead predictions for $H \in \{8,\ldots, 15\}$ and repeat each simulation example $Q=500$ times.

We further evaluate the prediction inference performance in terms of prediction intervals. For the LLM-CPI, we exploit both the Box--Jenkins (BJ) prediction interval and the bootstrap (BOOT) prediction interval. We compare them with the BJ prediction interval based on the traditional AR model\footnote{ 
 See Equation \eqref{equ:ar_ci} in Section \ref{new.Sec.forecastmods} for details on the CI calculation.}. 

We consider two performance measures to assess the inference performance. The first measure is the $H$-step average coverage rate defined as 
$$
\operatorname{Coverage}_m(H) = \frac{1}{HQ} \sum_{i=1}^Q\sum_{h=1}^H \calI(y_{T+h} \in \operatorname{PI}_{m,i}^{(h)}),
$$
where $\operatorname{PI}^{(h)}_{m,i}$ denotes the $h$-step-ahead prediction interval constructed by each method $m$ for the $i$th simulation repetition. The second measure is the $H$-step average interval length defined as 
$$
\operatorname{Length}_m(H) = \frac{1}{HQ} \sum_{i=1}^Q  \sum_{h=1}^H |\operatorname{PI}^{(h)}_{m,i}|,
$$
where $|\operatorname{PI}_{m,i}^{(h)}|$ represents the length of the interval. These two quantities jointly measure the effectiveness of the prediction intervals given by different forecasting models. We set  $\alpha=0.05$ for all prediction intervals so 
a well performed prediction interval is expected to have the nominal coverage rate of $1 - \alpha = 95\%$. 
Prediction intervals with smaller lengths while maintaining the nominal coverage rate are desired.

\subsection{Simulation results} \label{Sec.simuresu}

Tables \ref{tab:simu_true_mse} and \ref{tab:simu_true_sign} present the results on the rPMSE and rSign for all the forecasting models across different prediction horizons $H$ and correlation levels $\rho$. As seen in Tables \ref{tab:simu_true_mse} and \ref{tab:simu_true_sign}, the LLM-CPI model consistently achieves both lower rPMSE and rSign than the traditional benchmark models, indicating its superior prediction performance. Such performance advantage becomes even more pronounced as the model error correlation level $\rho$ increases. 
Importantly, the improvement holds consistently across both short and long term horizons, demonstrating the advantages and robustness 
of the LLM-CPI method in forecasting tasks compared to classical approaches.

Table \ref{tab:simu_true} summarizes the average coverage rates of prediction intervals and the average interval lengths (included in the parentheses) across different forecast horizons $H$ and correlation levels $\rho$. The AR method achieves high coverage consistently, and the variants of the LLM-CPI method with the BJ and bootstrap intervals generally achieve coverage rates that are close to the nominal level in the short term. For the long term, the coverage rates of both variants occasionally fall slightly below the nominal level but remain competitive overall. In terms of inference efficiency, both variants of LLM-CPI method consistently construct substantially tighter prediction intervals compared to the traditional AR model, highlighting their practical capability in producing sharper forecast intervals while maintaining good coverage rate. 

We also conduct three additional simulation experiments in Sections \ref{new.Sec.robustness.1}--\ref{new.Sec.robustness.3} to evaluate the robustness of the suggested LLM-CPI model under various model misspecification or overfitting scenarios. Specifically, these experiments consider scenarios involving the omitted relevant predictor, model overfitting, and non-Gaussian error distributions (e.g., \textit{t}-distributions). The results therein demonstrate that the LLM-CPI model still exhibits strong prediction inference performance across all three scenarios.

\section{Real data application} \label{sec:real_data}

In this section, we further showcase the practical utility and advantages of the LLM-CPI method for CPI prediction inference on a real data set in comparison to several popular benchmark methods\footnote{The corresponding R code can be found at \url{https://github.com/suntiansheng/LLM-CPI-prediction-and-inference}.}.

\subsection{Real data description and preprocessing} \label{sec:real_data.descripreproc}

The target variable in our real data application is the observed monthly inflation rate, measured using the monthly CPI data published by the National Bureau of Statistics of China (NBSC)\footnote{\url{https://data.stats.gov.cn/english/easyquery.htm?cn=A01}}, specifically from the data set titled ``Consumer Price Index (the last month = $100$).'' This index uses the previous month as the reference point (i.e., base value $= 100$), and the monthly inflation is reflected as the marginal change in CPI relative to the prior month. To standardize the data, we subtract $100$ from each observed CPI value and then divide it by the standard deviation of the series. The standardized CPI value at time $t$ is denoted as $y_t$ with $t =1,\ldots, T$. In contrast, the surrogate inflation response is constructed from the LLM-generated daily inflation index and denoted as $y^{S}_{t,k}$ with $t =1,\ldots,T$ for $1 \leq k \leq 3$, as described in Section \ref{Sec.jointmodel}.

The predictors considered in the LLM-CPI model consist of two parts. One part is a widely recognized inflation-related macroeconomic variable--the national urban unemployment rate, also reported monthly by the NBSC. Such indicator reflects the surveyed unemployment rate in urban areas. The unemployment rate is standardized by subtracting the mean and dividing by the standard deviation of the series. The standardized unemployment rate at time $t$ is denoted as $z_t$ with $t =1,\ldots, T$. The other part is online text embedding features. These text embedding features capture economic narratives extracted from the online Weibo posts. We summarize the latent semantic information using two types of text embeddings: the LDA embedding $\bx_t^{\text{LDA}}$ and the BERT embedding $\bx_t^{\text{BERT}}$ as discussed in Section \ref{Sec.textembed}. For easy reference, we will name the LLM-CPI method with the LDA and BERT embeddings as
LLM+LDA and LLM+BERT, respectively.

\begin{figure}[t]
    \centering
    \begin{subfigure}[b]{0.4\textwidth}  
        \includegraphics[width=\linewidth]{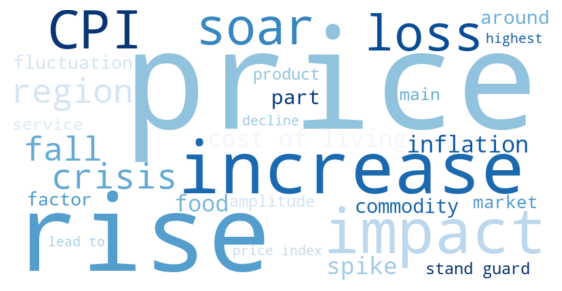}
        \caption{Topic on multidimensional inflation drivers: commodity markets and spatial price dynamics} 
        \label{fig:topic1_e}
    \end{subfigure}
    \hspace{0.5cm}
    \begin{subfigure}[b]{0.4\textwidth}
        \includegraphics[width=\linewidth]{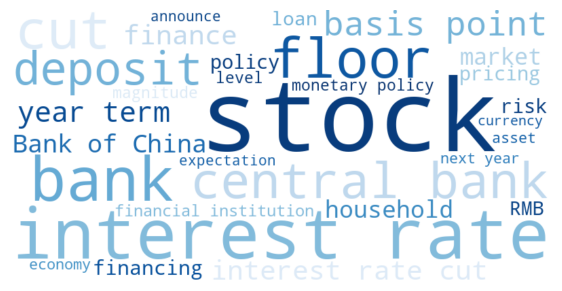}
        \caption{Topic on monetary policy transmission and inflationary expectation dynamics } 
        \label{fig:topic10_e}
    \end{subfigure}
 \caption{The word cloud plots of two selected LDA embedding features given by model selection criterion in \ref{app:sec:model_selection}.}
    \label{fig:wordcloud5year}
\end{figure}

To improve model interpretability, we apply a model selection criterion to select important features from the two text embeddings above based on the training data, which corresponds to the time period of January 2019 to October 2021. See Section \ref{app:sec:model_selection} for details on the time-series model selection. For the LDA embedding, the number of features given by model selection is $2$. Figure \ref{fig:wordcloud5year} presents the word cloud plots for the top keywords associated with the selected LDA embedding features. These keywords provide useful insights into the economic narratives captured by the LLM+LDA method and highlight the topics that are most strongly correlated with inflation dynamics. Specifically, the prominent keywords in Figure \ref{fig:topic1_e} are mainly related to multidimensional inflation drivers, especially the nonlinear interactions between commodity volatility (``food'', ``commodity'', ``product'', ``fluctuation''), regional economic disparities (``region'', ``crisis'', ``loss'', ``rise''), and consumer price formation mechanisms (``CPI'', ``price'', ``inflation''), capturing the spatial-temporal propagation of inflationary pressures. The topic words in Figure \ref{fig:topic10_e} reveal the interplay between central bank policy instruments (``central bank'', ``monetary policy'', ``RMB'', ``announce'') and market response mechanisms, focusing on two critical transmission channels. One is the borrowing and lending signals, especially how benchmark rate changes (``basis point'', ``floor'', ``stock'') influence commercial lending rates (``loan'') and deposit behavior (``deposit''). Another is the financial intermediation (``financial institution'', ``financing'', ``asset'', ``expectation'', ``bank''). See Tables \ref{tab:5year_19-23_tab1} and \ref{tab:5year_19-23_tab2} in Section \ref{new.Sec.add.real.LDAtopic} for the lists of topic words in Figures \ref{fig:topic1_e} and \ref{fig:topic10_e}, respectively, and related hashtags on Weibo. 

\begin{figure}[t]
    \centering
    \includegraphics[width=10cm]{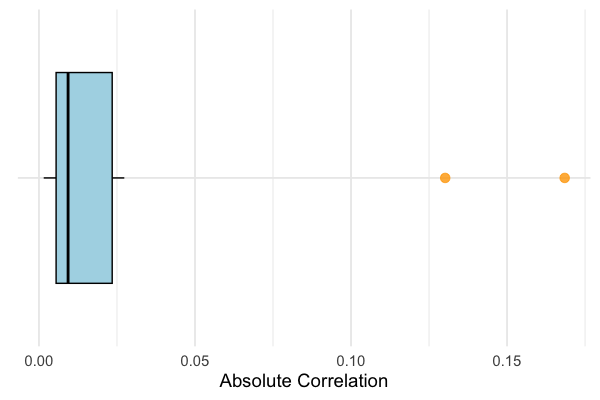}
    \caption{
     The boxplot of the absolute values of correlations between BERT embedding features and AR model residuals.
    The two selected BERT embedding features given by model selection criterion in \ref{app:sec:model_selection} are highlighted in yellow.}
    \label{fig:barplot}
\end{figure}

The number of the BERT embedding features given by model selection is also $2$. Although the BERT embeddings lack explicit interpretability, their correlations with the residuals of the AR model \eqref{equ:pure_AR} demonstrate the prediction power of each embedding feature. Notably, the selected BERT embedding features exhibit significantly stronger correlations with the residuals than the remaining ones, as seen in Figure \ref{fig:barplot}. With slight abuse of notation, we will denote the selected LDA embedding features as $\bx_t^{\text{LDA}} \in \R^2$ and the selected BERT embedding features as $\bx_t^{\text{BERT}} \in \R^2$ with $t = 1,\ldots,T$.

\subsection{Out-of-sample forecasting} \label{sec:real_data.forecasting}

We now assess the out-of-sample forecasting performance of the LLM+LDA and LLM+BERT, i.e., two variants of the LLM-CPI method. 
To isolate the effect of the text embedding features, we first exclude the macroeconomic predictor $z_t$ (i.e., the unemployment rate). The corresponding empirical results with unemployment rate are detailed in Section \ref{new.Sec.add.real.forecasting}. We compare the LLM+LDA and LLM+BERT models against three well-established inflation forecasting benchmarks: the AR model \eqref{equ:pure_AR}, RW model \eqref{equ:random_walk}, and AVE model \eqref{equ:mean_model}. In addition, we include the direct text-based model \eqref{equ:target2} (without unemployment rate) with the LDA and BERT embeddings (i.e., without the LLM-CPI model structure) in the comparison. For simplicity, we refer to the text-based prediction model with the LDA embeddings as the LDA model, and the text-based prediction model with the BERT embeddings as the BERT model (with slight abuse of terminology). Detailed model specifications are provided in Section \ref{new.Sec.forecastmods}.

The out-of-sample forecast period spans from $H$ months prior to December 2023 through December 2023, yielding $H$ forecast steps. Such evaluation window is strictly independent of the sample used for model fitting and model selection, ensuring the integrity of the out-of-sample assessment. We choose the AR model as the baseline and evaluate the performance using the relative root prediction mean squared error ($\operatorname{rPMSE}^{AR}(H)$) and the relative sign
prediction error ($\operatorname{rSign}^{AR}(H)$) defined in \eqref{equ:rPMSE} and \eqref{equ:rSign}, respectively. For both performance measures, we now have $Q = 1$ due to a single observation for each month.

\begin{table}[t]
\centering
\small
\caption{The $\operatorname{rPMSE}^{AR}(H)$ results across different horizons $H$ without unemployment rate}
\label{tab:rpmse_no_unemp}
\begin{tabular}{lccccccccc}
\toprule
Method & $8$ & $9$ & $10$ & $11$ & $12$ & $13$ & $14$ & $15$ & Ave. \\
\midrule
RW        & 0.846 & 1.133 & 1.673 & 2.389 & 0.921 & 1.036 & 0.984 & 1.272 & 1.282 \\
AVE       & 0.908 & 0.817 & 0.908 & 1.347 & 0.932 & 1.037 & 1.089 & 1.044 & 1.010 \\
LDA       & 0.818 & 0.840 & 0.851 & 0.819 & 0.892 & 0.916 & 0.902 & 0.886 & 0.865 \\
BERT        & 1.578 & 1.646 & 1.743 & 1.339 & 1.289 & 1.343 & 1.379 & 1.407 & 1.466 \\
LLM+LDA   & 0.834 & 0.775 & 0.786 & 0.589 & 0.894 & 0.954 & 0.972 & 0.978 & 0.848 \\
LLM+BERT    & 0.735 & 0.831 & 0.859 & 1.033 & 1.055 & 1.100 & 1.063 & 1.036 & 0.964 \\
\bottomrule
\end{tabular}\\
{\footnotesize Note: The relative PMSE values compared to the AR benchmark. Smaller values indicate better performance.}
\end{table}

\begin{table}[t]
\centering
\small
\caption{The $\operatorname{rSign}^{AR}(H)$ across different horizons $H$ without unemployment rate}
\label{tab:rsign_no_unemp}
\begin{tabular}{lccccccccc}
\toprule
Method    & $8$ & $9$ & $10$ & $11$ & $12$ & $13$ & $14$ & $15$ & Ave. \\
\midrule
RW        & 0.429 & 0.429 & 0.429 & 1.333 & 0.800 & 0.800 & 0.800 & 1.833 & 0.857 \\
AVE       & 0.429 & 0.429 & 0.571 & 1.333 & 1.200 & 1.000 & 1.600 & 1.500 & 1.008 \\
LDA       & 0.429 & 0.714 & 0.714 & 0.833 & 0.800 & 0.800 & 0.800 & 0.833 & 0.740 \\
BERT        & 0.653 & 1.286 & 1.224 & 1.833 & 2.400 & 2.080 & 3.360 & 3.750 & 2.073 \\
LLM+LDA   & 0.143 & 0.143 & 0.143 & 0.167 & 0.400 & 0.600 & 0.400 & 0.333 & 0.291 \\
LLM+BERT    & 0.143 & 0.286 & 0.286 & 0.833 & 1.200 & 1.200 & 1.000 & 0.833 & 0.723 \\
\bottomrule
\end{tabular}\\
{\footnotesize Note: The relative sign prediction error values compared to the AR benchmark. Smaller values indicate better performance.}
\end{table}

Tables \ref{tab:rpmse_no_unemp} and \ref{tab:rsign_no_unemp} summarize the results across different forecast horizons $H$. Several key findings emerge in view of Tables \ref{tab:rpmse_no_unemp} and \ref{tab:rsign_no_unemp}.
\begin{itemize}
    \item \textit{LDA embedding enhances prediction accuracy}. The inclusion of the LDA embedding in the LDA model (without the LLM-CPI model structure) for text-based prediction reduces prediction errors relative to traditional benchmarks. Specifically, the LDA model achieves an average $\operatorname{rPMSE}^{AR}(H)$ of $0.865$, outperforming the RW model ($1.282$), AVE model ($1.010$), and AR model ($1.000$), while also achieving a substantial reduction in the relative sign prediction error ($\operatorname{Sign}^{AR}(H)$ = $0.740$). Such improvement demonstrates that text embeddings can capture economically meaningful signals for inflation forecasting. In particular, we see that the reduction in the relative sign prediction error is consistent across different horizons $H$, from short-term ($H=8$) to long-term ($H=15$) forecasts, suggesting robust performance of the LDA model in identifying the inflation trends. In contrast, the BERT model (without the LLM-CPI model structure) exhibits poor performance, possibly due to the more noisy nature of the BERT embedding features.
    

    \item \textit{LLM-CPI method substantially improves prediction performance}. Incorporating the LLM-powered joint time series modeling in the LLM-CPI variants yields additional, consistent gains over their non-LLM-powered counterparts. The LLM+LDA model achieves the lowest $\operatorname{rPMSE}^{AR}(H)$ of $0.848$ as well as the lowest average $\operatorname{Sign}^{AR}(H)$ of $0.291$, demonstrating the strong effectiveness of the LLM-CPI model structure. Similarly, the LLM+BERT model improves over the standalone BERT model (without the LLM-CPI model structure), reducing $\operatorname{rPMSE}^{AR}(H)$ from $1.466$ to $0.964$ and $\operatorname{Sign}^{AR}(H)$ from $2.073$ to $0.723$, highlighting the benefits of incorporating the LLM-powered surrogate modeling in stabilizing noisy text signals.
\end{itemize}

We provide in Section \ref{new.Sec.add.real.forecasting} additional empirical results on including the unemployment rate as a predictor, with the autoregressive model with exogenous unemployment rate serving as the baseline. The results therein continue to demonstrate that the LLM-CPI model outperforms all the benchmark ones.

\subsection{High-frequency CPI prediction inference by LLM-CPI} \label{sec:predinf.LLM-CPI}

We further evaluate the prediction inference performance of the LLM-CPI method in the context of high-frequency CPI forecasting. We employ the Box--Jenkins (BJ) procedure for constructing the prediction intervals in LLM-CPI. The corresponding results for LLM-CPI with the bootstrap prediction interval are contained in Section \ref{new.Sec.add.real.inference.boot}. 

Table \ref{tab:real_no_unemp} reports the prediction interval coverage rates and average interval lengths across different forecast horizons $H = 8$ to $15$, under the setting that excludes the macroeconomic predictor $z_t$ (i.e., unemployment rate). The major findings are summarized below.

\begin{table}[t]
\centering
\small
\caption{The $\operatorname{Coverage}_m(H)$ and $\operatorname{Length}_m(H)$ results across different horizons $H$ without unemployment rate (LLM with BJ interval).}
\label{tab:real_no_unemp}
\small{
\begin{tabular}{lccccccccc}
\toprule
Method            & $8$ & $9$ & $10$ & $11$ & $12$ & $13$ & $14$ & $15$ & Ave. \\
\midrule
AR                & 1.000 & 1.000 & 1.000 & 1.000 & 1.000 & 1.000 & 1.000 & 1.000 & 1.000 \\
                  & (4.093) & (4.133) & (4.179) & (4.148) & (4.115) & (4.164) & (4.195) & (4.247) & (4.159) \\
LDA               & 1.000 & 1.000 & 1.000 & 1.000 & 1.000 & 1.000 & 1.000 & 1.000 & 1.000 \\
                  & (3.752) & (3.788) & (3.838) & (3.843) & (3.753) & (3.799) & (3.842) & (3.894) & (3.814) \\
BERT                & 1.000 & 1.000 & 1.000 & 1.000 & 1.000 & 1.000 & 1.000 & 1.000 & 1.000 \\
                  & (4.135) & (4.176) & (4.224) & (4.208) & (4.173) & (4.225) & (4.267) & (4.324) & (4.216) \\
LLM+LDA           & 1.000 & 1.000 & 1.000 & 1.000 & 0.917 & 0.923 & 0.929 & 0.933 & 0.963 \\
                  & (3.212) & (3.235) & (3.251) & (3.315) & (3.214) & (3.233) & (3.284) & (3.289) & (3.254) \\
LLM+BERT           & 1.000 & 1.000 & 1.000 & 1.000 & 1.000 & 1.000 & 1.000 & 1.000 & 1.000 \\
                  & (3.530) & (3.411) & (3.424) & (3.409) & (3.442) & (3.480) & (3.518) & (3.554) & (3.471) \\
\bottomrule
\end{tabular}}\\
{\footnotesize Note: The values in the parentheses are interval length, and coverage near the nominal level of $0.95$ with smaller interval length is preferred.}
\end{table}

\begin{itemize}
    \item \textit{LDA embedding improves prediction intervals}. Models incorporating the LDA text embeddings yield shorter prediction intervals while maintaining the nominal coverage. For instance, the LDA model (without the LLM-CPI model structure) achieves an average interval length of $3.814$ compared to that of $4.159$ for the AR benchmark, with coverage rate remaining at $100\%$ across all forecast horizons $H$. Such finding reinforces earlier results that text signals can contain valuable prediction information and also enhance the uncertainty quantification in inflation forecasting. Similarly as before, the BERT model (without the LLM-CPI model structure) exhibits relatively poor performance.

    \item \textit{LLM-CPI method yields further gains in inference efficiency}. Both the LLM+LDA and LLM+BERT models achieve substantial reductions in prediction interval length relative to their non-LLM-powered counterparts. In particular, the LLM+LDA model reduces the average length to $3.254$ while maintaining a high average coverage rate of $0.963$. The LLM+BERT model also matches the nominal coverage level with slightly longer intervals on average $(3.471)$. These improvements illustrate the LLM-CPI model's ability in producing tighter and more informative prediction intervals by leveraging the LLM-powered surrogate signals through the LLM-powered joint time series modeling.
\end{itemize}

Section \ref{new.Sec.add.real.inference.bj} documents additional empirical results on the LLM-CPI with the BJ and bootstrap prediction intervals, including unemployment rate as an additional predictor. The autoregressive model with exogenous unemployment rate is employed as the baseline for prediction inference. The results therein consistently show that the LLM-CPI model with the BJ prediction interval outperforms all the benchmark models. 

\subsection{The impact of COVID-19 on inflation} \label{Sec.covid19}

A fundamental question in the evaluation of machine learning-based economic forecasting models is whether the performance gains stem from genuine predictive content or merely from overfitting or spurious correlations. To address such concern, we utilize the COVID-19 pandemic as a natural exogenous shock and perform a structural robustness check on the LLM-CPI framework.

The COVID-19 pandemic, which began in early 2020, represents a significant exogenous shock with lasting effects on global economic conditions, particularly on the inflation dynamics. Such unprecedented event likely triggered structural shifts in both the macroeconomic environment and text narratives. To account for these potential disruptions, we divide the data into two distinct time periods: the  pre- and during-lockdown period from January 1, 2019 to December 31, 2021, and the  post-lockdown period from January 1, 2022 to December 31, 2023\footnote{ The timeline of the COVID-19 pandemic in mainland China is available at \url{https://en.wikipedia.org/wiki/COVID-19_pandemic_in_mainland_China.}}.  Such time segmentation enables us to examine the changes in text content, topic structures, and model performance before and after the peak of the COVID-19 pandemic.

\begin{figure}[t]
    \centering
    \includegraphics[width=8cm]{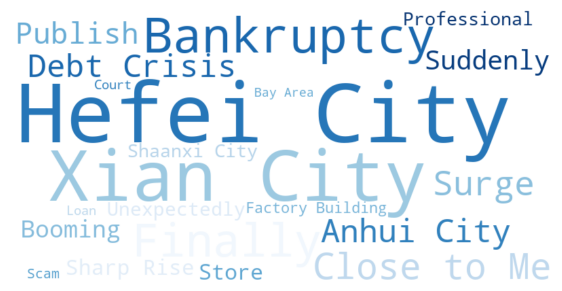}
    \caption{The word cloud plot for a selected LDA topic for the pre- and during-lockdown period, on the pre- and during-lockdown period urban financial fragility and structural debt contagion.
    }
    \label{fig:topic12_19}
\end{figure}

\begin{figure}[t]
    \centering
    \begin{subfigure}[b]{0.4\textwidth}  
        \includegraphics[width=1\linewidth]{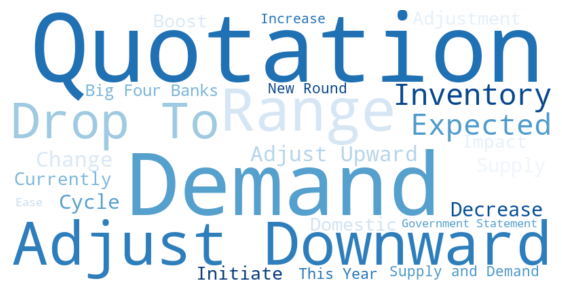}
        \caption{Topic on market and policy-driven supply-demand adjustment mechanisms} 
        \label{fig:topic16_21}
    \end{subfigure}
    \hspace{0.5cm}
    \begin{subfigure}[b]{0.4\textwidth}
        \includegraphics[width=1\linewidth,height=0.55\linewidth]{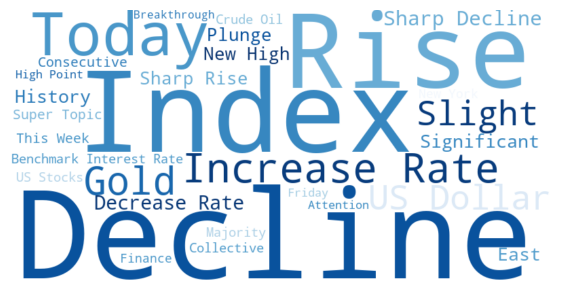}
        \caption{Topic on live financial index tracking and global commodity volatility } 
        \label{fig:topic13_21}
    \end{subfigure}

    \vspace{0.5cm} 

    \begin{subfigure}[b]{0.4\textwidth}
        \includegraphics[width=1\linewidth]{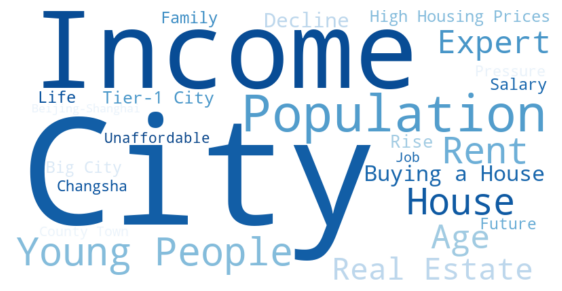}
        \caption{Topic on urban housing affordability crisis -- intergenerational pressures and economic mobility in tiered cities } 
        \label{fig:topic3_21}
    \end{subfigure}
    \hspace{0.5cm}
    \begin{subfigure}[b]{0.4\textwidth}
        \includegraphics[width=1\linewidth]{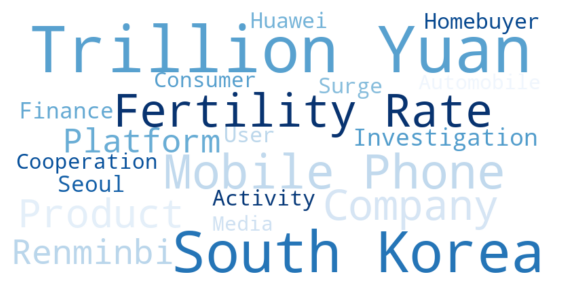}
        \caption{Topic on South Korea's socioeconomic nexus -- housing affordability, tech disruption, and demographic transition} 
        \label{fig:topic17_21}
    \end{subfigure}

    \caption{The word cloud plots for four selected LDA topics during the post-lockdown period.
    }
    \label{fig:wordcloud4}
\end{figure}

For each period, we independently train the LDA model to extract period-specific latent topics. To evaluate the prediction performance, we designate the final six months of each period as the testing sample, with the remaining (i.e., earlier) months used for training. We also apply the model selection procedure on the training sample as discussed in Section \ref{app:sec:model_selection} to reduce the dimensionality of the text embeddings. Tables \ref{tab:COV19_19-21_mse}--\ref{tab:COV19_21-23_sign} in Section \ref{new.Sec.add.real.forecast.pand} report the root prediction mean squared error (PMSE) and the sign prediction error (Sign) for both pre- and during-lockdown period and post-lockdown period\footnote{We do not use the relative errors here since the baseline AR model occasionally yields zero error when the forecast horizon $H$ is small.}. These results unveil that the LLM-CPI model usually outperforms the benchmark models in terms of both PMSE and rSign  across both periods. Figures \ref{fig:topic12_19} and \ref{fig:wordcloud4} display the corresponding topic visualizations before and after the lockdown. The differences in the dominant topics and words illustrate a clear shift in the text narratives.

To enhance the interpretability of the extracted topics, we analyze the original Weibo posts to uncover their semantic meaning. Many of these posts feature self-labeled hashtags by users that summarize their main themes, serving as human-generated, interpretable topic labels. To utilize such information, we adopt the following approach: for each topic, we first identify the top $10$ highest-probability keywords generated by the LDA model. We then retrieve all posts containing each of those keywords and extract the associated hashtags. From these, we select the $10$ most frequently occurring hashtags to represent the concrete, human-readable meaning of each topic. Table \ref{tab:COV19_19-21_tab} (corresponding to Figure \ref{fig:topic12_19}) in Section \ref{new.Sec.add.real.LDAtopic} presents the most-discussed original hashtags corresponding to our selected topic during the lockdown era, while Tables \ref{tab:COV19_21-23_tab3}--\ref{tab:COV19_21-23_tab4} (corresponding to Figure \ref{fig:wordcloud4}) in Section \ref{new.Sec.add.real.LDAtopic} list the four selected topics and related hashtags during the post-lockdown period.

In view of Figure \ref{fig:topic12_19} and Table \ref{tab:COV19_19-21_tab}, it is seen that the lockdown-era topics emphasize crisis-related themes such as debt defaults, bankruptcies, and city lockdowns, reflecting the immediate economic fallout. This topic also reveals latent vulnerabilities in regional financial ecosystems through lexical patterns documenting debt accumulation cycles, real estate overexposure, and professional service sector instability.

In contrast, the post-lockdown topics shown in Figure \ref{fig:wordcloud4} and Tables \ref{tab:COV19_21-23_tab3}--\ref{tab:COV19_21-23_tab4} revert to more regular macroeconomic themes, including pricing, trade, and demands, suggesting a normalization of economic discourse. Specifically, Figure \ref{fig:topic16_21} and Table \ref{tab:COV19_21-23_tab3} exhibit concentrated terms around supply chain resilience and energy transition costs, reflecting input-driven inflationary pressures. Figure \ref{fig:topic13_21} and Table \ref{tab:COV19_21-23_tab2} focus on monitoring daily market movements, currency fluctuations (particularly USD), commodity price trends (crude oil/gold), and benchmark index performance, with emphasis on rate-sensitive assets and short-term market reactions to macroeconomic signals. Figure \ref{fig:topic3_21} and Table \ref{tab:COV19_21-23_tab1} integrate macroeconomic trends with the grassroots experiences to map how housing costs reshape urban demographics, family structures, and long-term financial planning across socioeconomic strata. Figure \ref{fig:topic17_21} and Table \ref{tab:COV19_21-23_tab4} link geopolitical risk and commodity hoarding, encoding exogenous shock amplification, which incorporates multinational corporate strategies (``cooperation''), media-driven consumer narratives, and policy responses to Seoul's  housing-supply crisis. It also reveals how currency dynamics (RMB 
integration) and investigative regulatory frameworks attempt to balance the technological innovation with social stability in Asia's fourth-largest economy. 



\section{Discussions} \label{sec:discu}

We have investigated in this paper the problem of consumer price index (CPI) forecasting. Motivated by the recent developments in large language models (LLMs), our suggested method of LLM-powered CPI prediction inference (LLM-CPI) is rooted on the LLM-powered joint time series model of both observed monthly CPIs and LLM-generated daily CPI surrogates. Such a model exploits the correlations between the low-frequency survey-based inflation labels and the high-frequency LLM-based inflation labels generated using an online text data set we collected, conditional on the lagged monthly CPIs, lagged LLM-generated daily CPI surrogates, macroeconomic indicators, and online text embeddings. With theoretical guarantees, LLM-CPI has been shown to provide accurate and tight CPI prediction inference results at both monthly and daily levels empirically, thanks to the power of LLMs such as ChatGPT and the trained BERT models as well as text embeddings via LDA and BERT.

It would be interesting to incorporate LLM-based inflation labels generated by different LLM tools into the joint model. We may consider more advanced text embedding models to extract useful text features for this problem. It would also be beneficial to consider more general models beyond the ARX and VARX models for the joint modeling with certain sparsity and latent factor structures. These problems are beyond the scope of the current paper and will be interesting topics for future research.

\bibliographystyle{chicago}
\bibliography{ref}


\newpage
\appendix
\setcounter{page}{1}
\setcounter{section}{0}
\renewcommand{\theequation}{A.\arabic{equation}}
\setcounter{equation}{0}

\begin{center}{\bf \Large Supplementary Material to ``LLM-Powered CPI Prediction Inference with Online Text Time Series''}

\bigskip

Yingying Fan, Jinchi Lv, Ao Sun and Yurou Wang
\end{center}

\noindent This Supplementary Material contains all the proofs of main results, details on the online text data collection and preprocessing, the implementation details of the suggested LLM-CPI method, and additional simulation results on the robustness of LLM-CPI, as well as some additional real data results. For a vector $\bx \in \R^p$, let $\|\bx\|$ denote its $\ell_2$-norm. For a matrix $\bB \in \R^{p_1 \times p_2}$, denote by $\|\bB\|$ the spectral norm, and $\|\bB\|_F$ the Frobenius norm.

\section{Proofs of main results} \label{sec.proofs}

\subsection{Proof of Theorem \ref{lem:prediction}} \label{sec.proof.lem:prediction}

For each $t \geq q_1$, let $Y_t = (y_t, \ldots, y_{t-q_1+1})^\top \in \R^{q_1}$ with the time index of $y$ running backward. We can rewrite the joint ARX model \eqref{equ:trans} as 
    \begin{equation}\label{equ:march06:01}
         Y_t = \bA Y_{t-1} + \bB \bx_{t} + \bTheta \bz_t  + \boldsymbol{\Gamma} \bD(\by_t^S) + \bE_t,
    \end{equation}
    where 
    \begin{equation}\label{def:A}
        \bA:=\left[\begin{array}{ccccc}
a_1 & a_2 & \ldots & a_{q_1-1} & a_{q_1} \\
1 & 0 & \ldots & 0 & 0 \\
0 & 1 & & 0 & 0 \\
\vdots & & \ddots & \vdots & \vdots \\
0 & 0 & \ldots & 1 & 0
\end{array}\right] \in \R^{q_1 \times q_1}, \quad \bE_t:=\left[\begin{array}{c}
e_t \\
0 \\
\vdots \\
0
\end{array}\right]  \in \R^{q_1},
    \end{equation}
$\bB = (\bdelta^\top, \mathbf{0}^\top,\ldots, \mathbf{0}^\top) \in \R^{q_1 \times p}$, $\bTheta = (\btheta^\top, \mathbf{0}^\top,\ldots, \mathbf{0}^\top) \in \R^{q_1 \times b}$, and $\bGamma = (\bgamma^\top, \mathbf{0}^\top,\ldots, \mathbf{0}^\top) \in \R^{q_1 \times K}$. Then it holds that for each $h \geq 1$, 
\begin{equation}\label{equ:march05:equ1}
   \begin{aligned}
       Y_{T+h} =& \bA Y_{T+h-1} + \bB \bx_{T+h} + \bTheta \bz_{T+h}  + \boldsymbol{\Gamma} \bD(\by_{T+h}^S) + \bE_{T+h}\\
       =&\bA^h Y_{T} + \sum_{r=0}^{h-1} \bA^r\left\{ \bB \bx_{T+h-r} + \bTheta \bz_{T+h-r}  + \boldsymbol{\Gamma} \bD(\by_{T+h-r}^S)\right\} + \sum_{r=0}^{h-1} \bA^r \bE_{T+h-r},
   \end{aligned} 
\end{equation}
where in the last step above we have sequentially used identity \eqref{equ:march06:01} with $t= T+h-1, \ldots, T$.  

Similar to \eqref{equ:march06:01}, letting $\hat{Y}_{T+1} = (\hat{y}_{T+1}, y_{T}, \ldots, y_{T-q_1+2})^\top \in \R^{q_1}$, we can also rewrite the one-step-ahead forecast \eqref{equ:prediction} given by the joint ARX model \eqref{equ:trans} as
$$
\hat{Y}_{T+1} =  \hbA Y_{T} + \hbB \bx_{T+1} + \hbTheta \bz_{T+1}+\widehat{\boldsymbol{\Gamma}} \hbD(\by_{T+1}^S),
$$
where 
    $$
    \hbA:=\left[\begin{array}{ccccc}
\hat{a}_1 & \hat{a}_2 & \ldots & \hat{a}_{q_1-1} & \hat{a}_{q_1} \\
1 & 0 & \ldots & 0 & 0 \\
0 & 1 & & 0 & 0 \\
\vdots & & \ddots & \vdots & \vdots \\
0 & 0 & \ldots & 1 & 0
\end{array}\right] \in \R^{q_1 \times q_1},
    $$
$\hbB = (\hat{\bdelta}^\top, \mathbf{0}^\top,\ldots, \mathbf{0}^\top) \in \R^{q_1 \times p}$, $\widehat{\bTheta}= (\hat{\btheta}^\top, \mathbf{0}^\top,\ldots, \mathbf{0}^\top) \in \R^{q_1 \times b}$, and $\widehat{\bGamma} = (\hat{\bgamma}^\top,\mathbf{0}^\top,\ldots, \mathbf{0}^\top) \in \R^{q_1 \times K}$.

Further, for each given $h \geq 1$, let us define $\hat{Y}_{T+h} = (\hat{y}_{T+h}, \ldots, \hat{y}_{T+1}, y_{T}, \ldots, y_{T-q_1+h+1})^\top \in \R^{q_1}$. 
Similarly, the $h$-step-ahead forecast \eqref{equ:hstep} given by the joint ARX model \eqref{equ:trans} can be recursively rewritten as 
\begin{equation}\label{equ:march05:equ2}
 \begin{aligned}
 \hat{Y}_{T+h} =& \hbA \hat{Y}_{T+h-1} + \hbB \bx_{T+h} + \hbTheta \bz_{T+h} +\widehat{\boldsymbol{\Gamma}} \hbD(\by_{T+h}^S),\\
 =& \hbA^h Y_{T} + \sum_{r=0}^{h-1} \hbA^r \left\{\hbB \bx_{T+h-r} + \widehat{\bTheta} \bz_{T+h-r}+\widehat{\boldsymbol{\Gamma}} \hbD(\by_{T+h-r}^S)\right\}.   
\end{aligned}   
\end{equation}
In view of \eqref{equ:march05:equ1} and \eqref{equ:march05:equ2}, the $h$-step prediction error is given by 
\begin{equation}
\begin{aligned} \label{new.eq.001}
\hat{Y}_{T+h} - Y_{T+h} =& (\hbA^h -\bA^h )Y_{T} + \sum_{r=0}^{h-1} \hbA^r \left\{\hbB \bx_{T+h-r} + \widehat{\bTheta} \bz_{T+h-r}+ \widehat{\boldsymbol{\Gamma}} \hbD(\by_{T+h-r}^S)\right\}\\
&- \sum_{r=0}^{h-1} \bA^r\left\{ \bB \bx_{T+h-r} + \bTheta \bz_{T+h-r} +  \boldsymbol{\Gamma} \bD(\by_{T+h-r}^S)\right\}  -\sum_{r=0}^{h-1} \bA^r \bE_{T+h-r}\\
:=& v_1 + v_2 -\sum_{r=0}^{h-1} \bA^r \bE_{T+h-r},
\end{aligned}
\end{equation}
where $v_1 = (\hbA^h -\bA^h )Y_{T}$ and $v_2 = \sum_{r=0}^{h-1} \hbA^r \left\{\hbB \bx_{T+h-r} + \widehat{\bTheta} \bz_{T+h-r}+ \widehat{\boldsymbol{\Gamma}} \hbD(\by_{T+h-r}^S)\right\} - \sum_{r=0}^{h-1} \bA^r\left\{ \bB \bx_{T+h-r} + \bTheta \bz_{T+h-r} +  \boldsymbol{\Gamma} \bD(\by_{T+h-r}^S)\right\}$. We will prove that $\| v_1\| = O_p(\zeta_T^{-1})$ and $ \|v_2\| = O_p(\zeta_T^{-1})$. Then in light of (\ref{new.eq.001}), we can obtain that 
$$
\|\hat{Y}_{T+h} - Y_{T+h}  +\sum_{r=0}^{h-1} \bA^r \bE_{T+h-r}\| = O_p(\zeta_T^{-1}), 
$$
which yields the desired conclusion.

It remains to establish the above claim on $v_1$ and $v_2$. 
For term $v_1$, we will make use of the identity 
$$
\begin{aligned}
  \hbA^h -\bA^h =& \left(\sum_{r=0}^{h-1} \hbA^{h-1-r} \bA^{r}  \right)(\hbA -\bA)
\end{aligned}
$$
for each $h \geq 1$. With the aid of the above identity and noting that $\|\hbA\| \le \|\hbA -\bA\| + \| \bA\| = O_p(1)$ since $\|\bA\| = O(1)$ and $\hbA$ is a $\zeta_T$-consistent estimator of $\bA$ by the conditions, 
we can deduce that 
\begin{align*}
\| \hat{\bA}^h -\bA^h\| & \le \sum_{r=0}^{h-1}  \left\| \widehat{\bA}^{h-1-r} \bA^{r} \right\| \left\| \widehat{\bA} -\bA\right\| \leq \sum_{r=0}^{h-1}  \left\| \widehat{\bA}^{h-1-r} \bA^{r} \right\| \left\| \widehat{\bA} -\bA\right\|_F \\ &= 
 O_p\left(\left\| \widehat{\bA} -\bA\right\|_F\right) = O_p(\zeta_T^{-1}).
\end{align*}
Using this fact and $\| Y_{T}\| = O_p(1)$ due to the stationarity of the process, it holds that 
$$
\|v_1\| \le \|\hat{\bA}^h -\bA^h \| \| Y_{T}\| = O_p(\zeta_T^{-1}).
$$

We now turn to term $v_2$. Observe that 
\begin{equation} \label{new.eq.002}
\begin{aligned}
 v_2 =& \sum_{r=0}^{h-1} (\widehat{\bA}^r - \bA^r)\left\{\widehat{\bB} \bx_{T+h-r} + \widehat{\bTheta} \bz_{T+h-r} +\hat{\boldsymbol{\Gamma}} \widehat{\bD}(\by_{T+h-r}^S)\right\}  \\
 &+ \sum_{r=0}^{h-1} \bA^r\Big\{ (\widehat{\bB} - \bB) \bx_{T+h-r} + (\widehat{\bTheta} - \bTheta )\bz_{T+h-r} \\
 &\quad+ (\widehat{\boldsymbol{\Gamma}} \widehat{\bD}(\by_{T+h-r}^S) - \boldsymbol{\Gamma} \bD(\by_{T+h-r}^S))\Big\}\\
 =& v_{21} + v_{22}.
\end{aligned}
\end{equation}
Let us deal with term $v_{21}$ first. With an application of similar arguments as for term $v_1$, we can show that $\| \widehat{\bA}^r - \bA^r\| = O_p(\zeta_T^{-1})$, and the $\zeta_T$-consistency of the coefficient estimators entails that 
$$\widehat{\bB} \bx_{T+h-r} + \widehat{\bTheta} \bz_{T+h-r}+\hat{\boldsymbol{\Gamma}} \widehat{\bD}(\by_{T+h-r}^S) = O_p(1).$$ Hence, it follows from the Cauchy--Schwarz inequality that 
\begin{equation} \label{new.eq.003}
\|v_{21}\| = O_p(\| \widehat{\bA}^r - \bA^r\|)= O_p(\zeta_T^{-1}).
\end{equation}

For term $v_{22}$, note that
$$
\begin{aligned}
&\hat{\bgamma}^\top \widehat{\bD}(\by^S_t) - \bgamma^\top \bD(\by^S_t)\\
=&  \hat{\bgamma}^\top \left( \widehat{\bD}(\by^S_t) -\bD(\by^S_t)\right) +  (\widehat{\bgamma} - \bgamma)^\top\bD(\by^S_t)\\
=& \widehat{\bgamma}^\top \sum_{l=1}^{q_2} \left( \bA^S_{l} - \widehat{\bA}^S_{l} \right) \by^S_{t-l} + (\widehat{\bgamma} - \bgamma)^\top\bD(\by^S_t),\\
\end{aligned}
$$
where we have used the representation 
\begin{equation}\label{equ:D_error}
 \begin{aligned}
   \widehat{\bD}(\by^S_t) -\bD(\by^S_t)= \by^S_t - \sum_{l=1}^{q_2} \widehat{\bA}^S_{l}  \by^S_{t-l} - \left( \by^S_t - \sum_{l=1}^{q_2} \bA^S_{l}  \by^S_{t-l}\right) = \sum_{l=1}^{q_2} \left( \bA^S_{l} - \widehat{\bA}^S_{l} \right) \by^S_{t-l}.
\end{aligned}   
\end{equation}
Since $q_2$ is a constant and all the coefficient estimators are $\zeta_T$-consistent, we can obtain that 
\begin{equation} \label{new.eq.004}
\begin{aligned}
\|v_{22}\| & { \le} \sum_{r=0}^{h-1} \|\bA^r\|\Big\{ \|\widehat{\bB} - \bB\| \|\bx_{T+h-r}\|+ \|\widehat{\bTheta} - \bTheta \| \|\bz_{T+h-r}\| \\
 & \quad + \|\widehat{\bgamma}\| \sum_{l=1}^{q_2} \| \bA^S_{l} - \widehat{\bA}^S_{l} \| \|\by^S_{t-l}\| + \|\widehat{\bgamma} - \bgamma\| \|\bD(\by^S_t)\| \Big\}   \\
 & = O_p(\zeta_T^{-1}).
\end{aligned}
\end{equation}
Therefore, combining (\ref{new.eq.002})--(\ref{new.eq.003}) and (\ref{new.eq.004}) completes the proof of Theorem \ref{lem:prediction}.

\subsection{Proof of Theorem \ref{lem:variance}} \label{sec.proof.lem:variance}

We will first establish the consistency of the error variance estimate $\hat{\sigma}_e^2$. Denote by $\Delta_t = \bgamma^\top \left( \bD(\by^S_t) -\hbD(\by^S_t)\right)$ the approximation error. We can rewrite model \eqref{equ:trans} as 
\begin{equation}\label{equ:apr17:01}
\begin{aligned}
    y_{t} =& \sum_{l=1}^{q_1}\alpha_l y_{t-l} + \bz_t^\top \btheta +\bx_t^\top \bdelta  + \bgamma^\top\hbD(\by^S_t) +\Delta_t + e_t\\
 :=& \sum_{l=1}^{q_1}\alpha_l y_{t-l} + \bz_t^\top \btheta +\bx_t^\top \bdelta  + \bgamma^\top\hbD(\by^S_t) + \tilde{e}_t,
\end{aligned}
\end{equation}
where $\tilde{e}_t := e_t + \Delta_t$. To prove the consistency, we only need to show that 
\begin{equation}\label{lem2:goal1}
    \sqrt{\frac{1}{T-q_1}\sum_{t=q_1+1}^T (\hat{e}_t - \tilde{e}_t)^2} = o_p(1)
\end{equation}
and 
\begin{equation}\label{lem2:goal2}
    \sqrt{\frac{1}{T-q_1}\sum_{t=q_1+1}^T (\tilde{e}_t - e_t)^2} = o_p(1).
\end{equation} 
Then it follows from \eqref{lem2:goal1}, \eqref{lem2:goal2}, and the triangle inequality that 
\begin{align*}
   \left|\hat{\sigma}_e - \sqrt{\frac{1}{T-q_1}\sum_{t=q_1+1}^T e^2_t }\right| \leq \sqrt{\frac{1}{T-q_1}\sum_{t=q_1+1}^T (\hat{e}_t - \tilde{e}_t)^2} +  \sqrt{\frac{1}{T-q_1}\sum_{t=q_1+1}^T (\tilde{e}_t - e_t)^2} = o_p(1).
\end{align*}
Therefore, we can obtain that 
\begin{equation*}
    \begin{aligned}
\hat{\sigma}_e & = \sqrt{\frac{1}{T-q_1}\sum_{t=q_1+1}^T \hat{e}^2_t } = \sqrt{\frac{1}{T-q_1}\sum_{t=q_1+1}^T e^2_t } + o_p(1) \\
&= \sigma_e + o_p(1),
\end{aligned}
\end{equation*}
where in the last step above we have applied the law of large numbers (LLN).

To show the consistency of the variance estimator, it remains to establish \eqref{lem2:goal1} and \eqref{lem2:goal2} above. We will start with proving \eqref{lem2:goal1}. Using \eqref{equ:D_error} and basic algebra, we have
$$
\begin{aligned}
& \frac{1}{T-q_1}\sum_{t=q_1+1}^T \left((\hat{\bgamma} - \bgamma)^\top \hbD(\by_t^S) \right)^2\\
=&  \frac{1}{T-q_1}\sum_{t=q_1+1}^T \left[(\hat{\bgamma} - \bgamma)^\top \left(\bD(\by_t^S) + \sum_{l=1}^{q_2} \left( \bA^S_{l} - \widehat{\bA}^S_{l} \right) \by^S_{t-l} \right) \right]^2\\
{\le}&   \frac{q_2+1}{T-q_1}\left(\sum_{t=q_1+1}^T  \left( (\hat{\bgamma} - \bgamma)^\top \bD(\by_t^S) \right)^2 +\sum_{l=1}^{q_2}  \sum_{t=q_1+1}^T \left((\hat{\bgamma} - \bgamma)^\top(\bA^S_{l} - \widehat{\bA}^S_{l} ) \by^S_{t-l})\right)^2 \right)\\
{\le}&   \frac{q_2+1}{T-q_1}\left(\sum_{t=q_1+1}^T  \left( (\hat{\bgamma} - \bgamma)^\top \bD(\by_t^S) \right)^2 +\sum_{l=1}^{q_2}  \sum_{t=q_1+1}^T \| \hat{\bgamma} - \bgamma\|^2\|\bA^S_{l} - \widehat{\bA}^S_{l} \|^2 \|\by^S_{t-l} \|^2 \right)\\
=&O_p\left( \zeta_T^{-2} \lambda_{\max,D} \right) + O_p\left( T^{-1}\zeta_T^{-4} \sum_{t=1}^T  \|\by^S_{t} \|^2\right).
\end{aligned}
$$
The above result and some basic calculations yield that 
\begin{equation}\label{equ:june03:01}
    \begin{aligned}
    &\frac{1}{T-q_1} \sum_{t=q_1+1}^T (\hat{e}_t -\tilde{e}_t)^2  \\
  { =}& \frac{1}{T-q_1} \sum_{t=q_1+1}^T \left( \sum_{l=1}^{q_1} (\hat{\alpha}_l - \alpha_l) y_{t-l} + \bz_t^\top(\hat{\btheta} - \btheta) + \bx_t^\top(\hat{\bdelta} - \bdelta) + (\hat{\bgamma} - \bgamma)^\top \hbD(\by_t^S) \right)^2 \\
  \le& \frac{4}{T-q_1} \sum_{t=q_1+1}^T \Big\{ \Big( \sum_{l=1}^{q_1} (\hat{\alpha}_l - \alpha_l) y_{t-l} \Big)^2 + \left( \bz_t^\top(\hat{\btheta} - \btheta)  \right)^2 + \left( \bx_t^\top(\hat{\bdelta} - \bdelta)\right)^2 \\
  &\quad\quad\quad + \left((\hat{\bgamma} - \bgamma)^\top \hbD(\by_t^S) \right)^2 \Big\}\\
  =& O_p\left( {T^{-1} \zeta_T^{-2}} \sum_{t=1}^T y^2_{t} \right) + O_p\left( \zeta_T^{-2}\lambda_{\max,z} \right)+ O_p\left( \zeta_T^{-2}\lambda_{\max,x}\right) 
  + O_p\left({T^{-1} \zeta_T^{-4}}  \sum_{t=1}^T \|\by^S_{t} \|^2 \right)\\
  &+O_p \left( \zeta_T^{-2} \lambda_{\max,D}\right),\\
    \end{aligned}
\end{equation}
where the last step above is due to the $\zeta_T$-consistency of all the coefficient estimators and \eqref{equ:D_error}.

Moreover, all the terms on the right-hand side of \eqref{equ:june03:01} vanish asymptotically. Specifically, from Condition \eqref{con:lem2:2}, we see that the second, third, and last terms vanish in probability. For the first term, it follows from Condition \eqref{con:lem2:1} and the Cauchy--Schwarz inequality that 
\begin{equation*}
    \begin{aligned}
\E(\sum_{t=1}^T y^2_{t})^2
& \leq T \sum_{t=1}^T \E(y^4_t)  = O(T^2).
    \end{aligned}
\end{equation*}
Using this result and Markov's inequality, we can obtain that 
$$\sum_{t=1}^T y^2_{t} = O_p(T),$$ 
which entails $ T^{-1} \zeta_T^{-2} \sum_{t=1}^T y^2_{t}  = O_p(\zeta_T^{-2})=o_p(1)$. Similarly, we can show that 
\[\E(\sum_{t=1}^T \|\by^S_{t-l} \|^2)^2 = O\left(T \sum_{t=1}^T \E \left(\|\by^S_{t-l} \|^4 \right) \right) = O(T^2)\] 
by invoking Condition \eqref{con:lem2:1}, and thus $T^{-1}\zeta_T^{-4} \sum_{t=1}^T \|\by^S_{t-l} \|^2 = o_p(1)$. Hence, bound \eqref{lem2:goal1} is established.


We now proceed to prove bound \eqref{lem2:goal2}. By the $\zeta_T$-consistency of all the coefficient estimators and Conditions \eqref{con:lem2:1}--\eqref{con:lem2:2}, we can deduce that 
\begin{equation*}
   \begin{aligned}
  \frac{1}{T-q_1} \sum_{t=q_1+1}^T (\tilde{e}_t - e_t)^2= &\frac{1}{T-q_1} \sum_{t=q_1+1}^T (\Delta_t)^2\\
 =& \frac{1}{T-q_1} \sum_{t=q_1+1}^T  \left(\sum_{l=1}^{q_2} \bgamma^\top \left( \bA^S_{l} - \widehat{\bA}^S_{l} \right) \by^S_{t-l}\right)^2\\
 \le &\frac{q_2}{T-q_1} \sum_{t=q_1+1}^T \sum_{l=1}^{q_2} \left(  \bgamma^\top \left( \bA^S_{l} - \widehat{\bA}^S_{l} \right) \by^S_{t-l}\right)^2\\
 =& O_p\left(T^{-1} \zeta_T^{-2} \sum_{t=1}^T \| \by_{t}^S\|^2 \right) = o_p(1),
\end{aligned}
\end{equation*}
which yields bound \eqref{lem2:goal2}.

Finally, we establish the asymptotic coverage of the Box--Jenkins (BJ) prediction interval. First, note that
$$
\frac{y_{T+h} - \hat{y}_{T+h}}{\sqrt{\sum_{r=0}^{h-1}(\bA^r)^2_{11}} {\sigma}_e} \overset{d}{\to} N(0,1).
$$
Since all the coefficient estimators and the variance estimator $\hat{\sigma}_e^2$ are consistent, an application of Slutsky's theorem gives that 
$$
\frac{y_{T+h} - \hat{y}_{T+h}}{\sqrt{\sum_{r=0}^{h-1}(\hbA^r)^2_{11}} \hat{\sigma}_e} \overset{d}{\to} N(0,1).
$$
Therefore, using this fact, we can obtain that 
\begin{equation*}
   \begin{aligned}
\Pr\left\{ y_{T+h} \in \operatorname{PI}^{BJ}(\hat{y}_{T+h}) \right\} & = \Pr\left\{\left| \frac{y_{T+h} - \hat{y}_{T+h}}{\sqrt{\sum_{r=0}^{h-1}(\hbA^r)^2_{11}} \hat{\sigma}_e} \right|\le z_{\alpha/2} \right\} \\
&\to \Pr\left\{|N(0,1)| \le z_{\alpha/2}  \right\} = 1-\alpha,
\end{aligned}
\end{equation*}
which concludes the proof of Theorem \ref{lem:variance}.


\subsection{Proof of Theorem \ref{lem:variance.boot}} \label{sec.proof.lem:variance.boot}



Let $F$ be the distribution of random error $e_t$ (i.e., $F = N(0,\sigma_e^2)$), $\tilde{F}_n$ the empirical distribution of residuals $\tilde{e}_t$ with $t=q_1+1,\ldots, T$ given in \eqref{equ:apr17:01}, and $\hat{F}_n$ the empirical distribution of residuals  $\hat{e}_t - \hat{\mu}$ with  $t=q_1+1,\ldots,T$ given in \eqref{equ:residual}. Similar to \eqref{equ:march06:01}, we can rewrite the bootstrap sample as 
\begin{equation}\label{equ:apr15:equ1}
   \begin{aligned}
       Y^*_{T+h} =& \hbA Y^*_{T+h-1} + \hbB \bx_{T+h} + \hbTheta \bz_{T+h} +\widehat{\bGamma} \hbD(\by_{T+h}^S) + \bE^*_{T+h}\\
       =&\hbA^h Y^*_{T} + \sum_{r=0}^{h-1} \hbA^r\left\{ \hbB \bx_{T+h-r} + \hbTheta \bz_{T+h-r} + \hbGamma \hbD(\by_{T+h-r}^S)\right\} + \sum_{r=0}^{h-1} \hbA^r \bE^*_{T+h-r},
   \end{aligned} 
\end{equation}
where $Y^*_t = (y^*_{t}, \ldots, y^*_{t-q_1+1})^\top \in \R^{q_1}$ and $\bE^*_t = (e^*_{t}, 0,\ldots, 0)^\top \in \R^{q_1}$. The $h$-step-ahead prediction based on the bootstrap sample is given by 
\begin{equation}\label{equ:apr15:equ2}
 \begin{aligned}
 \hat{Y}^*_{T+h} =& \hbA^* \hat{Y}^*_{T+h-1} + \hbB^* \bx_{T+h} + \hbTheta^* \bz_{T+h} +\widehat{\boldsymbol{\Gamma}}^* \hbD(\by_{T+h}^S)\\
 =& (\hbA^*)^h \hat{Y}^*_{T} + \sum_{r=0}^{h-1} (\hbA^*)^r \left\{\hbB^* \bx_{T+h-r} + \widehat{\bTheta}^* \bz_{T+h-r}+\widehat{\boldsymbol{\Gamma}}^* \hbD(\by_{T+h-r}^S)\right\}, 
\end{aligned}   
\end{equation}
where $\hat{Y}^*_t = (\hat{y}^*_{t}, \ldots, \hat{y}^*_{t-q_1+1})^\top \in \R^{q_1}$ with $\hat{y}^*_t$ defined as $y^*_t$ for each $t \le T$ for the notational simplicity, and $\hbA^*$, $\hbB^*$, $\widehat{\bTheta}^*$, and $\widehat{\boldsymbol{\Gamma}}^*$ represent the bootstrap counterparts of the corresponding matrices. 

We can show that all the coefficient estimators based on the bootstrap sample are consistent with rate $\sqrt{\lambda_{\min,G}^{-1} \log(\lambda_{\max,G})} \to 0$. 
To this end, denote by 
$$\mathbf{g}^*_t := (y^*_{t-1},\ldots, y^*_{t-l}, \bz_t^\top, \bx_t^\top, (\hbD(\by_t^S))^\top )^\top \in \R^{q_1 + b + p + K}$$ 
the joint feature vector, and $\mathbf{G}^* := (\mathbf{g}^*_{q_1+1},\ldots, \mathbf{g}^*_{T})^\top  \in \R^{(T-q_1) \times (q_1 + b + p + K)}$ the bootstrap design matrix. Let $\lambda_{\max, G^*} = \lambda_{\max}((\mathbf{G}^*)^\top \mathbf{G}^*)$ and $\lambda_{\min, G^*} = \lambda_{\min}((\mathbf{G}^*)^\top \mathbf{G}^*)$ be the maximum and minimum eigenvalues of matrix $(\mathbf{G}^*)^\top \mathbf{G}^*$, respectively. Observe that conditional on the original sample, with randomness arising solely from the bootstrap resampling the random errors $e_t^*$ with $t=q_1+1,\ldots,T$ have zero mean.  To establish the consistency, it suffices to show that  $\lambda_{\max, G^*} \to \lambda_{\max, G}$ and $\lambda_{\min, G^*} \to \lambda_{\min, G}$ in probability. Then an application of Theorem 1 of \cite{lai1982least} guarantees the consistency of the bootstrap coefficient estimators with rate $\sqrt{\lambda_{\min,G}^{-1} \log(\lambda_{\max,G})}$.
To verify this, we apply Weyl's inequality.

Note that $\mathbf{g}_t := (y^*_{t-1}, \ldots, y^*_{t-q_1}, \bz_t^\top, \bx_t^\top, (\bD(\by_t^S))^\top )^\top \in \R^{q_1 + b + p + K}$ is the joint feature vector, and $\mathbf{G} := (\mathbf{g}_{q_1+1},\ldots, \mathbf{g}_{T})^\top  \in \R^{(T-q_1) \times (q_1 + b + p + K)}$ is the bootstrap design matrix. It can be seen that $\mathbf{g}^*_t = \mathbf{g}_t + \mathbf{r}_t$, where $\mathbf{r}_t$ is the residual vector with the first $q_1 + b + p$ components being zeros and the last $K$ components being $\hbD(\by_t^S) - \bD(\by_t^S)$. Let us define $\mathbf{R} = [\mathbf{r}_{q_1+1}, \ldots, \mathbf{r}_T]^\top \in \R^{(T-q_1) \times (q_1 + b + p + K)}$. Then it holds that 
$$\mathbf{G}^* = \mathbf{G} + \mathbf{G}^\top\mathbf{R} +  \mathbf{R}^\top\mathbf{G} + \mathbf{R}^\top \mathbf{R} := \mathbf{G} + \mathbf{\Delta}.$$ 
We can write
$$
\operatorname{Trace}(\mathbf{\Delta}) = 2 \operatorname{Trace}(\mathbf{R}^\top\mathbf{G}) + \operatorname{Trace}(\mathbf{R}^\top \mathbf{R}). 
$$
Using \eqref{equ:D_error}, we can deduce that 
$$
\begin{aligned}
\operatorname{Trace}(\mathbf{R}^\top\mathbf{G}) &= \operatorname{Trace}(\mathbf{G} \mathbf{R}^\top) = \sum_{t=q_1+1}^T \mathbf{g}_{t}^\top  \mathbf{r}_{t}\\
&=  \sum_{t=q_1+1}^T \bD(\by_t^S)^\top \left( \hbD(\by_t^S)- \bD(\by_t^S)^\top\right) \\
&= \sum_{t=q_1+1}^T \sum_{l=1}^{q_2} \bD(\by_t^S)^\top \left( \bA^S_{l} - \widehat{\bA}^S_{l} \right) \by^S_{t-l} \\
&= \sum_{l=1}^{q_2}  \operatorname{Trace}\left\{\left( \bA^S_{l} - \widehat{\bA}^S_{l} \right) \left(\sum_{t=q_1+1}^T \by^S_{t-l} \bD(\by_t^S)^\top \right)  \right\}\\
&:= \sum_{l=1}^{q_2} \operatorname{Trace}\left\{ \tilde{\bA}_l  \tilde{\bB}_l\right\},
\end{aligned}
$$
where $ \tilde{\bA}_l = \bA^S_{l} - \widehat{\bA}^S_{l} \in \R^{K \times K}$ and $\tilde{\bB}_l =  \sum_{t=q_1+1}^T \by^S_{t-l} \bD(\by_t^S)^\top \in \R^{K \times K}$. Furthermore, for each $(i,j) \in \{1,\ldots, K\} \times \{1,\ldots, K\}$, it holds that 
$$
\begin{aligned}
 \E( \tilde{\bB}_{l,i,j})^2 =& \E\left( \sum_{t=q_1+1}^T \by^S_{t-l,i} \bD_j(\by_t^S) \right)^2 \\
 \leq & 2T \sum_{t=1}^T\E\left(\| \by_{t}^S\|^4 \right) + 2T \sum_{t=q_1+1}^T\E\left(\| \bD(\by_{t}^S)\|^4 \right)  \\
\leq &  2T \sum_{t=1}^T\E\left(\| \by_{t}^S\|^4 \right) +2 T q_2^3\sum_{t=q_1+1}^T \sum_{l=1}^{q_2}\E\left(\| \bA_{l}^S \by_{t-l}^S\|^4 \right) \\
 =&O\left( T \sum_{t=1}^T\E\left(\| \by_{t}^S\|^4 \right)\right) = O(T^2).
\end{aligned}
$$
This entails that $\tilde{\bB}_{l,i,j} = O_p(T)$. 

Meanwhile, it follows from the $\zeta_T$-consistency of the coefficient estimators that $\tilde{\bA}_{l,i,j} = O_p(\zeta_T^{-1})$, and thus $(\tilde{\bA}_l  \tilde{\bB}_l)_{i,j} = O_p(T \zeta_T^{-1})$ for each $(i,j)$. Indeed, since $K$ and $q_2$ are fixed constants, this result holds uniformly for all $(i,j, l) \in \{1,\ldots,K\} \times \{1,\ldots, K\} \times \{1,\ldots,q_2\}$ by the union bound. Hence, we have 
$$
\operatorname{Trace}(\mathbf{R}^\top\mathbf{G}) = O_p(T \zeta_T^{-1}).
$$
An application of similar arguments as above leads to $\operatorname{Trace}(\mathbf{R}^\top\mathbf{R}) = O_p(T \zeta_T^{-1})$. Combining these results and invoking Weyl's inequality, we can obtain that 
\begin{equation}
\begin{aligned}
|\lambda_{\max, G^*} - \lambda_{\max, G}| & \le \lambda_{\max}(\mathbf{\Delta}) \le \operatorname{Trace}(\mathbf{\Delta}) \\
& = O_p(T \zeta_T^{-1}) =  o_p(\lambda_{\max, G}),
\end{aligned}
\end{equation}
where the last step follows from Condition \eqref{con:lem3:1}.
A similar bound also holds for $\lambda_{\min, G^*}$. In view of Condition \eqref{con:lem3:1}, the desired claim holds. 
Hence, an application of similar arguments as in the proof of Theorem \ref{lem:prediction} yields that 
$$
 y^*_{T+h} -  \hat{y}^*_{T+h} = \sum_{r=0}^{h-1} (\hbA^r)_{11} e^*_{T+h-r} + o_p(1) = \sum_{r=0}^{h-1} (\bA^r)_{11} \be^*_{T+h-r} + o_p(1),
$$
where the last step above is due to the consistency of the original coefficient estimators.

The convergence of the bootstrap prediction interval can be quantified by the Mallows metric \citep{bickel1981some,freedman1981bootstrapping}. Here, we use only the $\ell_2$-Mallows metric. Let  $U$ and $V$ be two random variables with distribution functions $F$ and $G$, respectively. The $\ell_2$-Mallows metric $d(F,G)$ (or written as $d(U,V)$) is defined as the infimum of $\mathbb{E}^{1/2}((U_1-V_1)^2)$ over all pairs of random variables $(U_1,V_1)$ with marginal laws $F$ and $G$. If we can show the claim 
\begin{equation}\label{equ:mallow}
    d\left(\sum_{r=0}^{h-1} (\bA^r)_{11} e^*_{T+h-r}, \sum_{r=0}^{h-1} (\bA^r)_{11} e_{T+h-r}\right) = o_p(1),
\end{equation}
an application of Lemma 8.2 of \cite{bickel1981some} leads to 
$${y}^*_{T+h} -  \hat{y}^*_{T+h} \overset{d}{\to} y_{T+h} -  \hat{y}_{T+h}.$$ 
Further, $\hat{q}^h_{\alpha/2}$ (respectively, $\hat{q}^h_{1-\alpha/2}$) is the $\alpha/2$ (respectively, $1-\alpha/2$) upper quantile of ${y}^*_{T+h} -  \hat{y}^*_{T+h}$, and $q^h_{\alpha/2}$ (respectively, $q^h_{1-\alpha/2}$) is the $\alpha/2$ (respectively, $1-\alpha/2$) upper quantile of $y_{T+h} -  \hat{y}_{T+h}$. Then we have that $\hat{q}^h_{\alpha/2} \overset{p}{\to} q^h_{\alpha/2}$ and $\hat{q}^h_{1-\alpha/2} \overset{p}{\to} q^h_{1-\alpha/2}$ for sufficient large $B$; see, e.g., Corollary 21.5 of \cite{van2000asymptotic}. A combination of these results gives that 
\begin{equation*}
   \begin{aligned}
&\Pr\left\{ y_{T+h} \in (\hat{y}_{T+h}+ \hat{q}^h_{\alpha/2}, \hat{y}_{T+h}+\hat{q}^h_{1-\alpha/2} ) \right\} \\
&\to \Pr\left\{ y_{T+h} \in (\hat{y}_{T+h}+ q^h_{\alpha/2}, \hat{y}_{T+h}+q^h_{1-\alpha/2} ) \right\} \\
&\to 1- \alpha.
\end{aligned}
\end{equation*}

It now remains to establish claim \eqref{equ:mallow}. The remaining proof is motivated by Lemma 2.1 of \cite{freedman1981bootstrapping}. With the aid of Lemmas 8.5 and 8.6 in \cite{bickel1981some}, we can show that 
\begin{equation}\label{equ:apr16:01}
    d\left(\sum_{r=0}^{h-1} (\bA^r)_{11} e^*_{T+h-r}, \sum_{r=0}^{h-1} (\bA^r)_{11} e_{T+h-r}\right) = \sum_{r=0}^{h-1}  (\bA^r)_{11} d\left(e^*_{T+h-r}, e_{T+h-r} \right).
\end{equation}
Moreover, for each $r = 0,\ldots, h-1$, the right-hand side of (\ref{equ:apr16:01}) above can be bounded by 
\begin{equation}\label{equ:apr16:02}
\begin{aligned}
 & (1/3) d\left(e^*_{T+h-r}, e_{T+h-r} \right)^2 =  (1/3) d\left(\hat{F}_n, F \right)^2 
 \\
 &\le d\left(\hat{F}_n, \tilde{F}_n \right)^2 + d\left(\tilde{F}_n, F_n \right)^2 + d\left(F_n, F \right)^2 \\
 &= o_p(1).   
\end{aligned}
\end{equation}
We provide more details on the last step above. Using \eqref{lem2:goal1} and $\hat{\mu} =\sum_{t=q_1 +1}^n e_t/(T-q_1) + o_p(1) =o_p(1)$, the first term on the right-hand side of (\ref{equ:apr16:02}) above can be bounded as 
 $$
 d\left(\hat{F}_n, \tilde{F}_n \right)^2 \le \frac{1}{T-q_1} \sum_{t=q_1+1}^T (\hat{e}_t - (\tilde{e}_t - \hat{\mu}))^2 \le \frac{2}{T-q_1} \sum_{t=q_1+1}^T (\hat{e}_t -\tilde{e}_t)^2 + 2\hat{\mu}^2 = o_p(1).
$$
In light of \eqref{lem2:goal2}, the second term on the right-hand side of (\ref{equ:apr16:02}) above can be bounded by
$$
\begin{aligned}
 &d\left(\tilde{F}_n, F \right)^2 \le \frac{1}{T-q_1} \sum_{t=q_1+1}^T (\tilde{e}_t - e_t)^2  = o_p(1).
\end{aligned}
$$
It follows from Lemma 8.4 of \cite{bickel1981some} that the third term on the right-hand side of (\ref{equ:apr16:02}) above satisfies $d\left(F_n, F \right)^2 = o_p(1)$. Therefore, combining \eqref{equ:apr16:01} and \eqref{equ:apr16:02} yields the desired claim \eqref{equ:mallow}. This completes the proof of Theorem \ref{lem:variance.boot}.

\section{Details on online text data collection and preprocessing} \label{new.Sec.textdatacoll}

In this section, we document our online text data collection protocol and the raw data preprocessing steps.

\subsection{Online text data collection} \label{new.Sec.textdatacoll.1}

Our data acquisition process utilizes the advanced search interface of Sina Weibo through a structured protocol that aims at guaranteeing both search coverage and precision. We begin by specifying a keyword lexicon consisting of $25$ price-related terms; see Section~\ref{Sec.textcollec} for the keyword lexicon. Such lexicon extends the price-related vocabulary from \cite{angelico2022can} by incorporating real estate specific terms to better capture the dynamics of Chinese housing market as it plays a significant role in shaping the consumption and savings expectations.

\begin{figure}[h]
    \centering
    \includegraphics[width=10cm]{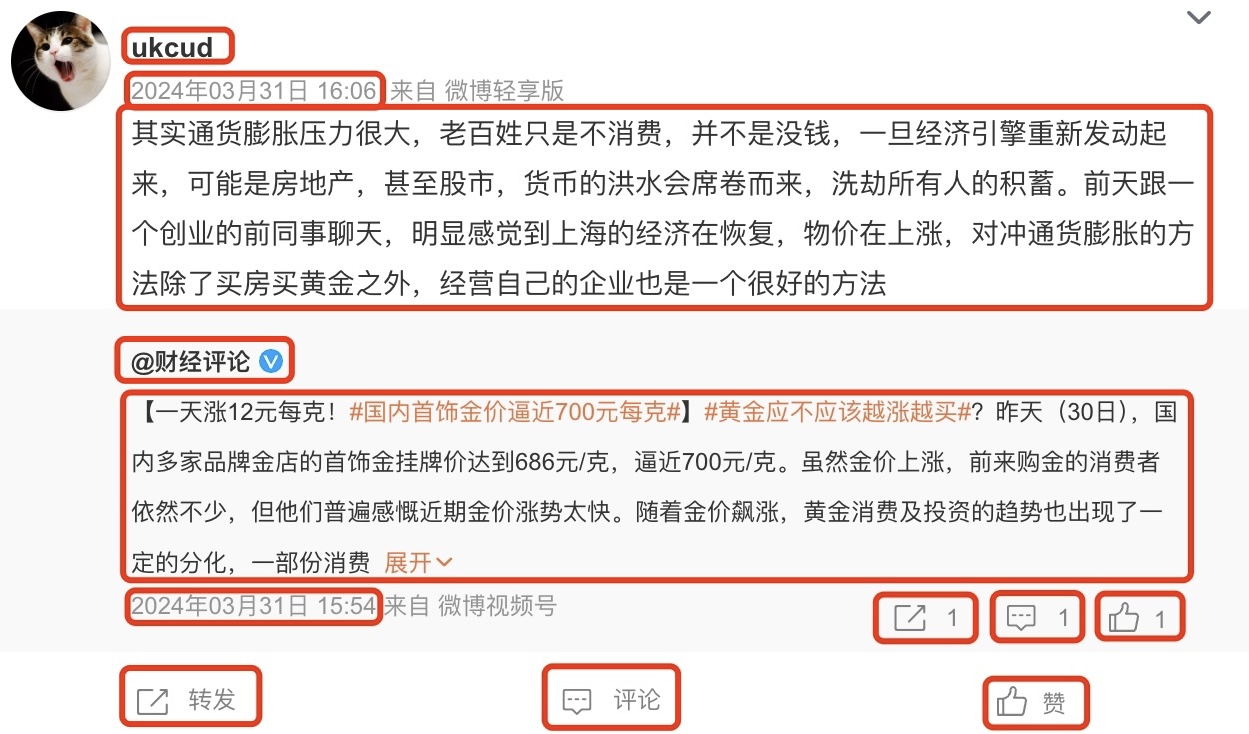}
    \caption{Snapshot of a Weibo post by the user ``ukcud,'' with the red rectangles indicating the areas that were crawled from the web.}
    \label{fig:weibo_sp}
\end{figure}

\begin{table}[t]
\centering
\small
\caption{Social media raw data statistics (2019--2023).}
\label{tab:social_media_stati}
\begin{tabular}{@{} l *{5}{S[table-format=10,group-four-digits = true]@{\hspace{10pt}}} @{}}
\toprule
\textbf{Year} & \textbf{2019} & \textbf{2020} & \textbf{2021} & \textbf{2022} & \textbf{2023} \\
\midrule
Total count        & 23867411  & 21478002  & 25055762  & 23558079  & 25898679  \\
Retweet count     & 3429038   & 3119749   & 3525874   & 3215841   & 3733858   \\
Unique users  & 4931671   & 4967343   & 4410642   & 4623454   & 4998838   \\
Unique hashtags & 1022506   & 1516800   & 1608734   & 1456434   & 1883884   \\
\midrule
Actions       &           &           &           &           &           \\
Likes       & 12839169800 & 20403910002 & 30258299263 & 20075445906 & 21861713953 \\
Retweets       & 8648416634 & 9356530796 & 19463395462 & 13335659391 & 14777487978 \\
Comments       & 3580772240 & 3616734933 & 6134589573 & 4715755638 & 4895875472 \\
\midrule
Keywords       &           &           &           &           &           \\
\quad House buying     & 1077183   & 829239    & 965077    & 830919    & 722778    \\
\quad Used house   & 99565     & 39877     & 75404     & 95167     & 92966     \\
\quad Price     & 2838008   & 2393859   & 2187561   & 3295425   & 2714885   \\
\quad Cheap     & 2238603   & 1978987   & 2147100   & 2473415   & 2543218   \\
\quad House selling     & 163358    & 98462     & 77197     & 78945     & 89563     \\
\quad Cost     & 1659565   & 2283366   & 1023571   & 1525205   & 1537220   \\
\quad House price     & 598757    & 508042    & 206019    & 412778    & 352712    \\
\quad Real estate    & 564631    & 639520    & 584665    & 880700    & 510788    \\
\quad House price (abbr.)  & 18322     & 23713     & 25921     & 25969     & 23023     \\
\quad Housing rent      & 316804    & 368030    & 255703    & 336635    & 281430    \\
\quad Mortgage     & 121491    & 191481    & 249711    & 225210    & 258122    \\
\quad New house     & 220649    & 159982    & 204317    & 259108    & 244952    \\
\quad Property market     & 248207    & 410944    & 296049    & 516941    & 211096    \\
\quad Oil price     & 109546    & 199619    & 112721    & 173054    & 88919     \\
\quad Rise/Increase       & 796561    & 2284562   & 4360591   & 1566195   & 3216657   \\
\quad Price rise     & 701838    & 619361    & 751806    & 636241    & 543230    \\
\quad Renting a house     & 939399    & 781312    & 321126    & 2210218   & 1678487   \\
\quad Rental fee     & 179092    & 271567    & 217554    & 257085    & 186540    \\
\quad Expensive       & 2343020   & 1796652   & 2367700   & 2132421   & 3024798   \\
\quad Fee     & 1550664   & 1090802   & 1222053   & 1072043   & 1374537   \\
\quad Decline/Decrease       & 1693393   & 1579567   & 3959107   & 1193505   & 2287394   \\
\quad Deflation  & 2097      & 3583      & 2399      & 2675      & 5555      \\
\quad Inflation  & 55822     & 68780     & 60048     & 98210     & 61676     \\
\quad Money       & 5002659   & 2560417   & 3080660   & 2955635   & 3380565   \\
\quad Price reduction     & 328163    & 296278    & 301702    & 304366    & 467568    \\
\bottomrule
\end{tabular}\\
{\footnotesize Note: ``Retweet'' indicates that the post includes content from another user's original post.}
\end{table}

We develop an adaptive temporal retrieval strategy to ensure comprehensive data collection within Weibo's page limits. The Weibo advanced search tool displays a maximum of $50$ pages of results for any keyword search. To overcome such limitation, we implement a step-by-step time-window approach to capture all relevant posts. The process starts with monthly searches for each keyword. If the results exceed the $50$-page limit, the search window is progressively narrowed to daily intervals, and further to hourly intervals if necessary, until all posts related to the keyword are retrieved (the displayed results are less than $50$ pages).

We also design several safeguard pipelines to ensure data integrity and prevent information leakage. These include validating metadata completeness by checking feature domains and implementing an error-logging system to retry failed requests, ensuring high data collection success.  Duplicate entries from overlapping keyword searches are removed after retrieval. Finally, the temporal validation checks ensure the continuity across the search windows. Such protocol resulted in a final corpus of approximately $119.8$ million posts.\footnote{Our data acquisition process strictly adheres to Weibo's Terms of Service, accessing only publicly available content without circumventing any access controls. We fully comply with the Weibo Service Agreement and did not collect any private user information.}

Table~\ref{tab:social_media_stati} provides the raw data statistics from January 2019 to December 2023, detailing the key metrics such as the total number of posts (Total count), the total number of retweets within the posts (Retweet count), the total number of unique users (Unique users), and the total number of unique hashtags (Unique hashtags). The total number of posts per year ranges from approximately $21.5$ million to $25.9$ million, with $10\%$ to $15\%$ being retweets of other users' posts. These retweets may include older posts from previous years. The number of unique users ranges from $4.4$ million to $4.9$ million, with each user posting an average of about five posts annually. Hashtags, identified by the symbol \textit{\#hashtag\#}, represent user-specified topics for posts. The total number of unique hashtags shows a consistent increase over the years.

The second part of Table~\ref{tab:social_media_stati} presents the total numbers of interactions for each Weibo post, including likes, retweets, and comments. A ``like'' indicates that a user has expressed approval of the post, a ``retweet'' means that the post has been shared by another user, and a ``comment'' refers to user responses below the post. On average, each post receives over hundreds of likes and retweets. However, the distribution of interactions exhibits a heavy-tailed pattern, where only a small proportion of posts gain significant attention, while the majority remain largely unnoticed. It is worth noting that the total number of ``retweets'' is significantly higher than that of retweeted posts within the same year. This discrepancy may occur because posts are often retweeted by users in subsequent years or are shared by posts that do not include the $25$ price-related keywords specified in our analysis.

The third part of Table~\ref{tab:social_media_stati} lists the count of posts collected for each keyword across the years. The most frequently occurring keywords are ``Price,'' ``Cheap,'' ``Cost,'' ``Expensive,'' ``Fee,'' and ``Money,'' reflecting general terms that are related to pricing. Posts containing house-related keywords also make up a significant proportion, highlighting the prominent role of real estate in the China market.

\subsection{Raw data preprocessing} \label{new.Sec.textdatacoll.2}


Our preprocessing pipeline involves several key steps to ensure the data quality and consistency. First, we remove posts with non-textual content, such as entries composed solely of numbers or symbols. We then exclude records lacking valid timestamps to avoid temporal incompleteness. Finally, we eliminate duplicate entries caused by the keyword-based web crawling process, which may capture the same posts multiple times occasionally. After these steps, the resulting text data set contains approximately $95.4$ million unique posts spanning the years of 2019 to 2023.

Despite the preprocessing, the obtained text data set still contains substantial noise from advertisements, E-commerce promotions, and unrelated posts. To reduce this, we employ an LLM-based framework to identify inflation-related content, as detailed in the next section.


\section{Implementation details of LLM-generated daily inflation index} \label{new.Sec.implemdetail}

This section provides the detailed steps for constructing the LLM-generated daily inflation index using the collected Weibo text data set mentioned in Section \ref{new.Sec.textdatacoll}.

\subsection{Noise reduction and high-frequency inflation measurement generation via LLMs} 
\label{new.Sec.inflation}

We introduce an LLM-based framework for text noise reduction and the LLM-generated daily inflation index construction, as summarized in Algorithm \ref{algo1}. 
To this end, we exploit the ChatGPT and a pre-trained BERT language model with fine-tuning for text noise reduction and the prediction tasks. Specifically, we build upon the BERT framework using the Chinese-optimized variant ``bert-base-chinese''\footnote{\url{https://huggingface.co/google-bert/bert-base-chinese}}, which is pre-trained on large-scale Chinese corpora.
This model features a $12$-layer Transformer architecture with multi-head self-attention mechanisms and feed-forward sublayers, containing approximately $110$ million parameters. Such large language model enables us to extract rich contextual features from Chinese text. 

To further fine-tune the BERT model, we collect a random sample of $S_1 = 20,000$ posts proportionally based on the keyword frequency and temporal distribution, denoted as set $\mathcal{S}_1$. We annotate these $S_1$ posts with high-quality labels with the aid of GPT-4 (GPT-4-turbo-2024-04-09). 

\begin{algorithm}[t]
\caption{Constructing the LLM-generated daily inflation index}\label{algo1}
\begin{algorithmic}[1]

\State \textbf{Input:} The preprocessed Weibo text data set.
\State \textbf{Output:} The LLM-generated daily inflation index.

\State \textbf{Step 1: Sampling}
\State Randomly sample a subset of $20,000$ posts from the complete data set.

\State \textbf{Step 2: Annotation}
\For{each post in the sampled subset}
    \State Use the \textit{chain-of-thought prompting} with GPT-4 to annotate the post:
    \State \hspace{1em} (a) Determine whether the post is an advertisement;
    \State \hspace{1em} (b) If not an advertisement, check whether it is related to inflation;
    \State \hspace{1em} (c) If related to inflation, assess the continuous degree of inflation it represents.
\EndFor

\State \textbf{Step 3: LLMs with fine-tuning}
\State Fine-tune the following BERT models using the annotated data set:
\State \hspace{1em} (a) \textbf{Advertisement-BERT:} Classify posts as advertisements or not;
\State \hspace{1em} (b) \textbf{Category-BERT:} Identify whether non-advertisement posts are related to inflation;
\State \hspace{1em} (c) \textbf{CPI-BERT:} Estimate the continuous degree of inflation for inflation-related posts.

\State \textbf{Step 4: Prediction}
\State Apply the fine-tuned Advertisement-BERT and Category-BERT models to filter out noise from the collected Weibo posts, and the fine-tuned CPI-BERT model to predict the continuous degree of inflation.
\State 
Compute the LLM-generated daily inflation index by Equation \eqref{equ:CPI_daily_inflation}.

\end{algorithmic}
\end{algorithm}

\textbf{High-quality annotation via GPT-4.} We implement the \textit{chain-of-thought prompting} strategy \citep{wei2022chain} to accurately annotate inflation-related posts, and precisely score the continuous inflation sentiment for each post. The chain-of-thought methodology emulates human reasoning by guiding the model through intermediate logical steps, thereby reducing semantic ambiguity in complex annotation tasks. The LLM prompts used for our labeling tasks are designed as follows: 


1). \textit{Your task is to make the best judgment based on the text I provide. Please judge whether the content of the text is an advertisement or not; if it is then output value $1$, and if not output value $0$. You only need to output a specific judgment number $0$ or $1$, and other text does not need an output. The content of the text is [\{text\}].}

2). \textit{Your task is to make the best judgment based on the text I provide.
Please determine whether the text content anticipates future inflation or deflation, and assign a score ranging from 0 to 1, where 1 indicates absolute inflation and 0 indicates absolute deflation.
For irrelevant information, classify it into one of the categories [Lifestyle, Entertainment, Emotion, Workplace, Socializing, News], or provide a category theme you deem appropriate.
Please make your judgment by comprehensively considering aspects such as consumption, savings, interest rates, and real estate market conditions mentioned in the text. You only need to output a specific score or category; no additional text is required.
The text content is [\{text\}].}

Based on prompt 1 above, GPT-4 assigns a preliminary binary label $\tilde{A}_{i} \in \{0, 1\}$ for each post, where $\tilde{A}_{i} = 1$ indicates that the $i$th post is classified as advertising content, and $\tilde{A}_{i} = 0$ denotes a non-advertising post. To ensure labeling accuracy and completeness, all preliminary binary labels $\tilde{A}_{i}$'s are manually reviewed. Then we obtain the final binary advertising labels denoted as $A_i \in \{0, 1\}$, $i =1,\ldots, S_1$. This results in a total of $S_2 = 13,970$ non-advertising posts (classified as $A_i=0$), collected in set $\mathcal{S}_2$.

Prompt 2 above for GPT-4 further classifies the $S_2$ non-advertising posts in set $\mathcal{S}_2$. Each post in this subset is initially assigned a category label $\tilde{C}_i$. If a post is preliminarily classified under the Inflation category, it is further evaluated and assigned a continuous inflation sentiment score $I_i \in [0,1]$. To enhance the interpretability and ensure sufficient sample sizes across categories, we manually consolidate infrequent categories, defined as those with fewer than $1000$ posts, into the most semantically appropriate groups. This yields five final content categories with labels $C_i \in \{\text{Inflation, Lifestyle, Entertainment, Emotion, News}\}$. We end up with a set $\mathcal{S}_3$ of $S_3 = 1576$ $\text{inflation}$ class posts (classified as $C_i = \text{Inflation}$), each with a continuous inflation sentiment score $I_i \in [0, 1]$. 

\begin{figure}[t]
    \centering
    \includegraphics[width=7cm]{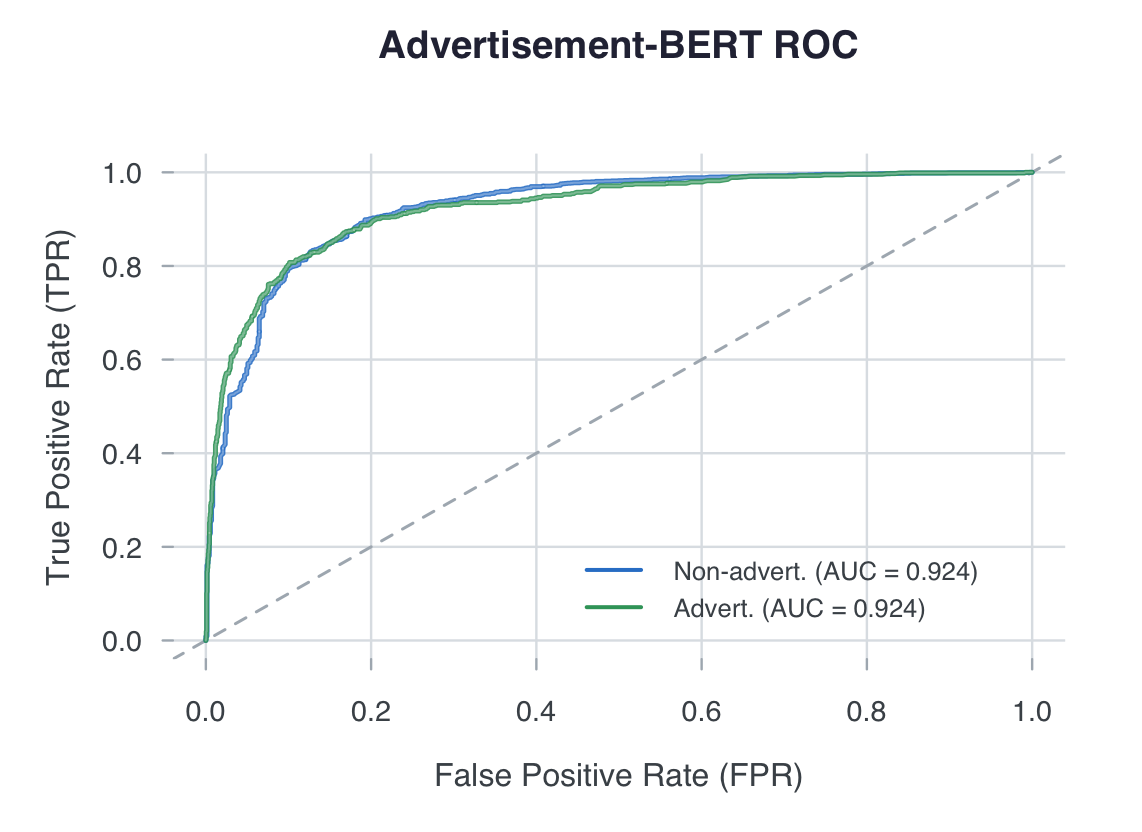}
    \includegraphics[width=7cm]{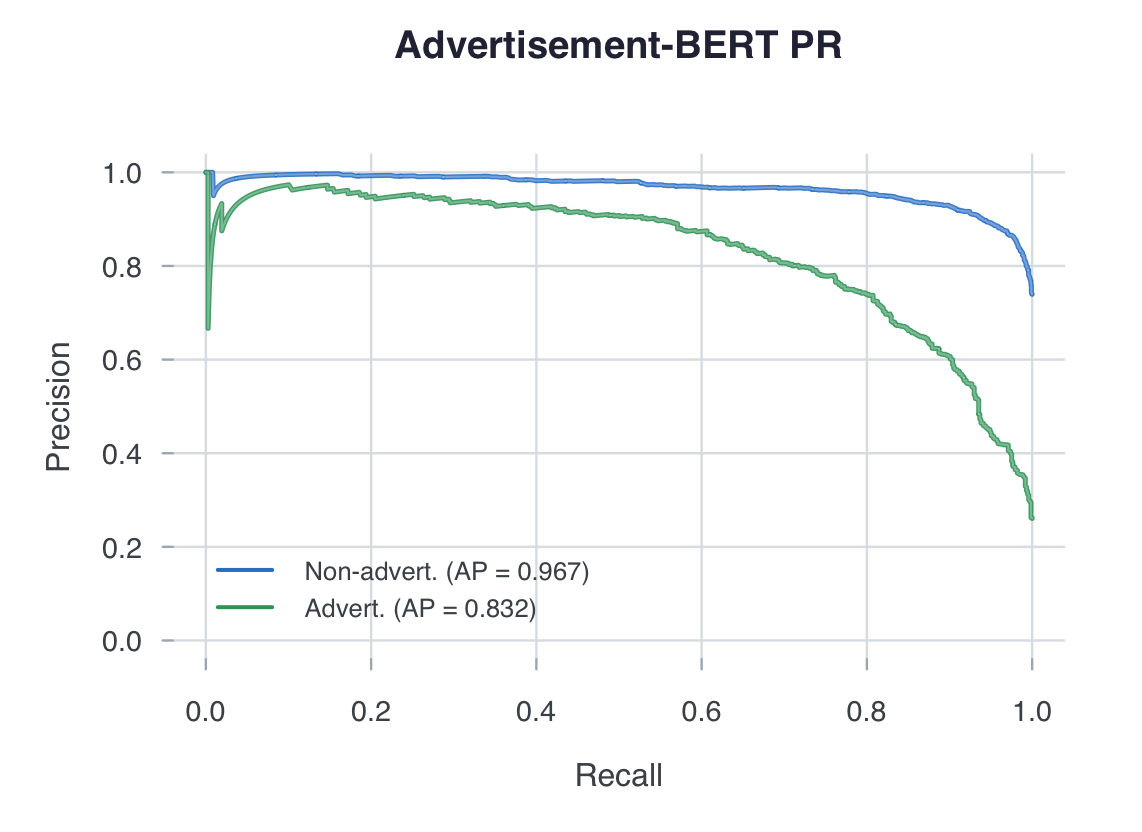}
    \caption{The performance measures of the fine-tuned Advertisement-BERT model.}
    \label{fig:bert_model1}
\end{figure}

\begin{figure}[t]
    \centering
    \includegraphics[width=7cm]{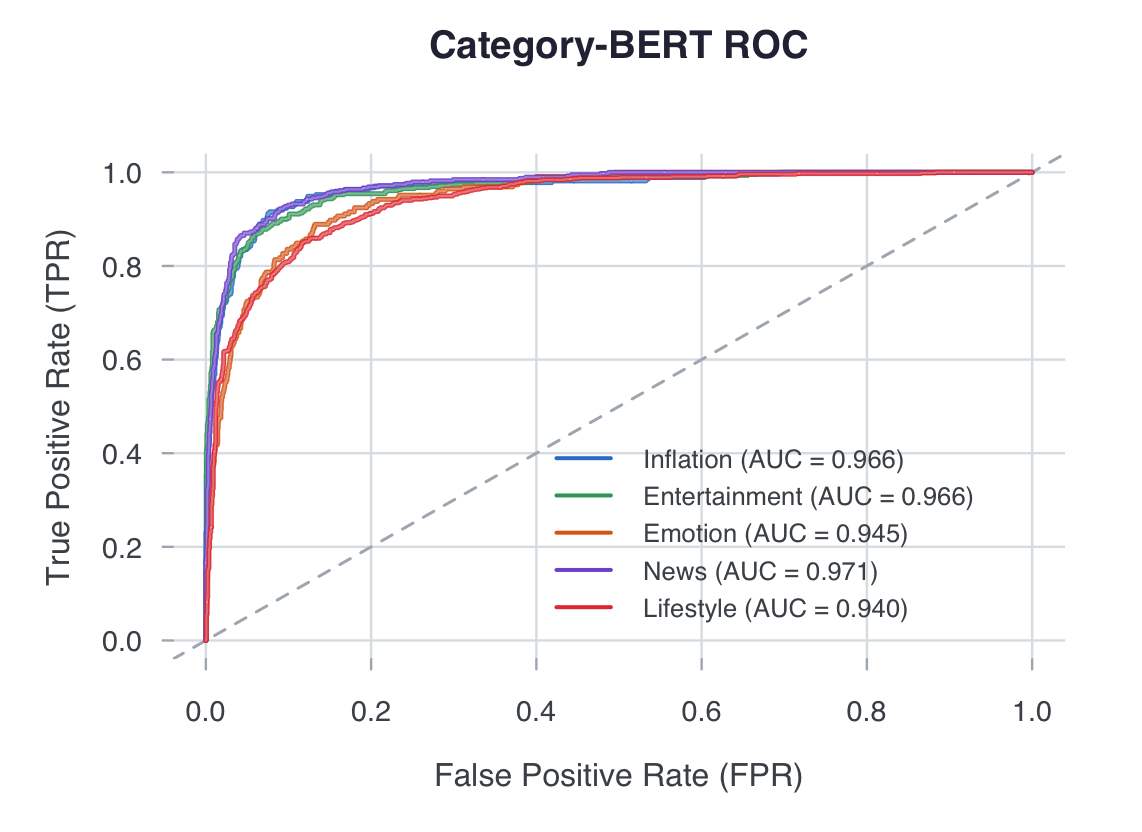}
    \includegraphics[width=7cm]{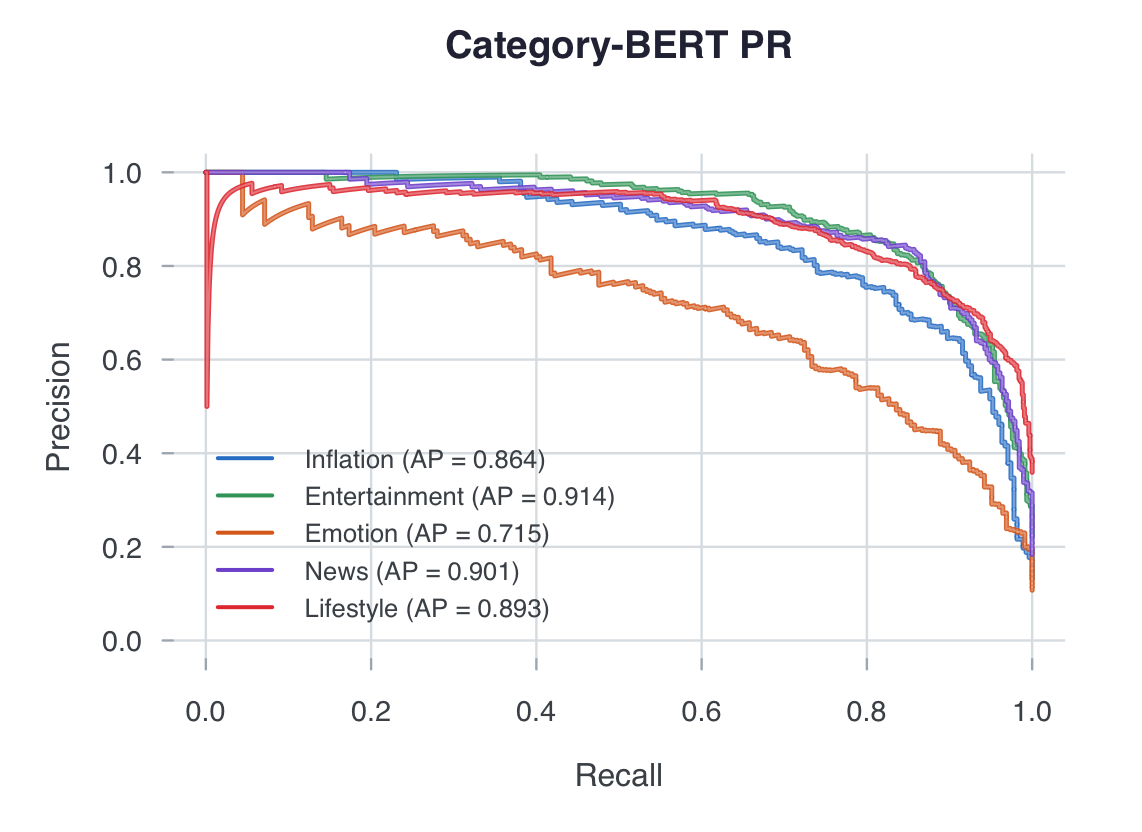}
    \caption{The performance measures of the fine-tuned Category-BERT model.}
    \label{fig:bert_model2}
\end{figure}

\begin{figure}[t]
    \centering
    \includegraphics[width=7cm]{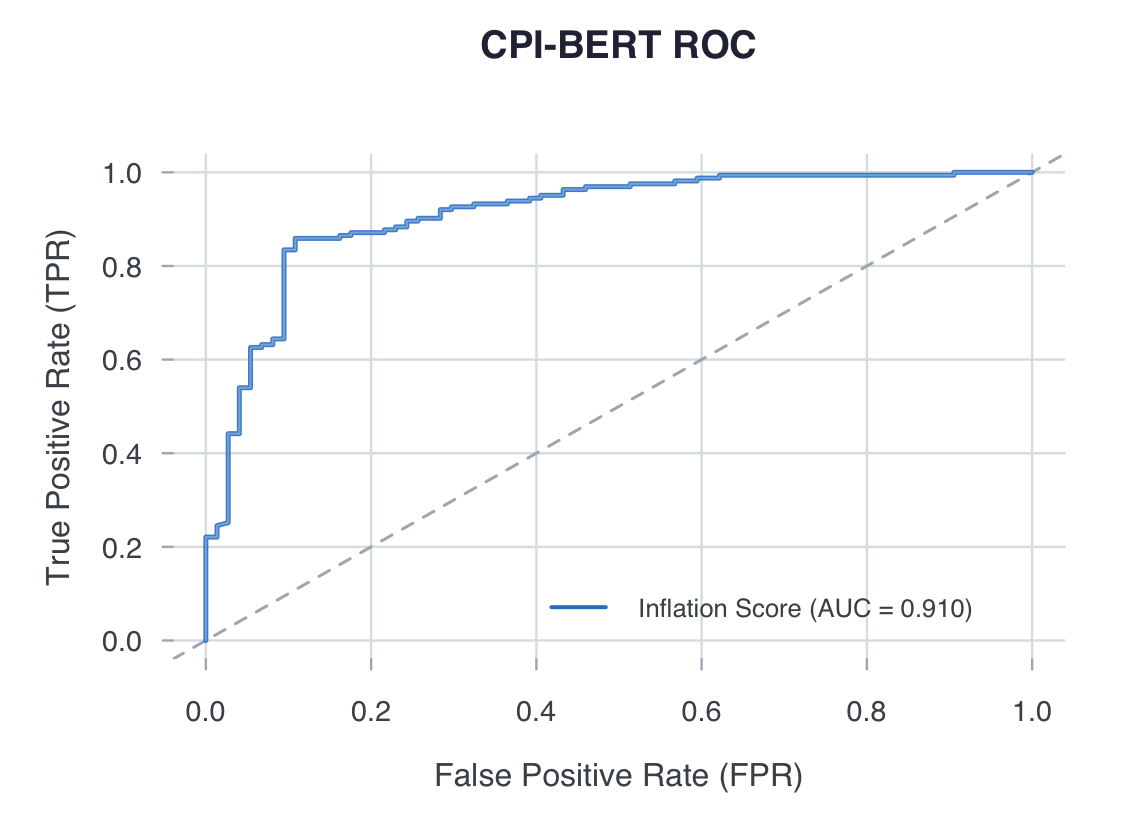}
    \includegraphics[width=7cm]{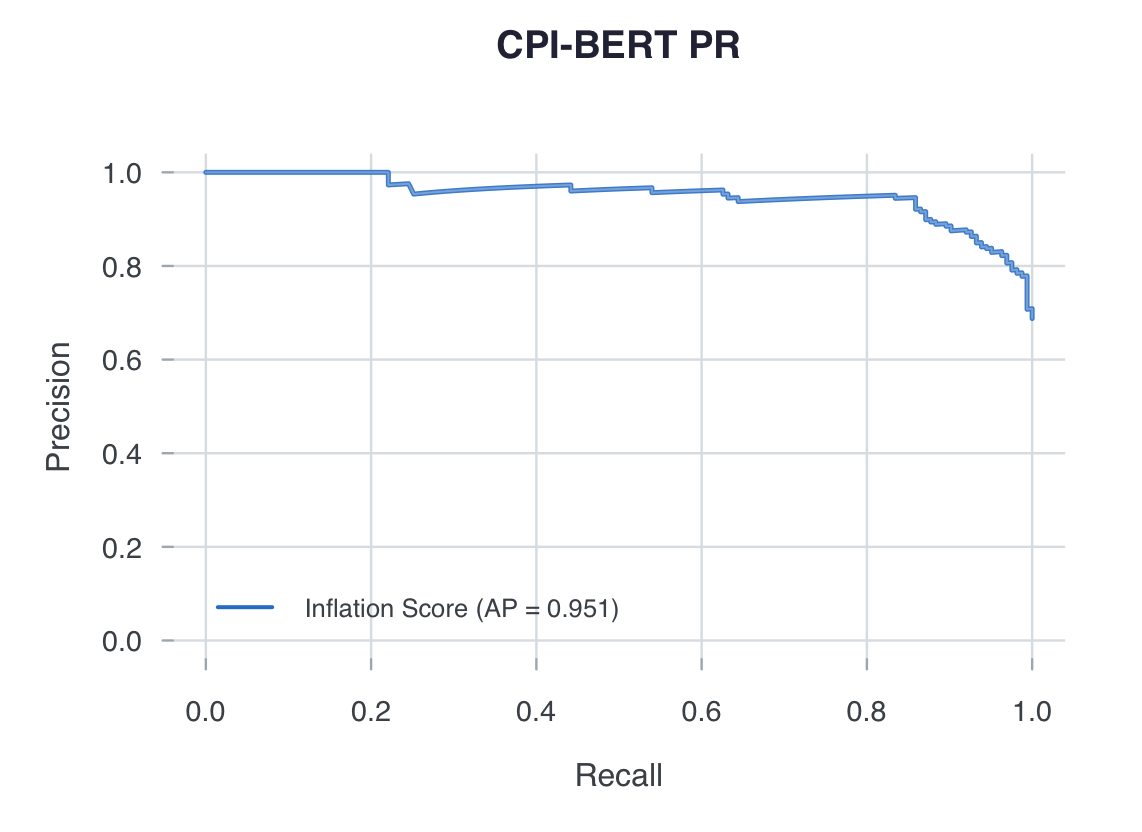}
    \caption{The performance measures of the fine-tuned CPI-BERT model.}
    \label{fig:bert_model3}
\end{figure}

\textbf{Identify and score inflation-related text with fine-tuned BERT models.} Using the annotated posts by GPT-4 as explained above, we further fine-tune three specialized BERT models to mimic the decision process of GPT-4: i) \textit{Advertisement-BERT} filters out the promotional and advertising posts, ii) \textit{Category-BERT} classifies the remaining non-advertising content to identify the inflation-related posts, and iii) \textit{CPI-BERT} scores the continuous inflation index of inflation-related posts. 

For each of the above three fine-tuned BERT models, the labeled (i.e., annotated) data is randomly partitioned into the training ($70\%$), validation ($15\%$), and testing ($15\%$) sets, and model performances are assessed on the testing sets using two evaluation metrics: the area under the receiver operating characteristic curve (AUC) and the average precision (AP).
The Advertisement-BERT is fine-tuned on set $\mathcal{S}_1$ with the loss function specified as the Softmax 
function. The testing AUC of the fine-tuned Advertisement-BERT is $0.924$ for both advertisement and non-advertisement classes. The corresponding testing AP scores are $0.832$ and $0.967$, respectively, as shown in Figure~\ref{fig:bert_model1}. The Category-BERT is fine-tuned on set $\mathcal{S}_2$ with the loss function also specified as the Softmax function. The testing AUC scores of the fine-tuned Category-BERT across the five categories (Inflation, Entertainment, Lifestyle, Emotion, News) are $0.966, 0.966, 0.940, 0.945$, and $0.971$, respectively, with the corresponding AP scores of $0.864, 0.914, 0.893, 0.715$, and $0.901$, as depicted in Figure~\ref{fig:bert_model2}. For the CPI-BERT, the loss function is chosen as the squared loss, and it is fine-tuned on set $\mathcal{S}_3$ with a continuous label of inflation score $I_i \in [0, 1]$.  To evaluate the model's ability to distinguish between textual indicators of inflation and deflation, we dichotomize the $ \operatorname{Score}_i, i \in \mathcal{S}_3$ into binary indicators with domain $\{0, 1\}$ using a threshold of 0.5, where 1 denotes inflation and 0 indicates deflation. Consequently, the testing AUC of the fine-tuned CPI-BERT is $0.910$ and the corresponding testing AP score is $0.951$, as displayed in Figure~\ref{fig:bert_model3}. In summary, the three fine-tuned specialized BERT models exhibit strong predictive performances across different tasks. 




We are now ready to apply the fine-tuned Advertisement-BERT model to the entire preprocessed Weibo text data set ($95,450,620$ posts), resulting in $70,743,705$ non-advertising posts. We next apply the fine-tuned Category-BERT model to the identified non-advertising posts, and assign a content category to each post. There are $5,790,457$ posts categorized under the \textit{Inflation} category, $23,884,070$ posts under the \textit{Lifestyle} category, $17,421,043$ posts under the \textit{Entertainment} category, $9,596,074$ posts under the \textit{Emotion} category, and $14,052,061$ posts under the \textit{News} category.
Finally, we apply the fine-tuned CPI-BERT model to the inflation-related posts. Such LLM assigns a continuous inflation sentiment score $\text{Score}_i \in [0, 1]$ to each post categorized under the \textit{Inflation} class. 

With the aid of the above three fine-tuned specialized BERT models, we can introduce our high-frequency LLM-generated inflation scores taking values in $[0, 1]$. By combining these scores with the posting dates, the final LLM-generated inflation results are represented as $\{(\text{Score}_i, \text{Date}_i): i = 1, \ldots, N\}$ with $N=5,790,457$, where $\text{Score}_i \in [0, 1]$ denotes the continuous LLM-generated inflation score and $\text{Date}_i$ represents the posting date of the $i$th post. Based on Equation \eqref{equ:CPI_daily_inflation}, we can finally calculate our LLM-generated daily inflation index. 



\begin{figure}[t]
    \centering
    \begin{subfigure}[b]{0.8\textwidth}  
        \includegraphics[width=\linewidth]{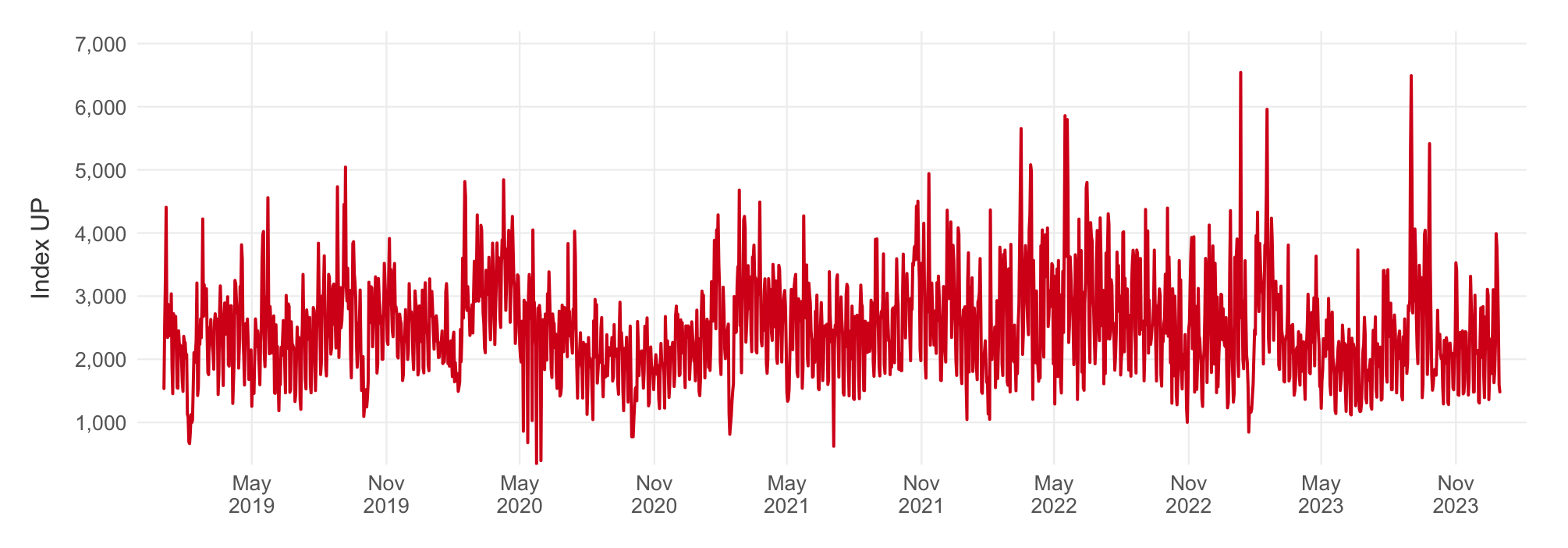}
        \caption{Daily count of Inflation-UP related posts on Weibo (2019--2023).} 
        \label{fig:index_up}
    \end{subfigure}
    \hspace{0.5cm}
    \begin{subfigure}[b]{0.8\textwidth}
        \includegraphics[width=\linewidth]{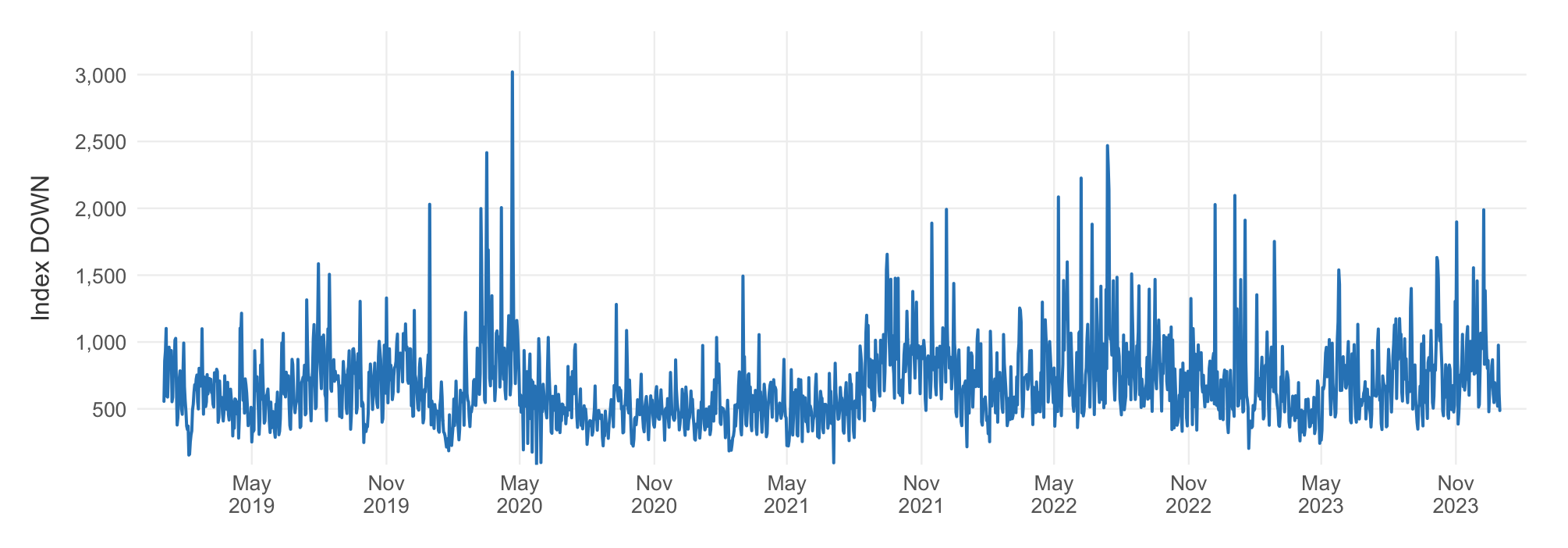}
        \caption{Daily count of Inflation-DOWN related posts on Weibo (2019--2023).} 
        \label{fig:index_down}
    \end{subfigure}
 \caption{Daily counts of inflation-increase versus inflation-decrease discussions on Weibo (January 2019 to December 2023).}
    \label{fig:indexup_down}
\end{figure}

\begin{figure}[t]
    \centering
    \includegraphics[width=10cm]{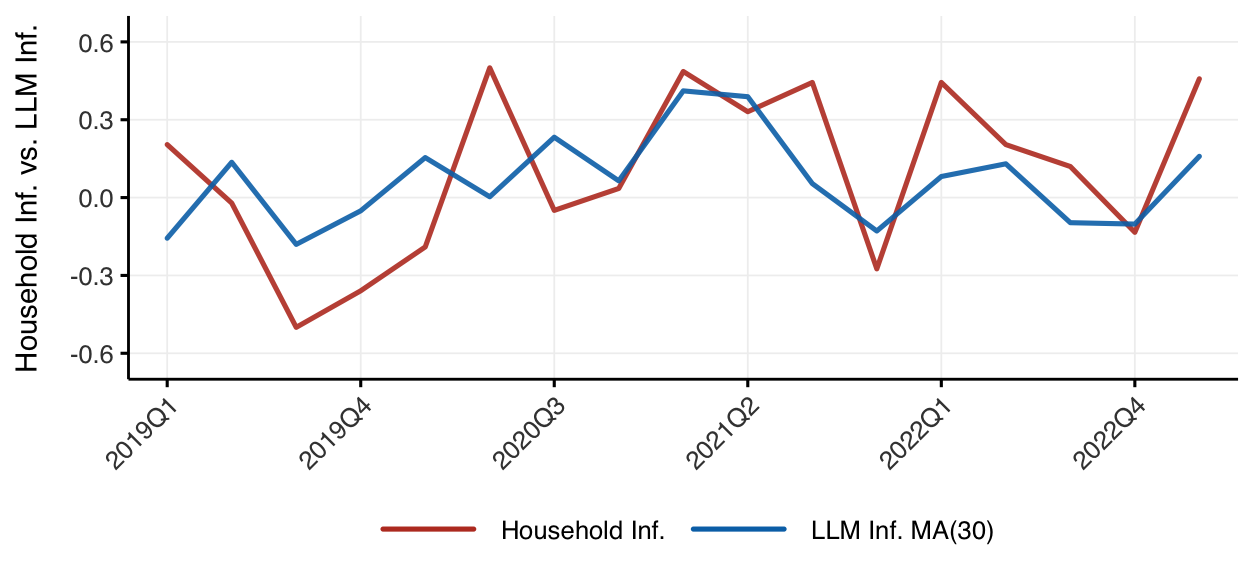}
    \caption{The quarterly household economic expectations (red curve) is the normalized index from surveys ($0$ for neutral, $\pm 1$ for extreme pessimism/optimism). The quarterly LLM-generated inflation index (blue curve) is the aggregated score from the LLM-generated daily inflation index, smoothed via the $30$-day moving average.}
    \label{fig:household_llm}
\end{figure}

\subsection{LLM-generated daily inflation fluctuation} \label{new.Sec.LLM-geninflfluc}

We examine the LLM-generated daily inflation fluctuation based on $\{(\Score_i, \operatorname{Date}_i): i=1,\ldots, N\}$ in this subsection. We first construct the Chinese LLM-generated daily inflation fluctuation index, similar to the approach in \cite{angelico2022can}, who studied a comparable index for Italy. We set a threshold of $0.5$ for the LLM-generated inflation score: if a post's score exceeds $0.5$, it indicates an Index UP; otherwise, it indicates an Index DOWN. Figure~\ref{fig:indexup_down} depicts the daily counts of Index UP and Index DOWN, respectively. Such index effectively captures the daily inflation fluctuations. For example, during the lockdown periods (e.g., COVID-19 in early 2020), the daily count of Inflation-UP posts dominates, whereas discussions about the inflation decreases rise during the post-lockdown recovery phases.

We also compare our LLM-generated inflation index \eqref{equ:CPI_daily_inflation} that is derived from online text data to the quarterly household expectations about future economic prospects based on the survey data obtained from the People's Bank of China (\url{http://www.pbc.gov.cn}). To process the LLM-generated daily inflation index, we first apply a $30$-day moving average and then aggregate it into the quarterly level by taking the average over three months. Figure~\ref{fig:household_llm} presents these two quarterly time series, revealing broadly synchronized cyclical patterns over the observed time period. Notably, there is significant co-movement between the two during exogenous economic shocks, such as the pandemic period. This alignment suggests that public economic expectations regarding inflation can be effectively extracted from unstructured online text data (e.g., Weibo posts) using LLMs.

\subsection{Online text embeddings}
\label{app:sec:embeddings}

\textbf{LDA embedding.} We exploit the popular tool of the latent Dirichlet allocation (LDA) \citep{blei2003latent} for constructing embeddings of the online posts. 
Such method has been successfully used for forecasting the U.S. inflation \citep{hong2025forecasting}. In the LDA framework, each post from the Weibo platform is assumed to be generated from a latent distribution over a list of topics, and each topic is characterized by a latent distribution over the vocabulary of words. These distributions are not directly observed but can be inferred from the text data. The LDA produces two key outputs. One of them is the post-topic distribution matrix $\mathbf{P} = (\mathbf{p}_1, \ldots, \mathbf{p}_N)^\top \in \R^{N \times D}$, where $N$ is the total number of posts and $D$ is the predefined number of topics. Each row vector $\mathbf{p}_i \in \R^{D}$ represents the topic distribution of the $i$th post and lies on a probability simplex. The other one is the topic-word distribution matrix $\mathbf{T} = (\mathbf{t}_1, \ldots, \mathbf{t}_V) \in \R^{D \times V}$, where $V$ is the size of (unique) vocabulary words. Each column $\mathbf{t}_d \in \R^{V}$ stands for the vocabulary distribution for the $d$th topic. The input to the LDA model is a Document-Term Matrix (DTM) of the post data. The LDA maps these high-dimensional representations to a lower-dimensional topic space. The topic distribution vector $\mathbf{p}_i$ serves as the embedding of the $i$th post in such topic space. The optimal number of topics is given by $D = 20$ in our text analysis, which is selected by minimizing the perplexity criterion \citep{blei2003latent}. 

\textbf{BERT embedding.} We also construct the BERT embeddings of the online posts by utilizing the fine-tuned CPI-BERT model. Given a predefined batch size $L$, we divide the entire data set of $N$ posts into approximately $[N/L] + 1$ mini-batches, denoted as $\operatorname{Batch}_k$ with $k = 1,\ldots, [N/L] + 1$. The final batch may contain fewer than $L$ posts if $N$ cannot be divided by $L$. For each $\operatorname{Batch}_k$, we extract the final hidden layer right before the output layer of the fine-tuned CPI-BERT model (i.e., a deep neural network). As a result, from each batch we obtain a sequence of matrices $[\mathbf{W}^{(k)}_1,\ldots,\mathbf{W}^{(k)}_l,\ldots, \mathbf{W}^{(k)}_L]$, where each $\mathbf{W}^{(k)}_l \in \R^{768 \times q_l}$ corresponds to the collection of token-level embeddings for the $l$th post in the $k$th batch, and $q_l$ is the number of tokens in that post. Each column of $\mathbf{W}^{(k)}_l$ represents the $768$-dimensional embedding of a specific word. To obtain a single embedding for each post, we calculate the mean pooling over all token embeddings in the post; that is, we take the average of all columns of $\mathbf{W}^{(k)}_l$. The resulting $768$-dimensional vector serves as the BERT embedding for that post based on the fine-tuned CPI-BERT model.

\textbf{Monthly aggregation.}
We further incorporate the timestamp of each post to generate a monthly aggregated representation of the online posts. Specifically, we aggregate all post embeddings within the same month by calculating their average. This results in a single embedding per month, which serves as the representation for the economic narrative index for that month. We denote by $\bx^{\text{LDA}}_t$ the monthly LDA embedding of the online text and $\bx^{\text{BERT}}_t$ the monthly BERT embedding of the online text for $t =1,\ldots, T$.

\subsection{Time-series model selection}
\label{app:sec:model_selection}


The online text data is divided into a training sample and a testing sample, where the training sample is of size $T_1$ and the testing sample is of size $T_2$ with $T_1 + T_2 = T$. Model selection is performed on the training sample using a combination of the correlation pursuit method \citep{zhong2012correlation} and the corrected Akaike information criterion (AIC) for time series models \citep{hurvich1989regression}. 
In particular, the correlation pursuit method offers an efficient alternative to forward selection, as discussed in \cite{borboudakis2019forward}.

Initially, an AR model specified in Equation \eqref{equ:pure_AR} is fitted, with the optimal lag order selected by minimizing the AIC. The residuals from such fitted AR model are calculated and denoted as $\hat{\epsilon}^{(0)}_t$ with $t=1,\ldots, T_1$. The $p$ latent embedding features are ranked by the absolute values of their correlations with the residuals, giving rise to the ranked latent embedding features $\bx_{(1)},\ldots, \bx_{(p)}$. Feature $\bx_{(1)}$ with the highest correlation is included in the model. We then calculate the residuals $\hat{\epsilon}^{(1)}_t$ with $t=1,\ldots, T_1$ and the (corrected) AIC 
$$
\operatorname{AIC}^{(m)} = T_1 \log\left(\sum_{t=1}^{T_1} (\hat{\epsilon}^{(m)}_t)^2/T_1+1\right) + 2\operatorname{Pen}(m),
$$
where $\operatorname{Pen}(m)= ((m+1)(m+2))/(T_1-m-2)$ and $m=1$ for the current step. Such process is iterated by sequentially adding the next most correlated latent feature, refitting the model, and recalculating the $\operatorname{AIC}^{(m)}$ at each step. The procedure continues as long as the $\operatorname{AIC}^{(m)}$ decreases sufficiently, and terminates when the inclusion of an additional feature no longer improves it, yielding the final set of selected latent features. Other model selection criteria can be invoked in combination with our LLM-CPI framework; see, e.g., \cite{FanTang2013,LvLiu2014}.

We emphasize that the embedding features in both the target CPI model (Equation \ref{equ:cpi}) and the surrogate model (Equation \eqref{equ:sur}) are set to be identical. This choice enables us to pinpoint the marginal accuracy increment of the LLM-CPI framework. Otherwise, such improvement could also be attributed to the inclusion of additional covariates in addition to the LLM-powered joint time series modeling.

\begin{table}[t]
\centering
\small
\caption{The $\operatorname{rPMSE}^{AR}_m(H)$ results across different prediction steps $H$ and correlation levels $\rho$ under the omitted relevant predictor setting.}
\label{tab:simu_miss_mse}
\begin{tabular}{cccccccccc}
\toprule
Method & 8 & 9 & 10 & 11 & 12 & 13 & 14 & 15 & Ave. \\
\midrule
\multicolumn{10}{c}{$\rho = 0.1$} \\
\midrule
RW       & 0.768 & 1.174 & 0.802 & 0.857 & 2.073 & 3.556 & 14.661 & 19.358 & 5.781 \\
AVE      & 0.686 & 1.102 & 2.327 & 4.933 & 7.849 & 9.188 & 10.753 & 8.133 & 5.371 \\
LLM-CPI  & 0.393 & 0.386 & 0.407 & 0.451 & 0.548 & 0.494 & 0.527 & 0.663 & 0.484 \\
\midrule
\multicolumn{10}{c}{$\rho = 0.2$} \\
\midrule
RW       & 0.761 & 1.164 & 0.798 & 0.871 & 2.152 & 3.638 & 15.100 & 19.845 & 5.791 \\
AVE      & 0.677 & 1.130 & 2.397 & 5.084 & 8.082 & 9.416 & 10.999 & 8.199 & 5.748 \\
LLM-CPI  & 0.382 & 0.369 & 0.393 & 0.437 & 0.539 & 0.484 & 0.515 & 0.663 & 0.472 \\
\midrule
\multicolumn{10}{c}{$\rho = 0.3$} \\
\midrule
RW       & 0.761 & 1.171 & 0.795 & 0.833 & 2.096 & 3.721 & 15.065 & 19.784 & 5.691 \\
AVE      & 0.675 & 1.115 & 2.366 & 5.010 & 8.044 & 9.488 & 10.987 & 8.218 & 5.738 \\
LLM-CPI  & 0.379 & 0.366 & 0.389 & 0.432 & 0.522 & 0.478 & 0.508 & 0.652 & 0.466 \\
\midrule
\multicolumn{10}{c}{$\rho = 0.4$} \\
\midrule
RW       & 0.759 & 1.173 & 0.796 & 0.855 & 2.082 & 3.606 & 15.014 & 19.751 & 5.754 \\
AVE      & 0.676 & 1.120 & 2.380 & 5.058 & 8.023 & 9.376 & 10.989 & 8.217 & 5.730 \\
LLM-CPI  & 0.369 & 0.364 & 0.396 & 0.442 & 0.527 & 0.478 & 0.509 & 0.641 & 0.466 \\
\bottomrule
\end{tabular}\\
{\footnotesize Note: The relative PMSE values compared to the AR benchmark. Smaller values indicate better performance.}
\end{table}

\begin{table}[t]
\centering
\small
\caption{The $\operatorname{rSign}^{AR}_m(H)$ results across different prediction steps $H$ and correlation levels $\rho$ under the omitted relevant predictor setting.}
\label{tab:simu_miss_sign}
\begin{tabular}{cccccccccc}
\toprule
Method & 8 & 9 & 10 & 11 & 12 & 13 & 14 & 15 & Ave. \\
\midrule
\multicolumn{10}{c}{$\rho = 0.1$} \\
\midrule
RW       & 0.637 & 0.932 & 0.613 & 0.591 & 0.675 & 0.545 & 0.688 & 0.695 & 0.672 \\
AVE      & 0.631 & 0.932 & 0.517 & 0.588 & 0.675 & 0.545 & 0.688 & 0.695 & 0.659 \\
LLM-CPI  & 0.398 & 0.768 & 0.392 & 0.531 & 0.697 & 0.542 & 0.682 & 0.694 & 0.588 \\
\midrule
\multicolumn{10}{c}{$\rho = 0.2$} \\
\midrule
RW       & 0.638 & 0.937 & 0.597 & 0.583 & 0.681 & 0.543 & 0.700 & 0.707 & 0.673 \\
AVE      & 0.630 & 0.937 & 0.513 & 0.583 & 0.681 & 0.543 & 0.700 & 0.707 & 0.662 \\
LLM-CPI  & 0.399 & 0.764 & 0.381 & 0.516 & 0.696 & 0.534 & 0.683 & 0.700 & 0.584 \\
\midrule
\multicolumn{10}{c}{$\rho = 0.3$} \\
\midrule
RW       & 0.630 & 0.930 & 0.615 & 0.581 & 0.671 & 0.541 & 0.698 & 0.711 & 0.672 \\
AVE      & 0.625 & 0.930 & 0.511 & 0.581 & 0.671 & 0.541 & 0.698 & 0.711 & 0.658 \\
LLM-CPI  & 0.385 & 0.738 & 0.373 & 0.509 & 0.667 & 0.519 & 0.666 & 0.683 & 0.567 \\
\midrule
\multicolumn{10}{c}{$\rho = 0.4$} \\
\midrule
RW       & 0.609 & 0.930 & 0.604 & 0.587 & 0.663 & 0.532 & 0.682 & 0.688 & 0.662 \\
AVE      & 0.605 & 0.930 & 0.514 & 0.587 & 0.663 & 0.532 & 0.682 & 0.688 & 0.638 \\
LLM-CPI  & 0.353 & 0.744 & 0.378 & 0.514 & 0.643 & 0.511 & 0.654 & 0.665 & 0.545 \\
\bottomrule
\end{tabular}\\
{\footnotesize Note: The relative sign prediction error values compared to the AR benchmark. Smaller values indicate better performance.}
\end{table}

\begin{table}[t]
\centering
\small
\caption{The $\operatorname{Coverage}_m(H)$ and $\operatorname{Length}_m(H)$ results across different prediction steps $H$ and correlation levels $\rho$ under the omitted relevant predictor setting.}
\label{tab:simu_miss}
\small{
\begin{tabular}{cccccccccc}
\toprule
Method & 8 & 9 & 10 & 11 & 12 & 13 & 14 & 15 & Ave. \\
\midrule
\multicolumn{10}{c}{$\rho = 0.1$} \\
\midrule
AR          & 1.000 & 1.000 & 1.000 & 1.000 & 1.000 & 1.000 & 1.000 & 1.000 & 1.000 \\
            & (3.623) & (3.719) & (3.762) & (3.830) & (3.898) & (3.951) & (4.049) & (4.126) & (3.857) \\
BJ      & 0.958 & 0.948 & 0.954 & 0.953 & 0.958 & 0.963 & 0.964 & 0.945 & 0.956 \\
            & (1.225) & (1.199) & (1.202) & (1.209) & (1.239) & (1.251) & (1.250) & (1.290) & (1.233) \\
BOOT    & 0.937 & 0.910 & 0.929 & 0.931 & 0.930 & 0.934 & 0.924 & 0.860 & 0.932 \\
            & (1.385) & (1.358) & (1.378) & (1.383) & (1.432) & (1.448) & (1.452) & (1.476) & (1.414) \\
\midrule
\multicolumn{10}{c}{$\rho = 0.2$} \\
\midrule
AR          & 1.000 & 1.000 & 1.000 & 1.000 & 1.000 & 1.000 & 1.000 & 1.000 & 1.000 \\
            & (3.623) & (3.720) & (3.763) & (3.831) & (3.899) & (3.952) & (4.049) & (4.125) & (3.870) \\
BJ      & 0.962 & 0.947 & 0.954 & 0.954 & 0.957 & 0.963 & 0.963 & 0.943 & 0.956 \\
            & (1.209) & (1.185) & (1.185) & (1.191) & (1.229) & (1.233) & (1.240) & (1.265) & (1.217) \\
BOOT    & 0.937 & 0.910 & 0.929 & 0.936 & 0.929 & 0.929 & 0.930 & 0.865 & 0.933 \\
            & (1.373) & (1.348) & (1.366) & (1.377) & (1.432) & (1.426) & (1.460) & (1.459) & (1.405) \\
\midrule
\multicolumn{10}{c}{$\rho = 0.3$} \\
\midrule
AR          & 1.000 & 1.000 & 1.000 & 1.000 & 1.000 & 1.000 & 1.000 & 1.000 & 1.000 \\
            & (3.625) & (3.723) & (3.764) & (3.832) & (3.900) & (3.954) & (4.051) & (4.129) & (3.872) \\
BJ      & 0.962 & 0.946 & 0.952 & 0.953 & 0.956 & 0.961 & 0.963 & 0.946 & 0.956 \\
            & (1.198) & (1.171) & (1.168) & (1.172) & (1.216) & (1.223) & (1.228) & (1.274) & (1.206) \\
BOOT    & 0.939 & 0.908 & 0.926 & 0.932 & 0.930 & 0.936 & 0.928 & 0.861 & 0.933 \\
            & (1.360) & (1.334) & (1.344) & (1.356) & (1.412) & (1.426) & (1.444) & (1.469) & (1.393) \\
\midrule
\multicolumn{10}{c}{$\rho = 0.4$} \\
\midrule
AR          & 1.000 & 1.000 & 1.000 & 1.000 & 1.000 & 1.000 & 1.000 & 1.000 & 1.000 \\
            & (3.628) & (3.725) & (3.767) & (3.835) & (3.903) & (3.956) & (4.053) & (4.130) & (3.875) \\
BJ      & 0.953 & 0.945 & 0.946 & 0.947 & 0.955 & 0.961 & 0.961 & 0.941 & 0.951 \\
            & (1.159) & (1.150) & (1.149) & (1.159) & (1.203) & (1.213) & (1.218) & (1.255) & (1.188) \\
BOOT    & 0.943 & 0.918 & 0.936 & 0.936 & 0.934 & 0.943 & 0.929 & 0.863 & 0.925 \\
            & (1.338) & (1.325) & (1.338) & (1.356) & (1.409) & (1.428) & (1.453) & (1.451) & (1.387) \\
\bottomrule
\end{tabular}}\\
{\footnotesize Note: The values in the parentheses are interval length, and coverage near the nominal level of $0.95$ with smaller interval length is preferred.}
\end{table}

\section{Additional simulation results on the robustness of LLM-CPI} \label{new.Sec.robustness}

In this section, we present additional simulation results to evaluate the robustness of the suggested LLM-CPI method under various model misspecification or overfitting scenarios, where both the Box--Jenkins (BJ) prediction interval and the bootstrap (BOOT) prediction interval introduced in Section \ref{Sec.CPIpredinf} are examined.

\subsection{Omitted relevant predictor} \label{new.Sec.robustness.1}

We now conduct a simulation experiment in which a key predictor is intentionally omitted from both the target CPI and the LLM surrogate models. Specifically, we exclude the second predictor from the estimation procedure while keeping the data-generating process unchanged. Tables \ref{tab:simu_miss_mse} and \ref{tab:simu_miss_sign} summarize the relative root prediction mean squared error (PMSE) and the relative sign prediction error (rSign), respectively. The simulation results in Tables \ref{tab:simu_miss_mse} and \ref{tab:simu_miss_sign} reveal that the LLM-CPI is robust under the omitted relevant predictor setting. First, although the omission of a relevant predictor leads to a deterioration in the LLM-CPI model forecasting accuracy compared to the correctly specified case (cf. Table \ref{tab:simu_true_mse}), the model still significantly outperforms traditional benchmarks across different forecast horizons $H$ and correlation levels $\rho$. Second, the LLM-CPI model also exhibits stable performance in directional forecasting (i.e., in terms of the relative sign prediction error denoted as $\operatorname{rSign}$). Despite the model misspecification, its corresponding $\operatorname{rSign}$ metric remains the lowest among all methods.

In addition to the point forecasts, we assess the quality of prediction intervals for the LLM-CPI under the omitted relevant predictor setting. Table \ref{tab:simu_miss} presents the coverage rates and average interval lengths, respectively. Both the BJ and bootstrap intervals maintain coverage rates that are close to the nominal level $95\%$ across all forecast horizons and correlation levels. Notably, the BJ interval exhibits more stable coverage properties compared to the bootstrap interval, particularly for longer horizons. Despite the model misspecification, both LLM-CPI prediction intervals remain substantially narrower than those of the AR benchmark, with the average interval length being roughly one third of that of the AR interval.

In summary, the omitted relevant predictor experiment demonstrates that the LLM-CPI model retains robust forecasting capabilities and prediction interval efficiency even when some key predictors are missing.

\subsection{Model overfitting} \label{new.Sec.robustness.2}

To further assess the robustness of the LLM-CPI, we consider a simulation scenario characterized by overfitting, where the model is specified with irrelevant predictors. Specifically, we augment the surrogate model with additional predictors that are uncorrelated with the response, thereby simulating a common empirical situation in which noisy or spurious features are mistakenly included in the model specification. The true data-generating process remains unchanged from the baseline simulation setup.

Tables \ref{tab:simu_over_mse} and \ref{tab:simu_over_sign} report the results on the relative root prediction mean squared error (rPMSE) and the relative sign prediction error (rSign), respectively. Despite the inclusion of irrelevant features, the LLM-CPI model continues to outperform traditional benchmarks across all prediction horizons $H$ and correlation levels $\rho$. The average rPMSE remains low, indicating that the LLM-CPI framework effectively mitigates the estimation noise introduced by the irrelevant predictors. The LLM-CPI model also maintains low rSign values, reflecting strong directional accuracy even in the presence of overfitting. Although the relative sign prediction error values (i.e., rSign) are slightly higher than those in the correctly specified model, they remain substantially lower than those of the baseline AR model and other benchmarks.

In terms of forecast uncertainty quantification, Table \ref{tab:simu_over} lists the coverage rates and average interval lengths of the prediction intervals under the overfitting setting. 
Both the BJ and bootstrap prediction intervals exhibit slight under-coverage, particularly for the longer-term forecasts and lower correlation settings. This suggests that overfitting may lead to underestimated forecast uncertainty. Regardless, both LLM-CPI prediction intervals remain substantially narrower than those of the AR model, indicating a considerable gain in interval tightness.

Overall, the overfitting experiment highlights the robustness and adaptivity of the LLM-CPI framework in the presence of model overparameterization. While the forecast interval reliability is slightly affected, the point forecasts remain highly accurate, and the suggested method continues outperforming conventional alternatives. Such robustness makes the LLM-CPI method well-suited for practical forecasting settings where the risk of overfitting is non-negligible.

\begin{table}[t]
\centering
\small
\caption{The $\operatorname{rPMSE}^{AR}_m(H)$ results across different prediction steps $H$ and correlation levels $\rho$ under the model overfitting setting.}
\label{tab:simu_over_mse}
\begin{tabular}{cccccccccc}
\toprule
Method & 8 & 9 & 10 & 11 & 12 & 13 & 14 & 15 & Ave. \\
\midrule
\multicolumn{10}{c}{$\rho = 0.1$} \\
\midrule
RW       & 0.768 & 1.174 & 0.802 & 0.857 & 2.073 & 3.556 & 14.661 & 19.358 & 5.781 \\
AVE      & 0.686 & 1.102 & 2.327 & 4.933 & 7.849 & 9.188 & 10.753 & 8.133 & 5.622 \\
LLM-CPI  & 0.212 & 0.232 & 0.274 & 0.322 & 0.346 & 0.333 & 0.368 & 0.436 & 0.316 \\
\midrule
\multicolumn{10}{c}{$\rho = 0.2$} \\
\midrule
RW       & 0.761 & 1.164 & 0.798 & 0.871 & 2.152 & 3.638 & 15.100 & 19.845 & 5.791 \\
AVE      & 0.677 & 1.130 & 2.397 & 5.084 & 8.082 & 9.416 & 10.999 & 8.199 & 5.748 \\
LLM-CPI  & 0.215 & 0.231 & 0.271 & 0.320 & 0.346 & 0.332 & 0.365 & 0.433 & 0.314 \\
\midrule
\multicolumn{10}{c}{$\rho = 0.3$} \\
\midrule
RW       & 0.761 & 1.171 & 0.795 & 0.833 & 2.096 & 3.721 & 15.065 & 19.784 & 5.691 \\
AVE      & 0.675 & 1.115 & 2.366 & 5.010 & 8.044 & 9.488 & 10.987 & 8.218 & 5.738 \\
LLM-CPI  & 0.205 & 0.225 & 0.266 & 0.315 & 0.337 & 0.327 & 0.357 & 0.421 & 0.319 \\
\midrule
\multicolumn{10}{c}{$\rho = 0.4$} \\
\midrule
RW       & 0.759 & 1.173 & 0.796 & 0.855 & 2.082 & 3.606 & 15.014 & 19.751 & 5.742 \\
AVE      & 0.676 & 1.120 & 2.380 & 5.058 & 8.023 & 9.376 & 10.989 & 8.217 & 5.730 \\
LLM-CPI  & 0.195 & 0.215 & 0.260 & 0.305 & 0.323 & 0.314 & 0.348 & 0.409 & 0.296 \\
\bottomrule
\end{tabular}\\
{\footnotesize Note: The relative PMSE values compared to the AR benchmark. Smaller values indicate better performance.}
\end{table}

\begin{table}[t]
\centering
\small
\caption{The $\operatorname{rSign}^{AR}_m(H)$ results across different prediction steps $H$ and correlation levels $\rho$ under the model overfitting setting.}
\label{tab:simu_over_sign}
\begin{tabular}{cccccccccc}
\toprule
Method & 8 & 9 & 10 & 11 & 12 & 13 & 14 & 15 & Ave. \\
\midrule
\multicolumn{10}{c}{$\rho = 0.1$} \\
\midrule
RW        & 0.637 & 0.932 & 0.613 & 0.591 & 0.675 & 0.545 & 0.688 & 0.695 & 0.672 \\
AVE       & 0.631 & 0.932 & 0.517 & 0.588 & 0.675 & 0.545 & 0.688 & 0.695 & 0.659 \\
LLM-CPI   & 0.357 & 0.675 & 0.326 & 0.440 & 0.508 & 0.409 & 0.522 & 0.491 & 0.466 \\
\midrule
\multicolumn{10}{c}{$\rho = 0.2$} \\
\midrule
RW        & 0.638 & 0.937 & 0.597 & 0.583 & 0.681 & 0.543 & 0.700 & 0.707 & 0.673 \\
AVE       & 0.630 & 0.937 & 0.513 & 0.583 & 0.681 & 0.543 & 0.700 & 0.707 & 0.662 \\
LLM-CPI   & 0.360 & 0.697 & 0.323 & 0.430 & 0.507 & 0.405 & 0.531 & 0.505 & 0.470 \\
\midrule
\multicolumn{10}{c}{$\rho = 0.3$} \\
\midrule
RW        & 0.630 & 0.930 & 0.615 & 0.581 & 0.671 & 0.541 & 0.698 & 0.711 & 0.672 \\
AVE       & 0.625 & 0.930 & 0.511 & 0.581 & 0.671 & 0.541 & 0.698 & 0.711 & 0.658 \\
LLM-CPI   & 0.335 & 0.670 & 0.317 & 0.431 & 0.499 & 0.399 & 0.518 & 0.499 & 0.458 \\
\midrule
\multicolumn{10}{c}{$\rho = 0.4$} \\
\midrule
RW        & 0.609 & 0.930 & 0.604 & 0.587 & 0.663 & 0.532 & 0.682 & 0.688 & 0.662 \\
AVE       & 0.605 & 0.930 & 0.514 & 0.587 & 0.663 & 0.532 & 0.682 & 0.688 & 0.638 \\
LLM-CPI   & 0.306 & 0.661 & 0.313 & 0.421 & 0.475 & 0.382 & 0.500 & 0.473 & 0.441 \\
\bottomrule
\end{tabular}\\
{\footnotesize Note: The relative sign prediction error values  compared to the AR benchmark. Smaller values indicate better performance.}
\end{table}

\begin{table}[t]
\centering
\small
\caption{The $\operatorname{Coverage}_m(H)$ and $\operatorname{Length}_m(H)$ results across different prediction steps $H$ and correlation levels $\rho$ under the model overfitting setting.}
\label{tab:simu_over}
\small{
\begin{tabular}{cccccccccc}
\toprule
Method & 8 & 9 & 10 & 11 & 12 & 13 & 14 & 15 & Ave. \\
\midrule
\multicolumn{10}{c}{$\rho = 0.1$} \\
\midrule
AR          & 1.000 & 1.000 & 1.000 & 1.000 & 1.000 & 1.000 & 1.000 & 1.000 & 1.000 \\
            & (3.623) & (3.719) & (3.762) & (3.830) & (3.898) & (3.951) & (4.049) & (4.126) & (3.857) \\
BJ      & 0.933 & 0.908 & 0.911 & 0.895 & 0.893 & 0.902 & 0.907 & 0.882 & 0.917 \\
            & (0.777) & (0.774) & (0.772) & (0.772) & (0.775) & (0.778) & (0.778) & (0.780) & (0.776) \\
BOOT    & 0.940 & 0.914 & 0.907 & 0.914 & 0.910 & 0.914 & 0.871 & 0.849 & 0.902 \\
            & (1.054) & (1.057) & (1.059) & (1.099) & (1.119) & (1.123) & (1.121) & (1.121) & (1.094) \\
\midrule
\multicolumn{10}{c}{$\rho = 0.2$} \\
\midrule
AR          & 1.000 & 1.000 & 1.000 & 1.000 & 1.000 & 1.000 & 1.000 & 1.000 & 1.000 \\
            & (3.623) & (3.720) & (3.763) & (3.831) & (3.899) & (3.952) & (4.049) & (4.125) & (3.870) \\
BJ      & 0.936 & 0.915 & 0.914 & 0.901 & 0.902 & 0.908 & 0.913 & 0.886 & 0.909 \\
            & (0.763) & (0.761) & (0.758) & (0.759) & (0.763) & (0.766) & (0.768) & (0.770) & (0.764) \\
BOOT    & 0.933 & 0.916 & 0.905 & 0.908 & 0.904 & 0.913 & 0.871 & 0.853 & 0.900 \\
            & (1.045) & (1.052) & (1.051) & (1.093) & (1.112) & (1.117) & (1.114) & (1.117) & (1.088) \\
\midrule
\multicolumn{10}{c}{$\rho = 0.3$} \\
\midrule
AR          & 1.000 & 1.000 & 1.000 & 1.000 & 1.000 & 1.000 & 1.000 & 1.000 & 1.000 \\
            & (3.625) & (3.723) & (3.764) & (3.832) & (3.900) & (3.954) & (4.051) & (4.129) & (3.872) \\
BJ      & 0.933 & 0.911 & 0.918 & 0.898 & 0.897 & 0.909 & 0.913 & 0.883 & 0.920 \\
            & (0.759) & (0.756) & (0.753) & (0.753) & (0.755) & (0.758) & (0.758) & (0.761) & (0.757) \\
BOOT    & 0.936 & 0.915 & 0.902 & 0.910 & 0.900 & 0.912 & 0.868 & 0.854 & 0.912 \\
            & (1.039) & (1.044) & (1.044) & (1.086) & (1.105) & (1.108) & (1.107) & (1.110) & (1.080) \\
\midrule
\multicolumn{10}{c}{$\rho = 0.4$} \\
\midrule
AR          & 1.000 & 1.000 & 1.000 & 1.000 & 1.000 & 1.000 & 1.000 & 1.000 & 1.000 \\
            & (3.628) & (3.725) & (3.767) & (3.835) & (3.903) & (3.956) & (4.053) & (4.130) & (3.875) \\
BJ      & 0.939 & 0.916 & 0.916 & 0.897 & 0.902 & 0.913 & 0.919 & 0.893 & 0.912 \\
            & (0.748) & (0.747) & (0.744) & (0.744) & (0.747) & (0.752) & (0.753) & (0.756) & (0.749) \\
BOOT    & 0.938 & 0.917 & 0.906 & 0.911 & 0.902 & 0.916 & 0.873 & 0.850 & 0.902 \\
            & (1.031) & (1.036) & (1.036) & (1.075) & (1.097) & (1.100) & (1.099) & (1.100) & (1.072) \\
\bottomrule
\end{tabular}}\\
{\footnotesize Note: The values in the parentheses are interval length, and coverage near the nominal level of $0.95$ with smaller interval length is preferred.}
\end{table}

\begin{table}[t]
\centering
\small
\caption{The $\operatorname{rPMSE}^{AR}_m(H)$ results across different prediction steps $H$ and correlation levels $\rho$ under the $t$-distribution setting.}
\label{tab:simu_t_mse}
\begin{tabular}{cccccccccc}
\toprule
Method & 8 & 9 & 10 & 11 & 12 & 13 & 14 & 15 & Ave. \\
\midrule
\multicolumn{10}{c}{$\rho = 0.1$} \\
\midrule
RW        & 0.781 & 1.190 & 0.810 & 0.867 & 2.038 & 3.550 & 14.437 & 18.960 & 5.704 \\
AVE       & 0.689 & 1.111 & 2.337 & 4.896 & 7.739 & 9.098 & 10.550 & 7.917 & 5.792 \\
LLM-CPI   & 0.222 & 0.224 & 0.266 & 0.330 & 0.348 & 0.317 & 0.341 & 0.428 & 0.309 \\
\midrule
\multicolumn{10}{c}{$\rho = 0.2$} \\
\midrule
RW        & 0.781 & 1.177 & 0.811 & 0.864 & 2.068 & 3.597 & 14.524 & 19.198 & 5.740 \\
AVE       & 0.686 & 1.120 & 2.335 & 4.905 & 7.755 & 9.147 & 10.584 & 7.994 & 5.691 \\
LLM-CPI   & 0.217 & 0.222 & 0.262 & 0.327 & 0.344 & 0.317 & 0.338 & 0.429 & 0.307 \\
\midrule
\multicolumn{10}{c}{$\rho = 0.3$} \\
\midrule
RW        & 0.776 & 1.186 & 0.814 & 0.873 & 2.082 & 3.637 & 14.760 & 19.425 & 5.757 \\
AVE       & 0.694 & 1.136 & 2.366 & 4.984 & 7.858 & 9.241 & 10.719 & 8.083 & 5.760 \\
LLM-CPI   & 0.211 & 0.217 & 0.258 & 0.324 & 0.339 & 0.307 & 0.330 & 0.414 & 0.300 \\
\midrule
\multicolumn{10}{c}{$\rho = 0.4$} \\
\midrule
RW        & 0.788 & 1.177 & 0.822 & 0.893 & 2.103 & 3.605 & 14.640 & 19.410 & 5.805 \\
AVE       & 0.691 & 1.126 & 2.373 & 4.988 & 7.888 & 9.258 & 10.771 & 8.044 & 5.767 \\
LLM-CPI   & 0.202 & 0.203 & 0.245 & 0.312 & 0.327 & 0.295 & 0.319 & 0.397 & 0.288 \\
\bottomrule
\end{tabular}\\
{\footnotesize Note: The relative PMSE values compared to the AR benchmark. Smaller values indicate better performance.}
\end{table}

\begin{table}[t]
\centering
\small
\caption{The $\operatorname{rSign}^{AR}_m(H)$ results across different prediction steps $H$ and correlation levels $\rho$ under the $t$-distribution setting.}
\label{tab:simu_t_sign}
\begin{tabular}{cccccccccc}
\toprule
Method & 8 & 9 & 10 & 11 & 12 & 13 & 14 & 15 & Ave. \\
\midrule
\multicolumn{10}{c}{$\rho = 0.1$} \\
\midrule
RW        & 0.647 & 0.919 & 0.634 & 0.603 & 0.673 & 0.559 & 0.694 & 0.701 & 0.679 \\
AVE       & 0.635 & 0.919 & 0.538 & 0.600 & 0.673 & 0.559 & 0.694 & 0.701 & 0.665 \\
LLM-CPI   & 0.361 & 0.600 & 0.286 & 0.412 & 0.462 & 0.368 & 0.455 & 0.438 & 0.423 \\
\midrule
\multicolumn{10}{c}{$\rho = 0.2$} \\
\midrule
RW        & 0.636 & 0.914 & 0.614 & 0.583 & 0.665 & 0.543 & 0.686 & 0.693 & 0.667 \\
AVE       & 0.620 & 0.914 & 0.519 & 0.580 & 0.665 & 0.543 & 0.686 & 0.693 & 0.652 \\
LLM-CPI   & 0.361 & 0.613 & 0.283 & 0.405 & 0.460 & 0.365 & 0.463 & 0.443 & 0.424 \\
\midrule
\multicolumn{10}{c}{$\rho = 0.3$} \\
\midrule
RW        & 0.636 & 0.919 & 0.615 & 0.586 & 0.659 & 0.544 & 0.680 & 0.683 & 0.665 \\
AVE       & 0.623 & 0.920 & 0.514 & 0.583 & 0.659 & 0.544 & 0.680 & 0.683 & 0.651 \\
LLM-CPI   & 0.351 & 0.605 & 0.281 & 0.404 & 0.458 & 0.363 & 0.456 & 0.432 & 0.419 \\
\midrule
\multicolumn{10}{c}{$\rho = 0.4$} \\
\midrule
RW        & 0.638 & 0.913 & 0.631 & 0.593 & 0.682 & 0.544 & 0.700 & 0.702 & 0.675 \\
AVE       & 0.621 & 0.912 & 0.523 & 0.590 & 0.682 & 0.544 & 0.700 & 0.702 & 0.659 \\
LLM-CPI   & 0.326 & 0.557 & 0.261 & 0.385 & 0.442 & 0.339 & 0.440 & 0.417 & 0.396 \\
\bottomrule
\end{tabular}\\
{\footnotesize Note: The relative sign prediction error values  compared to the AR benchmark. Smaller values indicate better performance.}
\end{table}

\begin{table}[t]
\centering
\small
\caption{The $\operatorname{Coverage}_m(H)$ and $\operatorname{Length}_m(H)$ results across different prediction steps $H$ and correlation levels $\rho$ under the $t$-distribution setting.}
\label{tab:simu_t}
\begin{tabular}{cccccccccc}
\toprule
Method & 8 & 9 & 10 & 11 & 12 & 13 & 14 & 15 & Ave. \\
\midrule
\multicolumn{10}{c}{$\rho = 0.1$} \\
\midrule
AR          & 1.000 & 1.000 & 1.000 & 1.000 & 1.000 & 1.000 & 1.000 & 1.000 & 1.000 \\
            & (3.630) & (3.726) & (3.768) & (3.836) & (3.903) & (3.956) & (4.052) & (4.128) & (3.875) \\
BJ      & 0.947 & 0.930 & 0.930 & 0.914 & 0.912 & 0.927 & 0.932 & 0.905 & 0.925 \\
            & (0.823) & (0.825) & (0.817) & (0.811) & (0.813) & (0.818) & (0.821) & (0.819) & (0.818) \\
BOOT    & 0.932 & 0.921 & 0.932 & 0.925 & 0.925 & 0.936 & 0.911 & 0.882 & 0.921 \\
            & (1.064) & (1.068) & (1.077) & (1.087) & (1.105) & (1.110) & (1.113) & (1.111) & (1.092) \\
\midrule
\multicolumn{10}{c}{$\rho = 0.2$} \\
\midrule
AR          & 1.000 & 1.000 & 1.000 & 1.000 & 1.000 & 1.000 & 1.000 & 1.000 & 1.000 \\
            & (3.635) & (3.731) & (3.773) & (3.841) & (3.908) & (3.962) & (4.057) & (4.136) & (3.880) \\
BJ      & 0.941 & 0.927 & 0.926 & 0.906 & 0.907 & 0.922 & 0.931 & 0.905 & 0.921 \\
            & (0.817) & (0.819) & (0.811) & (0.805) & (0.807) & (0.812) & (0.814) & (0.815) & (0.812) \\
BOOT    & 0.933 & 0.930 & 0.932 & 0.923 & 0.925 & 0.936 & 0.910 & 0.881 & 0.921 \\
            & (1.053) & (1.059) & (1.061) & (1.075) & (1.095) & (1.099) & (1.099) & (1.099) & (1.080) \\
\midrule
\multicolumn{10}{c}{$\rho = 0.3$} \\
\midrule
AR          & 1.000 & 1.000 & 1.000 & 1.000 & 1.000 & 1.000 & 1.000 & 1.000 & 1.000 \\
            & (3.632) & (3.727) & (3.770) & (3.838) & (3.905) & (3.959) & (4.055) & (4.132) & (3.877) \\
BJ      & 0.948 & 0.933 & 0.934 & 0.912 & 0.914 & 0.927 & 0.937 & 0.905 & 0.926 \\
            & (0.811) & (0.814) & (0.805) & (0.799) & (0.802) & (0.806) & (0.809) & (0.809) & (0.807) \\
BOOT    & 0.924 & 0.917 & 0.930 & 0.921 & 0.925 & 0.936 & 0.917 & 0.881 & 0.919 \\
            & (1.056) & (1.060) & (1.064) & (1.076) & (1.095) & (1.100) & (1.101) & (1.099) & (1.081) \\
\midrule
\multicolumn{10}{c}{$\rho = 0.4$} \\
\midrule
AR          & 1.000 & 1.000 & 1.000 & 1.000 & 1.000 & 1.000 & 1.000 & 1.000 & 1.000 \\
            & (3.633) & (3.729) & (3.772) & (3.840) & (3.907) & (3.961) & (4.057) & (4.133) & (3.879) \\
BJ      & 0.942 & 0.929 & 0.931 & 0.908 & 0.907 & 0.920 & 0.927 & 0.898 & 0.920 \\
            & (0.791) & (0.794) & (0.786) & (0.779) & (0.781) & (0.786) & (0.788) & (0.789) & (0.787) \\
BOOT    & 0.930 & 0.926 & 0.931 & 0.922 & 0.921 & 0.932 & 0.907 & 0.878 & 0.919 \\
            & (1.035) & (1.042) & (1.045) & (1.059) & (1.076) & (1.081) & (1.079) & (1.085) & (1.063) \\
\bottomrule
\end{tabular}\\
{\footnotesize Note: The values in the parentheses are interval length, and coverage near the nominal level of $0.95$ with smaller interval length is preferred.}
\end{table}

\subsection{\textit{t}-distributions} \label{new.Sec.robustness.3}

To investigate the robustness of the LLM-CPI framework under non-Gaussian error distributions, we consider a simulated example in which the error terms of both the target CPI model and the LLM surrogate model follow a $t$-distribution with $10$ degrees of freedom. This setting introduces heavy-tailed innovations, which are common in macroeconomic and financial time series, and can pose challenges for parametric models relying on the normality assumptions. Importantly, the fitting procedure remains unchanged; that is, we still estimate the LLM-powered joint time series model using LLM-CPI assuming Gaussian errors.

Tables \ref{tab:simu_t_mse} and \ref{tab:simu_t_sign} present the rPMSE and rSign, respectively, results across different forecast horizons $H$ and correlation levels $\rho$. The LLM-CPI model continues to substantially outperform the benchmark models in both point and directional forecast accuracies, despite the presence of heavy-tailed innovations. Such robustness is particularly evident at higher correlation levels (e.g., $\rho = 0.3$ and $0.4$), where the average rPMSE drops below $0.3$, significantly better than all the competitors. Further, even under the $t$-distribution, the LLM-CPI model achieves substantially lower rSign values, indicating its ability to correctly predict the direction of CPI movements. While some slight increases in rSign are observed compared to the Gaussian baseline, the performance gap relative to the AR, RW, and AVE remains large.

The forecast interval performance under the $t$-distribution is documented in Table \ref{tab:simu_t}. Both the BJ and Bootstrap prediction intervals maintain coverage rates close to the nominal level ($95\%$), with only slight under-coverage for longer horizons. The LLM-CPI method continues to produce substantially narrower prediction intervals compared to the AR-based intervals. For instance, the average length of the BJ intervals remains under $0.82$ across all $\rho$ values, compared to the AR intervals exceeding $3.87$ in average length, representing a nearly $80\%$ reduction in interval length.

In summary, the $t$-distribution robustness check confirms that the LLM-CPI framework is robust to heavy-tailed innovations, delivering strong point forecasts, accurate directional predictions, and efficient prediction intervals even when the standard distributional assumptions are violated. These simulation results underscore the LLM-CPI's practical utility in real-world forecasting environments where the error distributions may deviate from normality.

\section{Additional real data results} \label{new.Sec.add.realdata}


\subsection{Details of forecasting models} \label{new.Sec.forecastmods}

We describe in this section four widely used inflation forecasting models. For a comprehensive comparison of different inflation prediction models, see, e.g., \cite{stock2008phillips}. Let $\E(y_{T+h})$ be the conditional expectation of the $h$-step-ahead inflation given the historical data $\{y_{1},\ldots, y_{T}\}$. The simplest approach to inflation forecasting relies on the random walk (RW) \citep{atkeson2001phillips} given by 
\begin{equation}\label{equ:random_walk}
   \hat{y}_{T+h, RW} = y_{T}. \tag{RW}
\end{equation}
This model is (surprisingly) a most powerful one in inflation prediction especially during the post-1984 ``Great Moderation'' period of the U.S. The second forecast model is the historical average (AVE) forecast defined as 
\begin{equation}\label{equ:mean_model}
    \hat{y}_{T+h, AVE}= \frac{1}{h} \sum_{l=0}^{h-1} y_{T-l}, \tag{AVE} 
\end{equation}
which is popular in financial market forecasting \citep{welch2008comprehensive}. The third prototype model is the autoregressive (AR) model specified as 
\begin{equation}\label{equ:pure_AR} 
   \hat{y}_{T+h, AR} = \sum_{l=1}^{q_1} \hat{\alpha}_{l,AR} \hat{y}_{T+h-l,AR}, \tag{AR}
\end{equation}
Here, $\hat{y}_{t,AR}$ represents the observed inflation for $t \le T$ and the prediction generated by this AR model for $t > T$, and $\hat{\alpha}_{l,AR}$'s are the estimated coefficients for the AR model. The lag length $q_1 \geq 1$ is typically determined using the corrected Akaike information criterion (AIC) \citep{hurvich1989regression}. {Using the notation introduced in Section \ref{sec.proof.lem:prediction}, the BJ prediction interval for the AR model is defined as 
\begin{equation}\label{equ:ar_ci}
\begin{aligned}
 &\operatorname{PI}^{AR}(\hat{y}_{T+h,AR})= \\
 &\left[\hat{y}_{T+h,AR} - |z_{\alpha/2}| \sqrt{\sum_{r=0}^{h-1}((\hbA^{AR})^r)^2_{11}}\hat{\sigma}^{AR}_e, \hat{y}_{T+h,AR} + |z_{\alpha/2}|\sqrt{ \sum_{r=0}^{h-1}(\hbA^{AR})^2_{11}} \hat{\sigma}^{AR}_e\right],   
\end{aligned}
\end{equation}
where $\hat{y}_{T+h,AR}$ is the $h$-step-ahead prediction from the AR model, $\hbA^{AR}$ is analogous to $\hbA$ defined in \eqref{equ:march05:equ2}, with its elements replaced by the estimated AR coefficients, and $\hat{\sigma}^{AR}_e$ is the standard error estimated from the AR residuals. 
}

The previous work also adapts the Phillips curve, particularly Gordon's ``triangle model'' \cite{gordon1988us}, and incorporates both lagged inflation and unemployment rates to predict inflation. The resulting autoregressive exogenous (ARX) model is defined as 
\begin{equation}\label{equ:ar_and_unem}
\hat{y}_{T+h, ARX} = \sum_{l=1}^{q_1} \hat{\alpha}_{l,ARX} \hat{y}_{T+h-l,ARX}  +  \hat{\beta}_{ARX} z_{T+h},    \tag{ARX}
\end{equation}
where inflation prediction depends on the lagged inflation, the unemployment rate $z_t$, and $\hat{\alpha}_{l,ARX}, \hat{\beta}_{ARX}$ are the estimated coefficients for the ARX model. There exist many other methods for predicting inflation, but the ones mentioned above are the popular prototype models in the literature.

Let us denote the selected embedding features (either LDA or BERT embeddings) for the $t$th month as $\bx_t$. Exploiting these embeddings, we can introduce two ARX models with text features. The first one is the AR model with text features, but excludes the unemployment rate 
\begin{equation}\label{equ:target1}
\hat{y}_{T+h,m} =  \sum_{l=1}^{q_1} \hat{\alpha}_{l,m} \hat{y}_{T+h-l,m} + \hat{\bbeta}_m^\top \bx_{T+h}.
\end{equation}
The second one is Gordon's ``triangle model'' with text features, i.e.,
\begin{equation}\label{equ:target2}
\hat{y}_{T+t,m} =  \sum_{l=1}^{q_1} \hat{\alpha}_{l} \hat{y}_{T+h-l,m} + \hat{\theta}_m z_{T+h} +  \hat{\bbeta}_m^\top\bx_{T+h}.
\end{equation}
Here, $\bx_t$ represents either the LDA embeddings $\bx_t^{\text{LDA}}$ (with $m = \text{LDA}$) or the BERT embeddings $\bx_t^{\text{BERT}}$ (with $m = \text{BERT}$). For ease of reference, we refer to the text-based prediction model with the LDA embeddings as the LDA model, and the text-based prediction model with the BERT embeddings as the BERT model (with slight abuse of terminology). In particular, the LDA embedding-based CPI prediction method has been successfully applied to the U.S. inflation forecasting \citep{hong2025forecasting}.

Our suggested LLM-powered CPI prediction inference (LLM-CPI) framework builds upon models \eqref{equ:target1} and \eqref{equ:target2}. Using the LLM-generated inflation index $\bx_t$, we can construct an LLM-based VARX$(q_2)$ surrogate model with $q_2 \geq 1$. Due to limited observations for the CPI index, we conservatively set $q_2=1$ to ensure the model identifiability 
\begin{equation}\label{equ:real_sur}
    \by^S_t = \bA^S  \by^S_{t-1} + \bB^S \bx_t + \beps^S_t. \tag{VARX}
\end{equation}
We consider two variants of the LLM-CPI model by combining different specifications:
\begin{itemize}
    \item[1)] LLM-CPI: integrating model \eqref{equ:target1} with the surrogate model \eqref{equ:real_sur}, excluding unemployment variables.

    \item[2)] LLM-CPI with unemployment rate: combining model \eqref{equ:target2} with the surrogate model \eqref{equ:real_sur}, incorporating both unemployment and text features.
\end{itemize}
Similarly, we abbreviate the LLM-CPI method with the LDA and BERT embeddings as LLM+LDA and LLM+BERT, respectively. 

\subsection{Out-of-sample forecasting with unemployment rate} \label{new.Sec.add.real.forecasting}

We expand our forecasting evaluation by incorporating the unemployment rate into the forecasting models. Specifically, we augment the LLM-CPI models (abbreviated as LLM+LDA and LLM+BERT) with the unemployment rate and compare them against models \eqref{equ:ar_and_unem}, \eqref{equ:random_walk}, and \eqref{equ:mean_model}, as well as the direct text-based model \eqref{equ:target2} with LDA and BERT embeddings (i.e., without the LLM-CPI model structure). The ARX model serves as the baseline in this setting, and the relative forecast performance is evaluated using the relative root mean squared prediction error with respective to the ARX model, where we replace the denominator of \eqref{equ:rPMSE} with the performance of ARX. We denote the resulting performance measure as $\operatorname{rPMSE}^{ARX}(H)$. Similarly, denote by $\operatorname{rSign}^{ARX}(H)$ the corresponding relative sign prediction error with respective to the ARX model.

Tables \ref{tab:real_rpmse_with_unemp} and \ref{tab:real_rsign_with_unemp} summarize the model performances across different horizons $H = 8$ to $15$. The empirical results align with the findings in Section \ref{sec:real_data}. First, including the LDA embeddings can improve the prediction performance. Both $\operatorname{rPMSE}^{ARX}(H)$ and $\operatorname{rSign}^{ARX}(H)$ decrease compared to the benchmark models. Second, the LLM-CPI method improves the prediction accuracy further. We see that the LLM+LDA model achieves the lowest average $\operatorname{rPMSE}^{ARX}(H)$ ($0.811$) and the lowest average $\operatorname{rSign}^{ARX}(H)$ ($0.230$) among all models, and the performance of the LLM+BERT model is significantly better than the non-LLM-powered BERT model.

\begin{table}[t]
\centering
\small
\caption{The $\operatorname{rPMSE}_m^{ARX}(H)$ results across different horizons $H$ with unemployment rate.}
\label{tab:real_rpmse_with_unemp}
\begin{tabular}{lccccccccc}
\toprule
Method     & $8$ & $9$ & $10$ & $11$ & $12$ & $13$ & $14$ & $15$ & Ave. \\
\midrule
RW         & 0.750 & 1.010 & 1.494 & 2.375 & 0.887 & 0.997 & 0.956 & 1.244 & 1.214 \\
AVE        & 0.805 & 0.728 & 0.811 & 1.339 & 0.898 & 0.998 & 1.059 & 1.021 & 0.957 \\
LDA        & 0.743 & 0.773 & 0.781 & 0.817 & 0.866 & 0.887 & 0.881 & 0.871 & 0.827 \\
BERT         & 1.608 & 1.685 & 1.798 & 1.400 & 1.321 & 1.374 & 1.416 & 1.454 & 1.507 \\
LLM+LDA    & 0.743 & 0.693 & 0.706 & 0.584 & 0.869 & 0.930 & 0.968 & 0.989 & 0.811 \\
LLM+BERT     & 0.719 & 0.806 & 0.836 & 1.021 & 1.006 & 1.046 & 1.032 & 1.012 & 0.935 \\
\bottomrule
\end{tabular}\\
{\footnotesize Note: The relative PMSE values compared to ARX with unemployment rate as the exogenous variable benchmark. Smaller values indicate better performance.}
\end{table}

\begin{table}[t]
\centering
\small
\caption{The $\operatorname{rSign}_m^{ARX}(H)$ results across different horizons $H$ with unemployment rate.}
\label{tab:real_rsign_with_unemp}
\begin{tabular}{lccccccccc}
\toprule
Method     & $8$ & $9$ & $10$ & $11$ & $12$ & $13$ & $14$ & $15$ & Ave. \\
\midrule
RW         & 0.600 & 0.429 & 0.429 & 1.333 & 0.571 & 0.571 & 0.500 & 1.222 & 0.707 \\
AVE        & 0.600 & 0.429 & 0.571 & 1.333 & 0.857 & 0.714 & 1.000 & 1.000 & 0.813 \\
LDA        & 0.600 & 0.714 & 0.714 & 0.833 & 0.571 & 0.571 & 0.500 & 0.556 & 0.632 \\
BERT         & 1.143 & 1.286 & 1.429 & 1.833 & 2.400 & 2.600 & 2.800 & 2.500 & 1.999 \\
LLM+LDA    & 0.200 & 0.143 & 0.143 & 0.167 & 0.286 & 0.429 & 0.250 & 0.222 & 0.230 \\
LLM+BERT     & 0.200 & 0.429 & 0.286 & 0.833 & 0.857 & 0.857 & 0.625 & 0.556 & 0.580 \\
\bottomrule
\end{tabular}\\
{\footnotesize Note: The relative sign prediction error values  compared to ARX with unemployment rate as the exogenous variable benchmark. Smaller values indicate better performance.}
\end{table}

\subsection{High-frequency CPI prediction inference by LLM-CPI with unemployment} \label{new.Sec.add.real.inference.bj}

We also extend the CPI prediction inference evaluation by incorporating the unemployment rate into the forecasting models. We adopt the Box--Jenkins (BJ) method for constructing the prediction interval and evaluate models over different horizons $H = 8$ to $15$.

Table \ref{tab:real_with_unemp} reports the prediction interval coverage rates and interval lengths under each setting. All models, including the ARX, LDA, BERT, and LLM-CPI variants maintain the nominal coverage across all horizons. In terms of the inference power, the LLM+LDA method achieves the smallest average interval length ($3.273$ with unemployment rate) while preserving high coverage, outperforming all the baselines. The LLM+BERT method also improves over its BERT-only counterpart in both efficiency and robustness. These empirical results confirm that the LLM-CPI model enhances both predictive accuracy and uncertainty quantification with unemployment rate.

\begin{table}[t]
\centering
\small
\caption{The $\operatorname{Coverage}_m(H)$ and $\operatorname{Length}_m(H)$ results across different horizons $H$ with unemployment rate (LLM with BJ interval).}
\label{tab:real_with_unemp}
\begin{tabular}{lccccccccc}
\toprule
Method             & $8$ & $9$ & $10$ & $11$ & $12$ & $13$ & $14$ & $15$ & Ave. \\
\midrule
 ARX          & 1.000 & 1.000 & 1.000 & 1.000 & 1.000 & 1.000 & 1.000 & 1.000 & 1.000 \\
            & (4.192) & (4.227) & (4.267) & (4.194) & (4.159) & (4.204) & (4.222) & (4.269) & (4.205) \\
LDA         & 1.000 & 1.000 & 1.000 & 1.000 & 1.000 & 1.000 & 1.000 & 1.000 & 1.000 \\
            & (3.792) & (3.828) & (3.880) & (3.886) & (3.796) & (3.843) & (3.889) & (3.941) & (3.857) \\
BERT          & 1.000 & 1.000 & 1.000 & 1.000 & 1.000 & 1.000 & 1.000 & 1.000 & 1.000 \\
            & (4.153) & (4.192) & (4.239) & (4.230) & (4.191) & (4.244) & (4.291) & (4.351) & (4.236) \\
LLM+LDA     & 1.000 & 1.000 & 1.000 & 1.000 & 0.917 & 0.923 & 0.929 & 0.933 & 0.963 \\
            & (3.225) & (3.248) & (3.263) & (3.329) & (3.230) & (3.250) & (3.320) & (3.320) & (3.273) \\
LLM+BERT      & 1.000 & 1.000 & 1.000 & 1.000 & 1.000 & 1.000 & 1.000 & 1.000 & 1.000 \\
            & (3.538) & (3.411) & (3.423) & (3.412) & (3.447) & (3.487) & (3.527) & (3.564) & (3.476) \\
\bottomrule
\end{tabular}\\
{\footnotesize Note: The values in the parentheses are interval length, and coverage near the nominal level of $0.95$ with smaller interval length is preferred.}
\end{table}

\subsection{High-frequency CPI prediction inference by LLM-CPI with bootstrap} \label{new.Sec.add.real.inference.boot}

In this subsection, we further assess the CPI prediction inference performance of various forecasting models using the bootstrap prediction interval. We consider two settings: one excluding the macroeconomic predictor (i.e., unemployment rate), and the other incorporating it. Across both cases, we evaluate the coverage probability and interval length over different horizons $H = 8$ to $15$.

We first examine the inference performance without the unemployment rate. Table \ref{tab:real_boot_no_unemployment} lists the coverage and length of bootstrap prediction intervals when the unemployment rate is excluded from all models. From Table \ref{tab:real_boot_no_unemployment}, we see that the LLM-CPI variants achieve nearly the nominal coverage rate across most horizons. More importantly, the LLM-CPI model yields tighter prediction intervals. For example, the LLM+LDA method achieves the smallest average interval length of $3.359$, significantly outperforming both the AR benchmark ($4.159$) and standalone LDA ($3.813$). The LLM+BERT method also reduces the interval length relative to the standalone BERT ($4.065$ vs. $4.216$), although its intervals are wider than those of the LLM+LDA method.

\begin{table}[t]
\centering
\small
\caption{The $\operatorname{Coverage}_m(H)$ and $\operatorname{Length}_m(H)$ results across different horizons $H$ without unemployment rate (LLM with bootstrap interval).}
\label{tab:real_boot_no_unemployment}
\begin{tabular}{lccccccccc}
\toprule
Method           & $8$ & $9$ & $10$ & $11$ & $12$ & $13$ & $14$ & $15$ & Ave. \\
\midrule
AR          & 1.000 & 1.000 & 1.000 & 1.000 & 1.000 & 1.000 & 1.000 & 1.000 & 1.000 \\
            & (4.093) & (4.133) & (4.179) & (4.148) & (4.115) & (4.164) & (4.195) & (4.247) & (4.159) \\
LDA         & 1.000 & 1.000 & 1.000 & 1.000 & 1.000 & 1.000 & 1.000 & 1.000 & 1.000 \\
            & (3.752) & (3.788) & (3.838) & (3.843) & (3.753) & (3.799) & (3.842) & (3.894) & (3.813) \\
BERT          & 1.000 & 1.000 & 1.000 & 1.000 & 1.000 & 1.000 & 1.000 & 1.000 & 1.000 \\
            & (4.135) & (4.176) & (4.224) & (4.208) & (4.173) & (4.225) & (4.267) & (4.324) & (4.216) \\
LLM+LDA     & 1.000 & 1.000 & 1.000 & 1.000 & 1.000 & 1.000 & 0.929 & 0.933 & 0.982 \\
            & (3.426) & (3.423) & (3.443) & (3.528) & (3.210) & (3.259) & (3.280) & (3.299) & (3.359) \\
LLM+BERT      & 1.000 & 1.000 & 1.000 & 0.909 & 1.000 & 1.000 & 1.000 & 1.000 & 0.989 \\
            & (3.972) & (3.920) & (4.026) & (4.148) & (4.062) & (4.229) & (4.226) & (4.246) & (4.065) \\
\bottomrule
\end{tabular}\\
{\footnotesize Note: The values in the parentheses are interval length, and coverage near the nominal level of $0.95$ with smaller interval length is preferred.}
\end{table}

We next evaluate the inference performance with  unemployment rate. The empirical results are presented in Table \ref{tab:real_combined_boot_with_unemp}. We note that the LLM+LDA and LLM+BERT methods with bootstrap prediction intervals achieve the nominal coverage rate. More importantly, both LLM+LDA and LLM-BERT maintain a small average interval length. The bootstrap interval results confirm the earlier findings from using the BJ interval, where the LLM-CPI model, particularly the LLM+LDA variant, consistently produces shorter prediction intervals without sacrificing coverage, demonstrating strong inference efficiency. The inclusion of macroeconomic information such as the unemployment rate introduces mild increase in interval length and slight coverage deterioration for the LLM-powered models, but the overall performance remains robust.

\begin{table}[t]
\centering
\small
\caption{The $\operatorname{Coverage}_m(H)$ and $\operatorname{Length}_m(H)$ results across different horizons $H$ with unemployment rate (LLM with bootstrap interval).}
\label{tab:real_combined_boot_with_unemp}
\begin{tabular}{lccccccccc}
\toprule
Method           & $8$ & $9$ & $10$ & $11$ & $12$ & $13$ & $14$ & $15$ & Ave. \\
\midrule
ARX         & 1.000 & 1.000 & 1.000 & 1.000 & 1.000 & 1.000 & 1.000 & 1.000 & 1.000 \\
            & (4.192) & (4.227) & (4.267) & (4.194) & (4.159) & (4.204) & (4.222) & (4.269) & (4.217) \\
LDA         & 1.000 & 1.000 & 1.000 & 1.000 & 1.000 & 1.000 & 1.000 & 1.000 & 1.000 \\
            & (3.792) & (3.828) & (3.880) & (3.886) & (3.796) & (3.843) & (3.889) & (3.941) & (3.857) \\
BERT          & 1.000 & 1.000 & 1.000 & 1.000 & 1.000 & 1.000 & 1.000 & 1.000 & 1.000 \\
            & (4.153) & (4.192) & (4.239) & (4.230) & (4.191) & (4.244) & (4.291) & (4.351) & (4.236) \\
LLM+LDA     & 1.000 & 1.000 & 1.000 & 1.000 & 1.000 & 0.923 & 1.000 & 0.933 & 0.982 \\
            & (3.478) & (3.477) & (3.523) & (3.608) & (3.238) & (3.331) & (3.337) & (3.308) & (3.413) \\
LLM+BERT      & 1.000 & 1.000 & 1.000 & 0.909 & 1.000 & 1.000 & 1.000 & 1.000 & 0.989 \\
            & (4.132) & (4.150) & (4.025) & (4.081) & (4.186) & (4.175) & (4.214) & (4.209) & (4.147) \\
\bottomrule
\end{tabular}\\
{\footnotesize Note: The values in the parentheses are interval length, and coverage near the nominal level of $0.95$ with smaller interval length is preferred.}
\end{table}

\subsection{Out-of-sample forecastings for the pre- and during-lockdown and post-lockdown} \label{new.Sec.add.real.forecast.pand}

This subsection presents the empirical results for the pre- and during-lockdown period, and the post-lockdown period. Since each period has only $36$ observations, we set prediction horizon $H \in \{2,\ldots, 6\}$. Tables \ref{tab:COV19_19-21_mse} and \ref{tab:COV19_19-21_sign} report the model performances across different horizons $H = 2$ to $6$ for the pre- and during-lockdown period. The empirical results highlight the advantages of the LLM-CPI method. First, including the LDA embeddings can improve the prediction performance. Both $\sqrt{\overline{\operatorname{PMSE}}(H)}$ and $\overline{\operatorname{Sign}}(H)$ decrease compared to the benchmark models. Second, the LLM-CPI variants improve the prediction accuracy further. In particular, we see that the LLM+LDA method achieves the lowest average $\sqrt{\overline{\operatorname{PMSE}}(H)}$ ($0.333$) and lowest average $\overline{\operatorname{Sign}}(H)$ ($0.123$) among all models. The LLM-CPI method also performs better than all benchmark methods for the post-lockdown period, as shown in Tables \ref{tab:COV19_21-23_mse} and \ref{tab:COV19_21-23_sign}.

\begin{table}[H]
\centering
\small
\caption{The $\sqrt{\overline{\operatorname{PMSE}}_m(H)}$ results across different horizons $H$ for the pre- and during-lockdown period.}
\label{tab:COV19_19-21_mse}
\begin{tabular}{lcccccc}
\toprule
  Method     & $2$ & $3$ & $4$ & $5$ & $6$ & Ave. \\
\midrule
AR        & 0.104 & 0.685 & 0.537 & 0.466 & 0.394 & 0.437 \\
RW        & 1.184 & 0.797 & 0.632 & 0.583 & 1.088 & 0.857 \\
AVE       & 0.812 & 0.829 & 0.557 & 0.714 & 0.986 & 0.779 \\
LDA       & 0.003 & 0.587 & 0.489 & 0.469 & 0.348 & 0.379 \\
LLM+LDA   & 0.056 & 0.396 & 0.257 & 0.534 & 0.423 & 0.333 \\
\bottomrule
\end{tabular}\\
{\footnotesize Note: The root PMSE values. Smaller values indicate better performance.}
\end{table}

\begin{table}[t]
\centering
\small
\caption{The $\overline{\operatorname{Sign}}_m(H)$ results across different horizons $H$ for the pre- and during-lockdown period.}
\label{tab:COV19_19-21_sign}
\begin{tabular}{lcccccc}
\toprule
Method    & $2$ & $3$ & $4$ & $5$ & $6$ & Ave. \\
\midrule
AR        & 0.000 & 1.000 & 1.000 & 0.600 & 0.500 & 0.620 \\
RW        & 0.500 & 1.000 & 0.500 & 0.400 & 0.833 & 0.647 \\
AVE       & 0.500 & 0.667 & 0.500 & 0.600 & 1.000 & 0.653 \\
LDA       & 0.000 & 0.333 & 0.500 & 0.600 & 0.500 & 0.387 \\
LLM+LDA   & 0.000 & 0.000 & 0.250 & 0.200 & 0.167 & 0.123 \\
\bottomrule
\end{tabular}\\
{\footnotesize Note: The sign prediction error values. Smaller values indicate better performance.}
\end{table}

\begin{table}[H]
\centering
\small
\caption{The $\sqrt{\overline{\operatorname{PMSE}}_m(H)}$ results across different horizons $H$ during the post-lockdown period.}
\label{tab:COV19_21-23_mse}
\begin{tabular}{lcccccc}
\toprule
Method       & $2$ & $3$ & $4$ & $5$ & $6$ & Ave.  \\
\midrule
AR      & 0.847 & 0.587 & 0.505 & 0.521 & 0.478 & 0.587 \\
RW       & 0.807 & 1.132 & 1.177 & 0.884 & 0.909 & 0.982 \\
AVE      & 0.728 & 1.207 & 0.901 & 0.604 & 0.837 & 0.855 \\
LDA      & 0.051 & 0.047 & 0.233 & 0.434 & 0.501 & 0.253 \\
LLM+LDA & 0.002 & 0.013 & 0.250 & 0.273 & 0.325 & 0.172 \\
\bottomrule
\end{tabular}\\
{\footnotesize Note: The root PMSE values. Smaller values indicate better performance.}
\end{table}

\begin{table}[H]
\centering
\small
\caption{The $\overline{\operatorname{Sign}}_m(H)$ results across different horizons $H$ during the post-lockdown period.}
\label{tab:COV19_21-23_sign}
\begin{tabular}{lcccccc}
\toprule
Method       & $2$ & $3$ & $4$ & $5$ & $6$ & Ave.\\
\midrule
AR       & 1.000 & 1.000 & 1.000 & 1.000 & 1.000 & 1.000 \\
RW       & 0.500 & 0.667 & 0.500 & 0.400 & 0.667 & 0.547 \\
AVE      & 0.000 & 0.667 & 0.500 & 0.400 & 0.667 & 0.447 \\
LDA      & 0.000 & 0.000 & 0.250 & 0.600 & 0.500 & 0.270 \\
LLM+LDA & 0.000 & 0.000 & 0.250 & 0.200 & 0.333 & 0.157 \\
\bottomrule
\end{tabular}\\
{\footnotesize Note: The sign prediction error values. Smaller values indicate better performance.}
\end{table}

\subsection{LDA topic words and related hashtags on Weibo} \label{new.Sec.add.real.LDAtopic}

To further explore the text factors influencing the CPI prediction, we analyze in this subsection the top $10$ keywords identified for each topic selected in the real data application using the LDA embeddings. Specifically, we match these keywords with the original post to find the hashtags in each post. We then calculate the frequencies of each keyword's top $10$ most-mentioned hashtags. As hashtags represent key discussion topics on social media (e.g., Weibo), their concise and headline-like format allows users to quickly identify, share, and discuss these critical issues. Such process is repeated for the keywords identified at different stages in our real data application, aiming to uncover the underlying issues that truly drive the inflation fluctuations. These results are presented in Tables \ref{tab:5year_19-23_tab1}--\ref{tab:COV19_21-23_tab4} that correspond to Figures \ref{fig:wordcloud5year} and \ref{fig:topic12_19}--\ref{fig:wordcloud4}.

\begin{table}[t]
\centering
\small
\caption{Topic words in Figure \ref{fig:topic1_e} and related hashtags on Weibo.}
\label{tab:5year_19-23_tab1}
\begin{tabular}{@{}l>{\raggedright\arraybackslash}p{10cm}@{}}
\toprule
\textbf{Topic Words} & \textbf{Related Hashtags on Weibo and Frequency} \\ 
\midrule
Price & Real Estate (10392); Finance (10300); Futures (8547); Stocks (6898); Quant Hedge Funds (5482); Housing Prices (5461); Home Purchase (5400); Market Watch (4832); Property (4795); Shenzhen Property Market (3968) \\

Rise & Finance (9188); Real Estate (9121); Stocks (7715); Housing Prices (6420); Futures (5725); Market Watch (5375); Property (4505); Home Purchase (4262); Gold (3410); Crude Oil (3328) \\

Increase & Finance (3609); Stocks (3511); Real Estate (3165); Market Watch (2801); Housing Prices (2589); Futures (1723); Home Buying Guide (1375); Property (1343); Home Purchase (1304); Crude Oil (1218) \\

Impact & Real Estate (7323); Finance (6544); Stocks (5448); Futures (4775); Market Watch (3654); Home Purchase (3058); Housing Prices (3020); Property (2894); Investment (2194); Quant Hedge Funds (2173) \\

CPI & Finance (1564); Stocks (1199); Gold (838); Quant Hedge Funds (743); Market Watch (736); Futures (700); Investment (559); Crude Oil (513); US Stocks (496); Forex (401) \\

Soar & Coconut Water Price Surge 4000\% (1103); KN95 Mask Price 600\% Increase (967); Real Estate (757); Stocks (752); Finance (693); Seoul Housing Price +52\% (521); Housing Prices (501); N95 Mask Search +715\% (427); Home Purchase (407); Central Bank Investigates Shenzhen Housing Price Surge (407) \\

Loss & Experts Warn Homebuyers Bear Brunt of Price Drops (855); Real Estate (219); Home Purchase (177); News Highlights (146); Housing Prices (121); Property (101); Property Discussion (62); Shenzhen Property Market (55); Property Headlines (53); Property Market (43) \\

Crisis & Real Estate (670); Property (302); Home Purchase (280); Housing Prices (216); Finance (191); Stocks (187); Property Market (108); Market Watch (94); Property Discussion (91); Shenzhen Property Market (69) \\

Region & Real Estate (2075); Futures (1811); Finance (1744); Stocks (1130); Housing Prices (901); Home Purchase (843); Market Watch (801); Investment (781); Property (775); Home Buying Guide (733) \\

Fall & Real Estate (5795); Finance (4898); Stocks (4483); Housing Prices (4159); Futures (3399); Market Watch (3085); Property (3021); Home Purchase (2907); Gold (2492); Crude Oil (2285) \\
\bottomrule
\end{tabular}\\
{\footnotesize Note: Match the hashtags in the original post where the topic word is located and count their frequencies of occurrence. The values in parentheses indicate the frequency of each hashtag.}
\end{table}

\begin{table}[t]
\centering
\small
\caption{Topic words in Figure \ref{fig:topic10_e} and related hashtags on Weibo.}
\label{tab:5year_19-23_tab2}
\begin{tabular}{@{}l>{\raggedright\arraybackslash}p{10cm}@{}}
\toprule
\textbf{Topic Words} & \textbf{Related Hashtags on Weibo and Frequency} \\ 
\midrule
Stock & Real Estate (1883); Existing First-home Mortgage Rate Reduction (1228); Finance (748); Existing Mortgage Rate Adjustment Demands (743); Outstanding Mortgage (705); Daily Market Brief (701); Property (694); Stocks (690); First-home Loan Rate Cut Implementation (642); Tianjin Housing Market (591) \\

Interest rate & Real Estate (5398); Finance (4716); Stocks (3828); Quant Hedge Funds (2564); Home Purchase (2525); Market Watch (2434); Mortgage (2237); Property (2111); Home Buying Guide (1882); Gold (1800) \\

Bank & Real Estate (6972); Finance (4655); Stocks (4424); Home Purchase (3676); Property (3205); Market Watch (2873); Home Buying Guide (2622); Housing Prices (2599); Quant Hedge Funds (2015); Investment (1684) \\

Floor & Regional First-home Loan Rate Floor Removal (364); Real Estate (351); Finance (263); Stocks (258); 16\% Mortgage Cost Reduction (218); City-specific Rate Floor Relaxation (215); Existing Loan Rate Adjustment (190); Mortgage (184); Home Buying Guide (180); Outstanding Mortgage (172) \\

Cut & Real Estate (2032); Finance (2028); Stocks (1561); Futures (1323); Oil Prices (1275); Home Buying Guide (1090); Market Watch (1078); Investment (710); Fuel Price Reduction (705); Mortgage (704) \\

Central bank & Finance (3488); Quant Hedge Funds (3099); Real Estate (2921); Stocks (2826); Home Buying Guide (1699); Market Watch (1629); Gold (1376); Investment (1237); Forex (1151); Futures (1121) \\

Deposit & Experts Suggest Using 1/3 Savings for Housing (3413); Real Estate (883); Finance (807); Stocks (790); Home Purchase (561); Market Watch (505); 0.25\% Reserve Ratio Cut (417); Property (399); Housing Prices (383); Reserve Requirement Reduction (355) \\

Basis point & Finance (1412); Stocks (1178); Market Watch (886); Quant Hedge Funds (774); Gold (632); Real Estate (515); Forex (507); Forex Gold (459); Futures (454); Mortgage (451) \\

Year term & Finance (1024); Stocks (789); Real Estate (729); Home Buying Guide (532); LPR Policy (497); Market Watch (485); Gold (473); Quant Hedge Funds (465); Mortgage (429); Investment (402) \\

Interest rate cut & Finance (1334); Stocks (1291); Rate Cut Policy (1079); Real Estate (808); Market Watch (733); Gold (561); Quant Hedge Funds (477); Investment (471); Central Bank Rate Cut (462); Federal Reserve (459) \\
\bottomrule
\end{tabular}\\
{\footnotesize Note: Match the hashtags in the original post where the topic word is located and count their frequencies of occurrence. The values in parentheses indicate the frequency of each hashtag.}
\end{table}

\begin{table}[t]
\centering
\small
\caption{Topic words in Figure \ref{fig:topic12_19} and related hashtags on Weibo.}
\label{tab:COV19_19-21_tab}
\begin{tabular}{@{}l>{\raggedright\arraybackslash}p{10cm}@{}}
\toprule
\textbf{Topic Words} & \textbf{Related Hashtags on Weibo and Frequency} \\ 
\midrule
Hefei City & Hefei Property Market (9267); Hefei Local Affairs (8132); Real Estate (6565); Hefei City Updates (3479); Hefei Home Purchase (1431); Hefei News Exposure (1108); Hefei Lifestyle (1026); Property (870); Property Discussion (825); Hefei-Anhui Hot Topics (655) \\

Xian City & Xian Property Market (6709); Xian City News (3266); Xian Local Affairs (2575); Xian Property Listings (2410); Xian Home Purchase (2181); Real Estate (2120); Xian Housing Prices (1991); Xian Market Updates (1424); Property (1057); Xian Property Market Insights (804) \\

Bankruptcy & Real Estate (972); Housing Prices (467); Home Purchase (429); Property (420); 100 Developers Bankruptcy 2023 (409); Finance (336); 208 Developer Bankruptcies 2022 (287); 271 Developer Bankruptcies (258); Stocks (254); Property Discussion (213) \\

Finally & Real Estate (1282); Cherry Price Decline (1254); Home Purchase (775); Property (488); Housing Prices (469); Shenzhen Property Market (445); Stocks (397); Finance (385); Property Discussion (362); 6 Students Co-buy Hangzhou Courtyard (326) \\

Close to Me & Real Estate (9132); Hefei Local Affairs (8132); Shenzhen Community News (3603); Xian Local Affairs (2575); Hefei Property Market (2446); Chongqing Community News (2439); Home Purchase (2353); Hefei City Updates (1968); Housing Prices (1866); Property Discussion (1848) \\

Surge & Coconut Water Price Surge 4000\% (1103); KN95 Mask Price 600\% Increase (967); Real Estate (757); Stocks (752); Finance (693); Seoul Housing Price +52\% (521); Housing Prices (501); N95 Mask Search +715\% (427); Home Purchase (407); Shenzhen Housing Price Investigation (407) \\

Anhui City & Anhui Fruit Price Hike (1208); Hefei Property Market (935); Hefei-Anhui Hot Topics (655); Hefei Local Affairs (489); Real Estate (418); Anhui Community News (298); Hefei City Updates (213); Property Market (170); Finance (168); Anhui Birth Rate Plunge (162) \\

Debt Crisis & Real Estate (670); Property (302); Home Purchase (280); Housing Prices (216); Finance (191); Stocks (187); Property Market (108); Market Watch (94); Property Discussion (91); Shenzhen Property Market (69) \\

Publish & Finance (400); Real Estate (344); Gold (311); Stocks (241); Crude Oil (206); Forex (169); Futures (167); Market Watch (159); Home Purchase (153); Investment (134) \\

Suddenly & Real Estate (832); Stocks (567); Finance (547); Home Purchase (448); Paper Mills Shutdown Wave (411); Property (402); Housing Prices (376); Holiday Rental Price Surge (368); Paper Price Rally (357); Forgotten Property Appreciates to ¥6M (312) \\
\bottomrule
\end{tabular}\\
{\footnotesize Note: Match the hashtags in the original post where the topic word is located and count their frequencies of occurrence. The values in parentheses indicate the frequency of each hashtag.}
\end{table}

\begin{table}[t]
\centering
\small
\caption{Topic words in Figure \ref{fig:topic16_21} and related hashtags on Weibo.}
\label{tab:COV19_21-23_tab3}
\begin{tabular}{@{}l>{\raggedright\arraybackslash}p{10cm}@{}}
\toprule
\textbf{Topic Words} & \textbf{Related Hashtags on Weibo and Frequency} \\ 
\midrule
Quotation & Futures (965); Finance (901); Real Estate (778); Home Buying Guide (652); Market Watch (532); Stocks (472); Mortgage Rates (466); LPR (377); Shenzhen Property Market (374); Home Purchase (303) \\

Demand & Futures (7434); Real Estate (6824); Finance (5967); Stocks (4032); Crude Oil (3750); Market Watch (2940); Home Purchase (2834); Property (2514); Quant Hedge Funds (2395); Housing Prices (2219) \\

Adjust Downward & Real Estate (2032); Finance (2028); Stocks (1561); Futures (1323); Oil Prices (1275); Home Buying Guide (1090); Market Watch (1078); Investment (710); Oil Price Reduction (705); Mortgage Rates (704) \\

Range & Finance (2003); Real Estate (1701); Stocks (1654); Futures (1441); Oil Prices (1169); Market Watch (1130); Housing Prices (1020); Property (748); Home Purchase (743); Crude Oil (725) \\

Drop To & Finance (889); Real Estate (857); Stocks (643); Quant Hedge Funds (627); Futures (602); Home Buying Guide (551); Investment (444); Market Watch (434); Crude Oil (365); Global Oil Storage Warning (350) \\

Inventory & Futures (5329); Crude Oil (2387); Finance (2228); Real Estate (1828); Market Watch (1049); Stocks (955); Gold (908); Coke (896); Property (858); Home Purchase (810) \\

Expected & Finance (4190); Futures (4129); Real Estate (3171); Quant Hedge Funds (3107); Market Watch (2487); Crude Oil (2343); Oil Prices (1876); Investment (1757); Gold (1477); Housing Prices (1372) \\

Adjust Upward & Oil Prices (1674); Oil Price Increase (1522); Finance (1424); Futures (1008); Stocks (1007); Nationwide First-home Mortgage Rate Increase (839); Market Watch (737); Quant Hedge Funds (730); Real Estate (563); Crude Oil (544) \\

Change & Borrower Saved ¥410k in Interest Excited All Night (158); Real Estate (152); Monthly Payment Reduction Calculator (134); Central Bank Supports Mortgage Rate Adjustments (133); Property News Daily (123); Seven-year Mortgage Principal Unchanged Shock (88); Xingtai Property Market (82); Existing Mortgage Rate Reduction Policy (82); Shenzhen Property Market (71); Property (63) \\

Cycle & Stocks (2691); Finance (2548); Real Estate (2268); Market Watch (1837); Futures (1165); Investment (1087); Home Purchase (893); Property (872); Oil Prices (747); A-shares (712) \\
\bottomrule
\end{tabular}\\
{\footnotesize Note: Match the hashtags in the original post where the topic word is located and count their frequencies of occurrence. The values in parentheses indicate the frequency of each hashtag.}
\end{table}

\begin{table}[t]
\centering
\small
\caption{Topic words in Figure \ref{fig:topic13_21} and related hashtags on Weibo.}
\label{tab:COV19_21-23_tab2}
\begin{tabular}{@{}l>{\raggedright\arraybackslash}p{10cm}@{}}
\toprule
\textbf{Topic Words} & \textbf{Related Hashtags on Weibo and Frequency} \\ 
\midrule
Decline & Real Estate (5795); Finance (4898); Stocks (4483); Housing Prices (4159); Futures (3399); Market Watch (3085); Property (3021); Home Purchase (2907); Gold (2492); Crude Oil (2285) \\

Index & Stocks (9063); Market Watch (7484); Finance (7108); Investment (3043); A-shares (2995); Gold (2573); Stock Market (2543); US Stocks (2263); Shanghai Composite Index (2033); Real Estate (1827) \\

Rise & Finance (9188); Real Estate (9121); Stocks (7715); Housing Prices (6420); Futures (5725); Market Watch (5375); Property (4505); Home Purchase (4262); Gold (3410); Crude Oil (3328) \\

Today & Market Watch (20910); Stocks (12761); Finance (9780); Stock Market (4197); A-shares (3612); Investment (3172); Real Estate (2723); Futures (2675); Housing Prices (2490); Property Discussion (2182) \\

Increase Rate & Finance (3609); Stocks (3511); Real Estate (3165); Market Watch (2801); Housing Prices (2590); Futures (1723); Home Buying Guide (1375); Property (1343); Home Purchase (1305); Crude Oil (1218) \\

Gold & Gold Market (8675); Crude Oil (4424); Forex Gold (4149); Forex Gold \& Crude Oil (4023); Futures (3590); Spot Gold (3283); Finance (3251); Market Watch (2622); Stocks (2493); Forex (2443) \\

US Dollar & Gold (5509); Crude Oil (4927); Finance (4835); Futures (3925); Stocks (3202); Forex Gold (2638); Forex Gold \& Crude Oil (2591); Market Watch (2374); Quant Hedge Funds (2226); Investment (2019) \\

Slight & Futures (1993); Finance (1602); Stocks (1399); Market Watch (1335); Gold (966); Crude Oil (888); Real Estate (670); Investment (646); Coke (626); Glass (573) \\

Sharp Decline & Stocks (1852); Finance (1758); Market Watch (1396); Real Estate (1218); Housing Prices (870); Futures (851); Property (753); Home Purchase (660); Gold (636); Crude Oil (549) \\

Decrease Rate & Stocks (1959); Finance (1702); Market Watch (1600); Real Estate (1241); Futures (984); US Stocks (951); Housing Prices (786); Stock Market (723); Crude Oil (716); A-shares (681) \\
\bottomrule
\end{tabular}\\
{\footnotesize Note: Match the hashtags in the original post where the topic word is located and count their frequencies of occurrence. The values in parentheses indicate the frequency of each hashtag.}
\end{table}

\begin{table}[t]
\centering
\small
\caption{Topic words in Figure \ref{fig:topic3_21} and related hashtags on Weibo.}
\label{tab:COV19_21-23_tab1}
\begin{tabular}{@{}l>{\raggedright\arraybackslash}p{10cm}@{}}
\toprule
\textbf{Topic Words} & \textbf{Related Hashtags on Weibo and Frequency} \\ 
\midrule
City & Real Estate (20705); Home Purchase (10299); Housing Prices (9692); Property (8986); Home Buying Guide (7663); Property Market (5400); Property Discussion (5162); Shenzhen Property Market (4941); Finance (3334); Stocks (2385) \\
Income & Real Estate (5052); Home Purchase (2901); Property (2822); Housing Prices (2745); Income Growth Exceeds Housing Price Growth (1719); Finance (1556); Stocks (1269); Property Discussion (1187); Shenzhen Property Market (947); Property Market (817) \\
Population & Real Estate (4915); Home Purchase (2700); Housing Prices (2690); Property (2630); Property Discussion (1288); Property Market (1183); Shenzhen Property Market (1056); Finance (1010); Stocks (769); Home Buying Guide (601) \\
Young People & Real Estate (3033); Home Purchase (2011); Housing Prices (1733); Experts Advise Against Depleting Savings for Down Payment (1687); Property (1577); Shenzhen Property Market (1337); Professor's Advice on Delaying Home Purchase (1079); Luxury Homes (866); Property Discussion (772); Property Market (601) \\
Rent & Emergency Fund Requests (6878); Real Estate (1322); Rent Surge in Hangzhou Dumpling Shop (782); Property (750); Home Purchase (702); Housing Prices (626); China's Rent-Price Ratio Analysis (595); Family Buys Property Due to High Vacation Rentals (550); Expert Calls for Affordable Rents (532); Property Discussion (508) \\
House & Real Estate (18221); Home Purchase (13968); Property (10938); Housing Prices (9740); Property Discussion (5206); Shenzhen Property Market (4947); Property Market (3474); News Highlights (2158); Property Headlines (2105); Finance (1916) \\
Age & Average Home Buying Age: 27 (699); Expert Suggests 80-Year Mortgage Terms (669); Nanning Extends Mortgage Age Limit (437); Real Estate (410); Home Purchase (364); Property (248); Housing Prices (210); Metro City Average Buying Age: 36.9 (200); Proposal to Lower Legal Marriage Age (184); Property Discussion (182) \\
Expert & Expert Suggests Using Savings for Home Purchase (3413); Real Estate (2584); Warning Against Financial Overextension (1687); Housing Price Impact Analysis (1421); Property Market Rebound Prediction (1178); Housing Prices (1173); Housing Price Stability Discussion (1168); 40-Year Mortgage Proposal (1160); Property (1148); Fertility Rate and Housing Costs (1147) \\
Real Estate & Property Transactions (37696); Real Estate Market (32228); Home Purchase (21931); Housing Prices (14111); Property Headlines (10232); Property Analysis (10021); Property Market (9229); Property Discussion (8912); Property News (8218); Shenzhen Property Market (8138) \\
Buying a House & Home Purchases (49095); Real Estate (33831); Home Buying Guide (29506); Property (20825); Housing Prices (16858); Shenzhen Home Purchases (16334); Shenzhen Property Market (12196); Beijing Home Purchases (8568); Property Discussion (8323); Property Market (7792) \\
\bottomrule
\end{tabular}\\
{\footnotesize Note: Match the hashtags in the original post where the topic word is located and count their frequencies of occurrence. The values in parentheses indicate the frequency of each hashtag.}
\end{table}

\begin{table}[t]
\centering
\small
\caption{Topic words in Figure \ref{fig:topic17_21} and related hashtags on Weibo.}
\label{tab:COV19_21-23_tab4}
\begin{tabular}{@{}l>{\raggedright\arraybackslash}p{10cm}@{}}
\toprule
\textbf{Topic Words} & \textbf{Related Hashtags on Weibo and Frequency} \\ 
\midrule
Trillion Yuan & Real Estate (1045); Finance (508); Home Buying Guide (504); Stocks (468); China's Personal Mortgage Balance (CNY38.94T) (269); Market Watch (267); Property (219); Property Discussion (205); Property Market (199); Investment (197) \\

South Korea & Seoul Home Purchase Requires 15.2 Years' Income (No Spending) (851); Korean Fried Chicken Price Surge (100/Portion) (669); 8.9 Years' Income for Home Purchase (632); Cabbage Price Hike (30/Head) (481); Finance (377); Cabbage Price Soars to 62 (349); Seoul Housing Price Soars 52\% (316); Real Estate (308); Wealthy Parents Assist Youth Home Buying (294); Stocks (274) \\

Mobile Phone & Mobile Phone Forum Content Sharing (362); Real Estate (272); Finance (245); Stocks (216); Saved 1.3M Interest Upgrades Devices (193); Mortgage Rate Conversion to LPR (182); Home Purchase (158); Market Watch (116); Housing Prices (115); iPhone14 Global Price Increase (114) \\

Fertility Rate & Housing Costs Impact Fertility Rates (1147); China's Fertility Rate Below Safety Line (360); Real Estate (278); Housing Prices (221); Property (208); Home Purchase (207); Education Duration vs Fertility (106); Home Buying Timing Analysis (82); Korea's 2018 Fertility Rate Crisis (76); Monetary Policy Fertility Impact (74) \\

Company & Real Estate (4576); Stocks (3883); Finance (3599); Market Watch (2074); Home Buying Guide (1795); Investment (1744); Home Purchase (1564); Property (1521); Shenzhen Property Market (1318); Housing Prices (1235) \\

Product & Real Estate (2184); Finance (2015); Stocks (1994); Market Watch (1288); Futures (1065); Home Purchase (807); Investment (761); Shenzhen Property Market (757); Property (659); HeyTea Product Price Increase (625) \\

Renminbi & Finance (1548); Stocks (1438); Real Estate (1027); Market Watch (772); Home Buying Guide (740); Investment (500); RMB Exchange Rate Surge (496); Housing Prices (396); Property (350); Home Purchase (303) \\

Platform & Real Estate (1058); Finance (862); Market Watch (673); Stocks (582); Futures (460); Shenzhen Property Market (432); Holiday Price Manipulation Case (388); Urban Women Home Buying Trend (383); Property Discussion (381); Shared Bike vs Public Transport Costs (380) \\

Investigation & Real Estate (919); Finance (637); Quant Hedge Funds (562); Home Buying Guide (532); Zibo Hotel Price Gouging Case (451); Property Discussion (443); Home Purchase (413); Housing Prices (391); Property (363); Stocks (349) \\

Finance & Real Estate (507); Stocks (379); Financial Affairs (326); Investment (274); Federal Reserve (239); Home Purchase (192); Home Buying Guide (184); Transactions (179); Property (174); Housing Prices (169) \\
\bottomrule
\end{tabular}\\
{\footnotesize Note: Match the hashtags in the original post where the topic word is located and count their frequencies of occurrence. The values in parentheses indicate the frequency of each hashtag.}
\end{table}

\end{CJK}
\end{document}